%% file: main.tex
\documentclass{article} 
\usepackage{conference,times}

\input{math_commands.tex}

\usepackage{hyperref}
\hypersetup{hidelinks} 
\usepackage{url}

\usepackage{enumitem}
\usepackage{graphicx}
\usepackage{float}
\usepackage{booktabs}
\usepackage{amssymb}
\usepackage{caption}
\usepackage{tikz}
\usepackage{algorithm}
\usepackage{algpseudocode}
\usepackage[table]{xcolor}
\definecolor{bestcell}{RGB}{246,210,210}
\usepackage{array}

\title{UMAS3D: Unified Mask-Free 3D Editing with Region-Adaptive Source Preservation\\[-1pt]
{\normalfont\footnotesize Project Page: \url{https://xianlin-zeng.github.io/UMAS3D}}\par\vspace{-6pt}}

\author{%
Xianlin Zeng\textsuperscript{1}\quad
Xiaolong Shen\textsuperscript{1}\quad
Ruijie Quan\textsuperscript{1}\thanks{Corresponding author: Ruijie Quan.}\quad
Yi Yang\textsuperscript{1}\\[2pt]
\normalfont\textsuperscript{1}ReLER, College of Artificial Intelligence, Zhejiang University
}

\begin{document}

\setcounter{footnote}{1}
\maketitle
\vspace{-4pt}

\begin{figure}[H]
\centering
\includegraphics[width=0.99\linewidth]{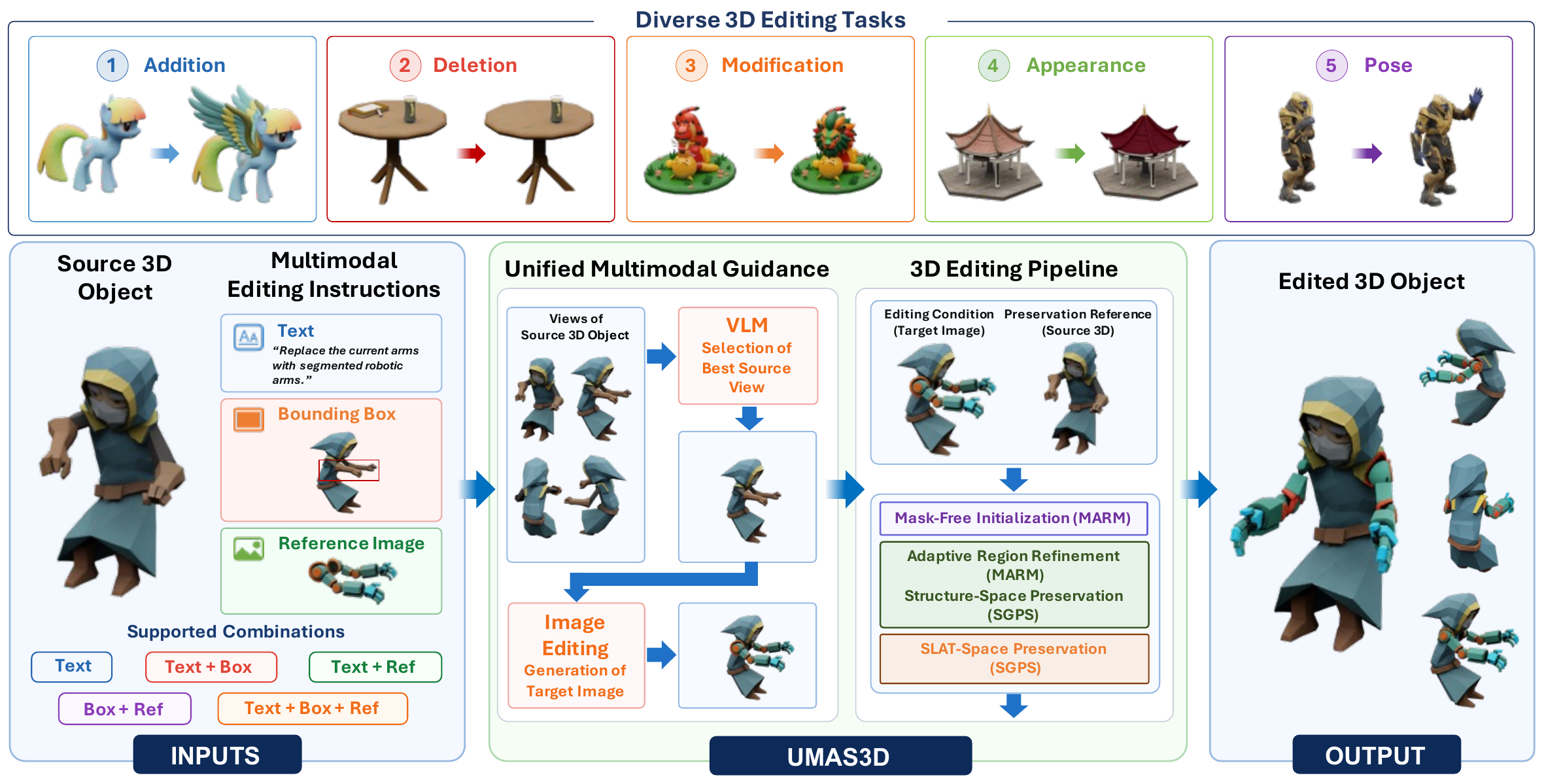}
\vspace{-2.5pt}
\caption{\textbf{Overview of UMAS3D.} A training-free framework that unifies multimodal conditions and five 3D editing tasks through adaptive region refinement and selective source preservation.}
\vspace{-5pt}
\label{fig:1}
\end{figure}


\begin{abstract}
3D editing aims to modify an existing 3D asset according to user intent while preserving geometry and appearance outside the intended edit region. Existing methods often rely on task-specific training, source-3D inversion, or external 3D masks, and lack a unified framework for diverse editing conditions and tasks. We propose UMAS3D, a unified 3D editing framework that is training-free, inversion-free, and free from external 3D masks. UMAS3D converts text, bounding boxes, reference images, and their combinations into common target-view guidance, after which addition, deletion, modification, appearance, and pose editing share the same downstream 3D generation process. Our key insight is to couple online edit-region estimation with region-aware source preservation. 
Mask-Free Adaptive Region Modeling (MARM) initializes an Edit/Keep prior from source-target visual differences and progressively refines it as the target structure evolves, while Source-Guided Preservation Steering (SGPS) constrains the current Keep-region states to retain source geometry and appearance without suppressing target-driven changes.
Experiments show that UMAS3D supports addition, deletion, modification, appearance, and pose editing while effectively preserving unedited source content. These advances make UMAS3D a unified framework for flexible and source-preserving 3D editing across diverse editing conditions.

\end{abstract}

\section{Introduction}
\input{secs/intro}

\section{Related Work}

\subsection{3D Editing Paradigms}
3D editing methods span several technical paradigms. A major line of work leverages pre-trained 2D diffusion priors to optimize 3D representations through SDS or related objectives \citep{chen2024gaussianeditor,wang2024gaussianeditor,palandra2024gsedit,kim2025dreamcatalyst,huang2025dacapo,sella2023vox,zhuang2023dreameditor,koo2024posterior,li2024focaldreamer}. Another class combines semantic localization, image-space editing signals, or generative trajectories with NeRF or 3D Gaussian representations \citep{sun2024gseditpro,lee2025editsplat,luo20243d,hong2025perturb,haque2023instruct,bao2023sine,song2023blending}. Multi-view and image-space methods propagate editing signals across views and subsequently reconstruct or optimize the underlying 3D representation \citep{chen2024dge,wu2024gaussctrl,chen2024generic,qi2024tailor3d,chen2024mvdrag3d,erkocc2025preditor3d,zhang2026vdfe,wu2026100editor}. Recent work further explores feed-forward editing directly in native 3D latent or voxel representations \citep{chen2024shap,weng2026feedforward,cai2025native,hu2026easy3e}. Spatial cues such as masks, local regions, and control points are also widely used for local 3D editing \citep{gao20253d,shen2024draggaussian,qu2025drag,kim2025robust}.

\subsection{Training-Free Native 3D Editing}
With the emergence of pre-trained 3D generative models, training-free methods leverage frozen 3D generative priors for editing \citep{ye2026nano3d,li2026voxhammer,chi2026vinedresser3d,zhou2026anchorflow,liu2026velocity,nie2026editflow3d,parelli20263d}. Some preserve unedited content through source-3D inversion. VoxHammer replaces preserved-region features with inverted latents and cached attention states, while Vinedresser3D performs inversion-based rectified-flow inpainting \citep{li2026voxhammer,chi2026vinedresser3d}. Others manipulate source-to-target generation in native voxel or latent spaces through flow trajectories, latent anchors, velocity fields, or internal features \citep{ye2026nano3d,zhou2026anchorflow,liu2026velocity,nie2026editflow3d,parelli20263d}. \textbf{UMAS3D} instead avoids source-3D inversion and updates Edit/Keep preservation support online within the same Stage-1 trajectory as the target structure evolves.

\subsection{Multimodal Guidance and 3D Edit Localization}
Visual editing interfaces span text instructions, reference images and multimodal instructions \citep{yu2025anyedit,xia2025dreamomni,xiao2025omnigen,chen2025unireal,he2025freeedit,tian2025mige}. Such controls have also been explored in 3D editing through text-image prompts, bounding boxes, and sketches \citep{zhuang2024tip,he2024customize,mikaeili2023sked}. Semantic components and local 3D regions can be localized with visual, linguistic, geometric, or interactive cues \citep{liu2025partfield,ma2025find,simonelli2025easy3d,lee2025mosaic3d,liang2026supergseg,wu2024opengaussian}. \textbf{UMAS3D} further unifies multimodal guidance and adaptive 3D edit localization, allowing Edit/Keep support to evolve with the target structure during generation.



\section{Method}
\label{Method}

\subsection{Preliminaries}

\textbf{TRELLIS.} We employ the pre-trained image-conditioned TRELLIS~\citep{xiang2025structured} as our 3D generation backbone. TRELLIS uses a two-stage process: Stage 1 produces a sparse 3D structure that determines active voxel locations and coarse geometry; Stage 2 generates Structured LATent (SLAT) features on these voxels to capture finer geometry and appearance. The resulting SLAT representation is decoded into the complete 3D asset. Throughout UMAS3D, all TRELLIS parameters remain frozen, and our editing method builds on this process without additional training.

\subsection{Framework Overview}

As shown in Figure~\ref{fig:1}, \textbf{UMAS3D} provides a unified pipeline from a source 3D asset and multimodal editing instructions to a target 3D asset. The source 3D object is first rendered into multi-view images, from which an editing-relevant source view is selected. A target image is then generated from the selected source view and multimodal editing conditions to guide the subsequent 3D editing process. Addition, deletion, modification, appearance, and pose share the same pipeline.

\begin{figure}[t]
\centering
\includegraphics[width=0.9\linewidth]{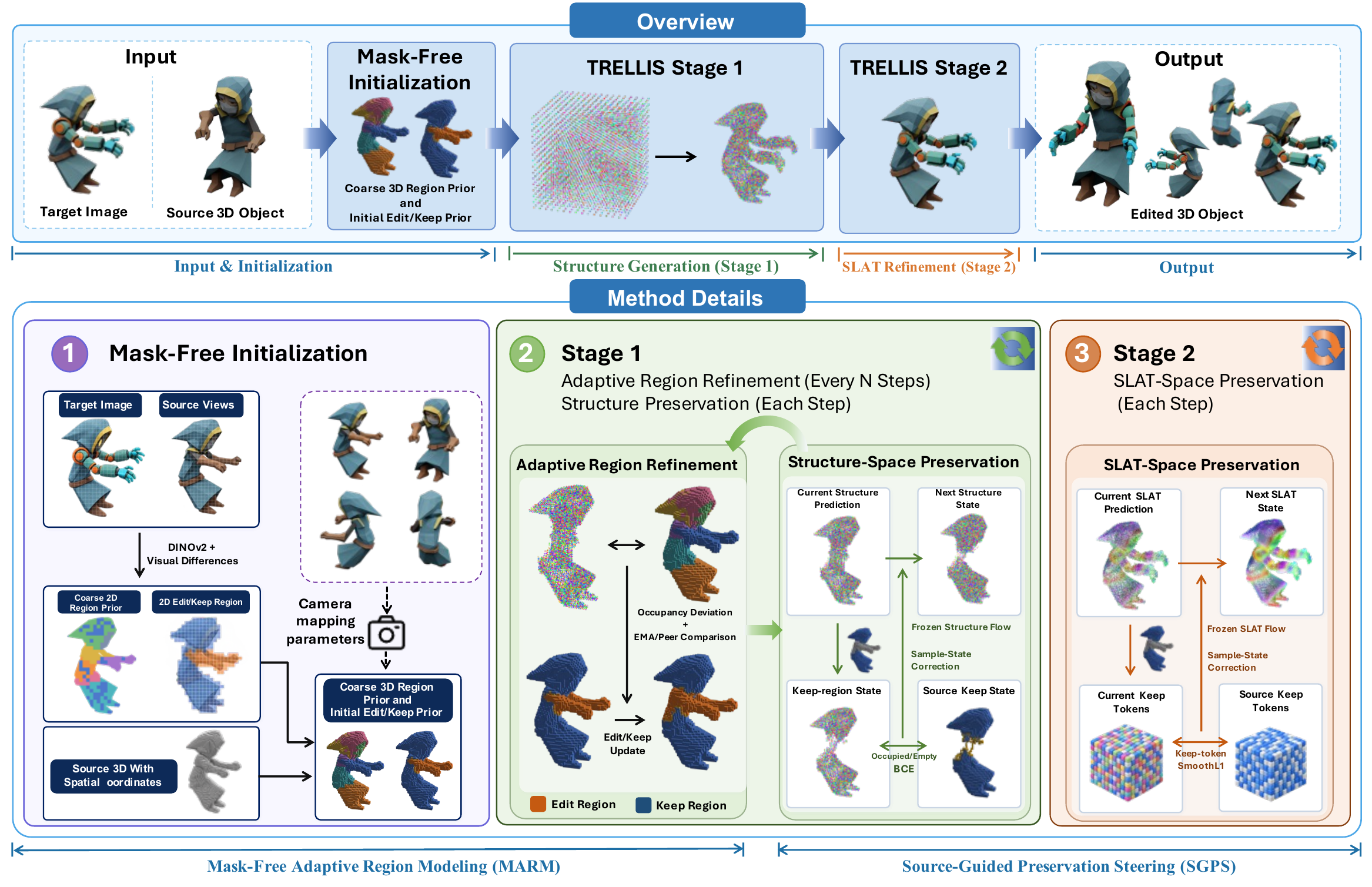}
\vspace{-0.5 em}
\caption{\textbf{Core 3D editing pipeline of UMAS3D.} MARM initializes and adaptively refines the Edit/Keep regions. SGPS preserves source-consistent content across structure and SLAT spaces.}
\vspace{-1. em}
\label{fig:2}

\end{figure}

As illustrated in Figure~\ref{fig:2}, to avoid external 3D masks while preserving unedited source content, we introduce \textbf{MARM} and \textbf{SGPS}. MARM constructs an initial 3D Edit/Keep prior and adaptively updates region assignments the target structure evolves (Section~\ref{4.3}). SGPS preserves source consistency in the current Keep regions, maintaining source-consistent structure and source-aligned SLAT features (Section~\ref{4.4}).

\subsection{Unified Multimodal Editing Guidance}
\label{4.2}

\noindent\textbf{Multimodal Condition Unification.}
UMAS3D supports text instructions, bounding boxes, and reference images, denoted by $T$, $B$, and $R$, respectively. We consider five input combinations:
\begin{equation}
\mathcal{C}\in
\{T,\;T+B,\;T+R,\;B+R,\;T+B+R\}.
\end{equation}
Given $\mathcal{C}$, Gemini 3.6 Flash~\citep{gemini36flash2026} interprets the available conditions and constructs a unified editing description $E$, while $B$ and $R$ provide spatial and visual guidance, respectively. The processing of different combinations is summarized in Table~\ref{tab:multimodal_processing}.

\noindent\textbf{Source-View Selection and Target-Image Generation.}
Given source views $\mathbf{I}_s=\{\mathbf{I}_s^v\}_{v=1}^{V}$ rendered from the source asset $\mathbf{X}_s$, the source view associated with $B$ is directly used when available; otherwise, Gemini 3.6 Flash selects the source view $\mathbf{I}_s^{v^*}$ most relevant to the intended edit. It then constructs the image-editing prompt from the selected view, $E$, and the available conditions. Gemini 3 Pro Image~\citep{gemini3proimage2025} then generates the target image $\mathbf{I}_t$:
\begin{equation}
E=\mathcal{F}_{pre}(\mathcal{C}), \qquad
\mathbf{I}_t=\mathcal{F}_{edit}
\left(
\mathbf{I}_s^{v^*},E,\delta_B B,\delta_R R
\right),
\end{equation}
where $\delta_B,\delta_R\in\{0,1\}$ indicate whether the corresponding modalities are present. For $B+R$, the editing semantics and prompt are inferred from the spatial and visual cues.

\noindent\textbf{Unified Downstream 3D Editing.}
The target image $\mathbf{I}_t$ serves as the common interface for 3D editing. Once obtained, addition, deletion, modification, appearance, and pose share the same pipeline, thereby achieving condition-level and task-level unification. Reference image conditions are not used for deletion and pose, as these edits are defined by content removal or articulation changes and do not require an external visual exemplar. The supported configurations are summarized in Table~\ref{tab:multimodal_tasks}, with further rationale provided in the Appendix.

\begin{table}[t]
\centering

\begin{minipage}[t]{0.47\textwidth}
\vspace{0pt}
\centering
\captionof{table}{Processing of multimodal conditions.}
\vspace{-.5em}
\label{tab:multimodal_processing}

\scriptsize
\setlength{\tabcolsep}{2.2pt}
\renewcommand{\arraystretch}{1.12}

\begin{tabular}{lcccc}
\toprule
\textbf{Condition} &
\textbf{Semantic} &
\textbf{Spatial} &
\textbf{Visual} &
\textbf{View Selection} \\
\midrule
$T$       & T  & -- & -- & VLM selection \\
$T+B$     & T  & B  & -- & BBox view \\
$T+R$     & T  & -- & R  & VLM selection \\
$B+R$     & VLM-inferred & B & R & BBox view \\
$T+B+R$   & T  & B  & R  & BBox view \\
\bottomrule
\end{tabular}
\end{minipage}%
\hfill%
\begin{minipage}[t]{0.50\textwidth}
\vspace{0pt}
\centering
\captionof{table}{Supported multimodal conditions.}
\vspace{-.5em}
\label{tab:multimodal_tasks}

\scriptsize
\setlength{\tabcolsep}{4.0pt}
\renewcommand{\arraystretch}{1.12}

\begin{tabular}{lccccc}
\toprule
\textbf{Edit Type} &
\textbf{T} &
\textbf{T+B} &
\textbf{T+R} &
\textbf{B+R} &
\textbf{T+B+R} \\
\midrule
Addition
& $\checkmark$ & $\checkmark$ & $\checkmark$ & $\checkmark$ & $\checkmark$ \\
Deletion
& $\checkmark$ & $\checkmark$ & $\times$ & $\times$ & $\times$ \\
Modification
& $\checkmark$ & $\checkmark$ & $\checkmark$ & $\checkmark$ & $\checkmark$ \\
Appearance
& $\checkmark$ & $\checkmark$ & $\checkmark$ & $\checkmark$ & $\checkmark$ \\
Pose
& $\checkmark$ & $\checkmark$ & $\times$ & $\times$ & $\times$ \\
\bottomrule
\end{tabular}
\end{minipage}
\vspace{-1. em}
\end{table}

\subsection{Mask-Free Adaptive Region Modeling}
\label{4.3}

\noindent\textbf{Overview.}
To avoid manual annotations and predefined 3D masks, we propose \textbf{Mask-Free Adaptive Region Modeling (MARM)}. As shown in Figure~\ref{fig:2}, \textbf{MARM} constructs a coarse 3D Edit/Keep prior from source-target visual differences and refines region states as the target structure evolves.

\noindent\textbf{Coarse Edit/Keep Initialization.}
Given the target image $\mathbf{I}_t$ and corresponding source view $\mathbf{I}_s^{v^*}$ from Section~\ref{4.2}, we reuse the frozen DINOv2~\citep{oquab2023dinov2} encoder $\phi(\cdot)$ in TRELLIS to extract patch-level features:
\begin{equation}
\mathbf{H}_t=\phi(\mathbf{I}_t),\qquad
\mathbf{H}_s=\phi(\mathbf{I}_s^{v^*}),
\end{equation}
We compare patch-level source-target DINOv2 correspondences at the same locations, together with foreground, appearance, and edge differences. Strong correspondence with small visual changes provides Keep evidence, whereas weakened correspondence with clear visual changes provides Edit evidence; target-only content without reliable source correspondence is treated as Edit. This yields
\begin{equation}
\mathbf{E}_{2D}
=
\left(
\mathbf{E}_{2D}^{E},
\mathbf{E}_{2D}^{K}
\right).
\end{equation}
To construct complete 3D regions, we use the source 3D, source views, and calibrated cameras. For each source view $v$, we extract its DINOv2 feature map as
\begin{equation}
\mathbf{H}_s^v=\phi(\mathbf{I}_s^v),\qquad v=1,\ldots,V,
\end{equation}
where $\mathbf{H}_s=\mathbf{H}_s^{v^*}$ corresponds to the selected source view. For each occupied source Grid64 voxel $x$, we project its center to visible views and aggregate the corresponding DINOv2 features:
\begin{equation}
p_v(x)=\pi(\Pi_vx),\qquad
\mathbf{f}_x=
\operatorname{Agg}_{v\in\mathcal V(x)}
\mathbf{H}_s^v\!\left(p_v(x)\right),
\end{equation}
where $\Pi_v$ is the camera projection of view $v$, $\pi(\cdot)$ denotes perspective projection, $\mathcal V(x)$ is the visible-view set, and $\mathbf f_x$ is the aggregated voxel feature. We then cluster voxels by DINOv2 feature similarity and 3D spatial distance, with connectivity regularization, to obtain 3D regions. Each region is projected through $\Pi_{v^*}$ to the 2D Edit/Keep evidence map, where the aggregated evidence determines its initial state:
\begin{equation}
\mathcal R=\{R_i\}_{i=1}^{N_R},
\qquad
\mathbf M_0=
\left(\mathbf M_0^{E},\mathbf M_0^{K}\right).
\end{equation}
$R_i$ denotes a 3D region. Edit-dominant and Keep-dominant regions are initialized as Edit and Keep; source-empty locations with strong target-side Edit evidence are opened as editable space. Thus, $\mathbf M_0$ matches 2D source-target evidence with complete multi-view 3D regions rather than directly back-projecting single 2D masks.

\noindent\textbf{Adaptive Region Refinement.}
The same 3D regions $\mathcal R$ serve as refinement units. Since $\mathbf M_0$ is estimated before the target structure is fully formed, we first allow the target structure to evolve during an initial warm-up period, after which the region states are re-evaluated every five Stage~1 steps. Let $\mathbf{O}_t$ and $\mathbf{O}_s$ denote the current target and aligned source Grid64 occupancy states used for region refinement. For each $R_i$, we compute its structural deviation and EMA:
\begin{equation}
\ell_i^t=
\frac{1}{|R_i|}
\sum_{x\in R_i}
|\mathbf O_t(x)-\mathbf O_s(x)|,\qquad
\bar{\ell}_i^t=
\lambda\bar{\ell}_i^{t-1}
+(1-\lambda)\ell_i^t .
\end{equation}
We further compare each region with others sharing the same current state:
\begin{equation}
d_i^t=
\bar{\ell}_i^t-
\underset{j\in\mathcal P(i)}
{\operatorname{Median}}
\left(\bar{\ell}_j^t\right),
\end{equation}
where $\mathcal P(i)$ denotes regions sharing the current Edit/Keep state of $R_i$. If an Edit region persistently shows smaller deviation, its structure remains close to the source and it is reassigned to Keep. Conversely, a Keep region with persistently larger target-driven deviation is released to Edit. Let $t^+$ denote the next refinement checkpoint:
\begin{equation}
\mathbf M_{t^+}(R_i)=
\begin{cases}
K, & \mathbf M_t(R_i)=E,\ d_i^t<-\tau^K,\\
E, & \mathbf M_t(R_i)=K,\ d_i^t>\tau^E,\\
\mathbf M_t(R_i), & \text{otherwise}.
\end{cases}
\end{equation}
Here, $\tau^K$ and $\tau^E$ are conservative transition margins. State changes require persistent evidence across consecutive refinement checkpoints, preventing transient fluctuations from causing unstable updates. \textbf{MARM} thus progressively updates the Edit/Keep partition as the target structure evolves and determines where SGPS applies source preservation.

\subsection{Source-Guided Preservation Steering}
\label{4.4}

\noindent\textbf{Overview.}
To preserve unedited content without restricting the intended edit, we introduce \textbf{Source-Guided Preservation Steering (SGPS)}. As shown in Figure~\ref{fig:2}, \textbf{MARM} determines where source preservation applies, while \textbf{SGPS} maintains source consistency within the current Keep regions during inference. \textbf{SGPS} preserves occupancy consistency in structure space and SLAT features. All operations update only the current generative state, with pretrained model parameters kept frozen.

\noindent\textbf{Structure-Space Preservation.}
In Stage~1, SGPS preserves source occupancy within the Keep regions. Let $\mathbf{x}_t$ denote the structure latent at timestep $t$, and $\hat{\mathbf{x}}_{0,t}$ the corresponding clean prediction, which is decoded into Grid64 occupancy logits $\mathbf{P}_t$. Given the source-occupied voxel set $\mathbf{S}_o$ and the valid non-editable empty support $\mathbf{S}_e$, the current Keep support is divided into
\begin{equation}
\mathbf{M}_t^{K_o}
=
\mathbf{M}_t^K \cap \mathbf{S}_o,
\qquad
\mathbf{M}_t^{K_e}
=
\mathbf{M}_t^K \cap \mathbf{S}_e .
\label{eq:stage1_occ_emp}
\end{equation}
\begin{equation}
\Delta\mathbf{x}_{0,t}
=
\mathbf{x}_{0,t}^{s}
-
\hat{\mathbf{x}}_{0,t},
\qquad
\tilde{\mathbf{x}}_{0,t}
=
\hat{\mathbf{x}}_{0,t}
+
\mathbf{W}_t^K
\odot
\Delta\mathbf{x}_{0,t}.
\label{eq:stage1_source_correction}
\end{equation}
The source-aligned clean state is then recombined with the corresponding noise endpoint under the original PNS rule, while non-Keep regions remain target-driven. In addition to the source-aligned clean-state correction above, the BCE term is applied only to source-occupancy violations within the current Keep regions:
\begin{equation}
\mathcal{L}_1^o
=
\operatorname{BCE}
\left(
\mathbf{P}_t[\mathcal{E}_t^o],1
\right),
\qquad
\mathcal{L}_1^e
=
\operatorname{BCE}
\left(
\mathbf{P}_t[\mathcal{E}_t^e],0
\right),
\qquad
\mathbf{x}_t
\leftarrow
\operatorname{AdamStep}
\left(
\mathbf{x}_t;
\mathcal{L}_1^o,\mathcal{L}_1^e
\right).
\label{eq:stage1_bce}
\end{equation}
where
$\mathcal{E}_t^o
=
\left\{
v\in\mathbf{M}_t^{K_o}
\mid
\mathbf{P}_t(v)\leq0
\right\}$
and
$\mathcal{E}_t^e
=
\left\{
v\in\mathbf{M}_t^{K_e}
\mid
\mathbf{P}_t(v)>0
\right\}$.
These sets capture occupied-to-empty and empty-to-occupied errors, respectively. Here, $\operatorname{AdamStep}$ denotes a sample-level IAS update of the current structure state without modifying model parameters. The updated state is re-evaluated by the frozen structure flow and advanced using the original TRELLIS update rule. After Stage~1, a final Keep-region Grid64 composition is applied.

\noindent\textbf{SLAT-Space Preservation.}
Occupancy preservation alone cannot fully retain local geometry, material, and appearance. In Stage~2, we therefore fix the final region partition from Stage~1 and extend source consistency to the SLAT feature space. Let $Z_t$ denote the current SLAT latent at timestep $t$, $Z_s$ the source SLAT feature set, and $\hat{Z}_{0,t}$ the corresponding clean SLAT prediction obtained from the frozen SLAT Flow. Source and current SLAT features are aligned through their 3D sparse coordinates. The final preserved region $\mathbf{M}_T^K$ is mapped to the Keep-token set $\Omega_{\mathrm{slat}}^K$, where
\begin{equation}
\tilde{Z}_{0,t}^{\,i}
=
Z_s^i,
\qquad
i\in\Omega_{\mathrm{slat}}^K .
\label{eq:stage2_source_alignment}
\end{equation}
$\tilde{Z}_{0,t}^{\,i}$ defines a clean state rather than directly overwriting the noisy SLAT state. It is then recombined with the fixed initial-noise endpoint; To smooth the transition between the final Keep and Edit regions, we derive a narrow boundary token set $\Omega_{\mathrm{slat}}^B$ from the final partition $\mathbf{M}_T$. This boundary band is used only for soft blending and does not introduce an additional \textbf{MARM} region state. Keep tokens retain source consistency, tokens in $\Omega_{\mathrm{slat}}^B$ are softly blended, and the remaining tokens stay target-driven. The recombined state is re-evaluated by the frozen SLAT flow, after which source supervision is applied only to Keep tokens:
\begin{equation}
\mathcal{L}_{2}
=
\frac{1}{|\Omega_{\mathrm{slat}}^K|}
\sum_{i\in\Omega_{\mathrm{slat}}^K}
\operatorname{SmoothL1}
\left(
\hat{Z}_{0,t}^{\,i},
Z_s^i
\right),
\qquad
Z_t
\leftarrow
\operatorname{AdamStep}
\left(
Z_t,\mathcal{L}_2
\right).
\label{eq:stage2_smoothl1}
\end{equation}
The SmoothL1 gradient updates only the current Keep-token state, while the remaining tokens receive no source-feature supervision and remain target-driven. The updated SLAT state is re-evaluated by the frozen SLAT Flow and advanced using the original TRELLIS update rule. Before decoding, a final Source-SLAT composition is applied on the Keep tokens. Thus, \textbf{SGPS} preserves source structure and fine-grained SLAT features while leaving the edited regions target-driven.

\section{Experiments}

\subsection{Experimental Setup}

\noindent\textbf{Datasets and Multimodal Instructions.}
We evaluate UMAS3D on 150 3D objects from
Pxform~\citep{weng2026feedforward} and Omni3DEdit~\citep{omni3dedit2026},
covering addition, deletion, modification, appearance, and pose. A small number of
Objaverse~\citep{deitke2023objaverse} assets are used only for qualitative
evaluation and excluded from quantitative statistics. Since these datasets
do not provide complete Text, Bounding Box, and Reference Image
conditions, we preserve the original editing intent and construct missing
modalities from source multi-view observations. Gemini 3.6
Flash~\citep{gemini36flash2026} generates Bounding Boxes and
modality-specific text variants, while Gemini 3 Pro
Image~\citep{gemini3proimage2025} generates Reference Images, yielding five
settings: $T$, $T+B$, $T+R$, $B+R$, and $T+B+R$. Some details are provided in the Appendix.

\noindent\textbf{Baselines.}
We compare against training-based 3DEditFormer~\citep{xia2025towards},
inversion-based VoxHammer~\citep{li2026voxhammer} and
Vinedresser3D~\citep{chi2026vinedresser3d}, training-free FlowEdit-based
Nano3D~\citep{ye2026nano3d}, and feed-forward EditP23~\citep{bar2025editp23}
and Easy3E~\citep{hu2026easy3e}. 
Details are provided in the Appendix.
For quantitative evaluation only, we render four canonical views from the ground-truth target 3D of each Pxform and Omni3DEdit sample and manually select one representative view as the target image for all methods; the resulting prediction is then evaluated against the
ground-truth target 3D. Qualitative evaluation follows the full editing pipeline.



\begin{table}[t]
\centering
\scriptsize
\setlength{\tabcolsep}{4.2pt}
\caption{
Quantitative comparison with existing 3D editing methods.
Best results are highlighted in \textbf{bold} with a light red background.
}
\vspace{-1 em}
\label{tab:comparison_results}

\resizebox{\textwidth}{!}{
\begin{tabular}{llccccccc}
\toprule
&
&
\multicolumn{3}{c}{Unedited-region Preservation} &
\multicolumn{4}{c}{Overall Target Quality} \\
\cmidrule(lr){3-5}
\cmidrule(lr){6-9}

Method
& Venue
& Pres-IoU$\uparrow$
& Pres-PSNR$\uparrow$
& Pres-SSIM$\uparrow$
& DINO-I$\uparrow$
& CD-L1$\downarrow$
& F@1\%$\uparrow$
& NC$\uparrow$ \\
\midrule

EditP23
& \textit{arXiv'25}
& 0.1745
& 7.8001
& 0.4363
& 0.5598
& 0.0612
& 0.3684
& 0.6905 \\

Easy3E
& \textit{CVPR'26}
& 0.5643
& 17.0366
& 0.7149
& 0.8568
& 0.0190
& 0.7719
& 0.8502 \\

VoxHammer
& \textit{3DV'26}
& 0.5691
& 16.1752
& 0.6056
& 0.8468
& 0.0186
& 0.7598
& 0.8521 \\

Nano3D
& \textit{ICLR'26}
& 0.5580
& 16.0651
& 0.6176
& 0.8554
& 0.0182
& 0.7696
& 0.8540 \\

Vinedresser3D
& \textit{CVPR'26}
& 0.6030
& 19.8795
& 0.7913
& 0.8797
& 0.0158
& 0.8031
& 0.8712 \\

3DEditFormer
& \textit{ICML'26}
& 0.5881
& 20.0870
& 0.7399
& 0.8878
& 0.0149
& 0.8096
& 0.8838 \\

\textbf{UMAS3D}
& \textit{Ours}
& \cellcolor{red!12}\textbf{0.7217}
& \cellcolor{red!12}\textbf{24.9473}
& \cellcolor{red!12}\textbf{0.8586}
& \cellcolor{red!12}\textbf{0.8936}
& \cellcolor{red!12}\textbf{0.0134}
& \cellcolor{red!12}\textbf{0.8194}
& \cellcolor{red!12}\textbf{0.8995} \\

\bottomrule
\end{tabular}
}
\vspace{-2em}
\end{table}

\noindent\textbf{Evaluation Metrics.}
We evaluate \emph{Unedited-region Preservation} using Pres-IoU, Pres-PSNR, and Pres-SSIM, which measure geometric and appearance preservation between the source and prediction within unedited regions. We evaluate \emph{Overall Target Quality} using DINO-I, CD-L1, F@1\%, and NC, measuring visual-semantic and geometric consistency between the prediction and the ground-truth target. Details are provided in the Appendix.


\subsection{Comparison Results}

As shown in Table~\ref{tab:comparison_results}, compared with existing methods,
UMAS3D achieves the best overall performance in both
\emph{Unedited-region Preservation} and \emph{Overall Target Quality},
demonstrating its ability to preserve unedited source content while producing
high-quality target edits. The results show that UMAS3D better preserves source
geometry and appearance, while maintaining higher consistency with the benchmark
target in terms of overall visual semantics, geometric structure, and surface
quality, thereby effectively reducing unintended changes during editing.

\begin{figure}[t]
\centering
\includegraphics[width=0.89\linewidth]{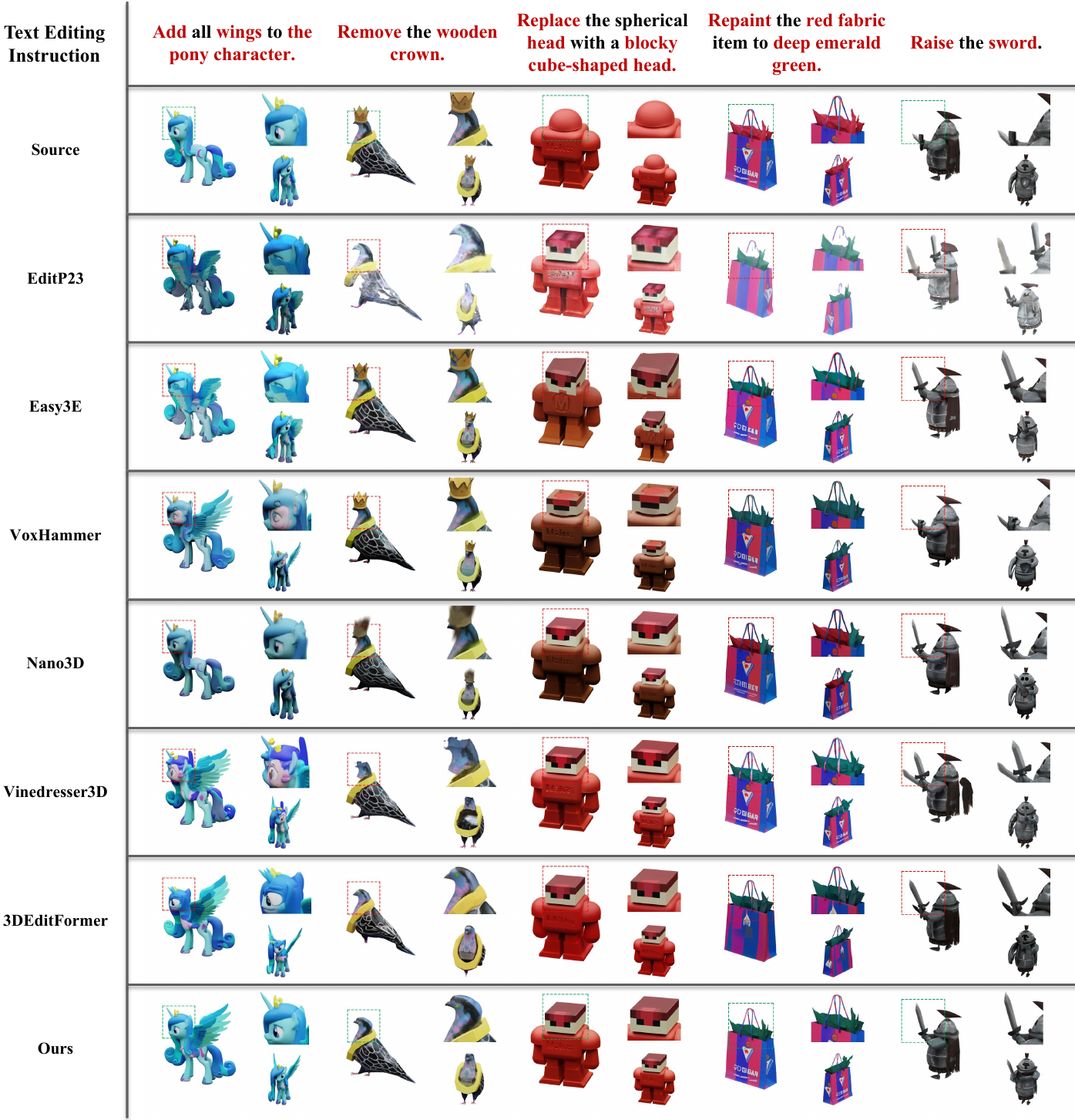}
\vspace{-1. em}
\caption{Qualitative comparison with existing 3D editing methods.}
\vspace{-1 em}
\label{fig:comparative_experiments}
\end{figure}
As shown in Figure~\ref{fig:comparative_experiments}, UMAS3D accurately follows
the specified editing intent while better preserving source structures and local
details unrelated to the edit. Compared with existing methods, its results exhibit
more complete edits, more stable geometric structures, and fewer unintended
changes in unrelated regions. Together, the quantitative and qualitative results
validate the complementary roles of \textbf{MARM} and \textbf{SGPS}, enabling \textbf{UMAS3D} to achieve
the intended edits while preserving source content more consistently.

\subsection{Human Evaluation of Editing Quality}
\label{sec:user_study}

\begin{table}[t]
\centering
\small
\setlength{\tabcolsep}{5.5pt}
\caption{Human evaluation results. Each value denotes the percentage of times a method is selected as the preferred result under the corresponding criterion.}
\vspace{-1 em}
\label{tab:user_study}
\begin{tabular}{lccc}
\toprule
Method &
\begin{tabular}[c]{@{}c@{}}Instruction Alignment $\uparrow$\end{tabular} &
\begin{tabular}[c]{@{}c@{}}Visual Quality $\uparrow$\end{tabular} &
\begin{tabular}[c]{@{}c@{}}Source Preservation $\uparrow$\end{tabular} \\
\midrule
VoxHammer        & 7\% & 7\% & 8\% \\
Nano3D     & 7\% & 9\% & 8\% \\
Vinedresser3D & 8\% & 11\% & 9\% \\
3DEditFormer  & 21\% & 19\% & 14\% \\
\textbf{UMAS3D}
               & \textbf{57\%}
               & \textbf{54\%}
               & \textbf{61\%} \\
\bottomrule
\end{tabular}
\vspace{-1. em}
\end{table}

To directly assess instruction fulfillment and perceived editing quality, we conduct a human evaluation with 40 participants. We compare UMAS3D with four representative baselines---Nano3D, VoxHammer, Vinedresser3D and 3DEditFormer. For each case, anonymized results are shown in random order under identical rendering settings, together with the source 3D asset and editing instruction. Participants select the best result in terms of Instruction Alignment, Visual Quality and Source Preservation. As shown in Table~\ref{tab:user_study}, UMAS3D receives the highest preference.

\subsection{Ablation Studies}

\begin{table}[t]
\centering
\scriptsize
\setlength{\tabcolsep}{2.2pt}
\caption{
Quantitative ablation of UMAS3D components.
\textit{Ext.}: external 3D-mask initialization;
\textit{Mask-free}: MARM initialization without external 3D masks;
\textit{S1/S2 SGPS}: Stage~1/Stage~2 SGPS;
\textit{Adapt.}: adaptive Edit/Keep refinement.
Best results are highlighted in \textbf{bold} with a light background.
}
\label{tab:ablation_studies}

\resizebox{\textwidth}{!}{
\begin{tabular}{lccccccccccc}
\toprule
&
\multicolumn{4}{c}{Components} &
\multicolumn{3}{c}{Unedited-region Preservation} &
\multicolumn{4}{c}{Overall Target Quality} \\
\cmidrule(lr){2-5}
\cmidrule(lr){6-8}
\cmidrule(lr){9-12}

Model
& Region Init.
& S1 SGPS
& S2 SGPS
& Adapt.
& Pres-IoU$\uparrow$
& Pres-PSNR$\uparrow$
& Pres-SSIM$\uparrow$
& DINO-I$\uparrow$
& CD-L1$\downarrow$
& F@1\%$\uparrow$
& NC$\uparrow$ \\
\midrule

A
& --
& --
& --
& --
& 0.1328 & 9.2586 & 0.4355 & 0.8644 & 0.0570 & 0.2861 & 0.6755 \\

B
& Ext.
& $\checkmark$
& --
& --
& 0.5437 & 16.0835 & 0.2619 & 0.8521 & 0.0162 & 0.7949 & 0.8633\\

C
& Ext.
& $\checkmark$
& $\checkmark$
& --
& 0.6438 & 17.2126 & 0.6770 & 0.8780 & 0.0153 & 0.8001 & 0.8800 \\

D
& Ext.
& $\checkmark$
& $\checkmark$
& $\checkmark$
& 0.6341 & 19.8069 & 0.6966 & 0.8798 & 0.0146 & 0.8048 & 0.8789 \\

E
& Mask-free
& $\checkmark$
& $\checkmark$
& --
& 0.7094 & 18.6103 & 0.7737 & 0.8778 & 0.0152 & 0.8022 & 0.8959 \\

\textbf{UMAS3D}
& \textbf{Mask-free}
& $\checkmark$
& $\checkmark$
& $\checkmark$
& \cellcolor{red!12}\textbf{0.7217}
& \cellcolor{red!12}\textbf{24.9473}
& \cellcolor{red!12}\textbf{0.8586}
& \cellcolor{red!12}\textbf{0.8936}
& \cellcolor{red!12}\textbf{0.0134}
& \cellcolor{red!12}\textbf{0.8194}
& \cellcolor{red!12}\textbf{0.8995} \\

\bottomrule
\end{tabular}
}
\end{table}

\begin{figure}[t]
\centering
\includegraphics[width=0.99\linewidth]{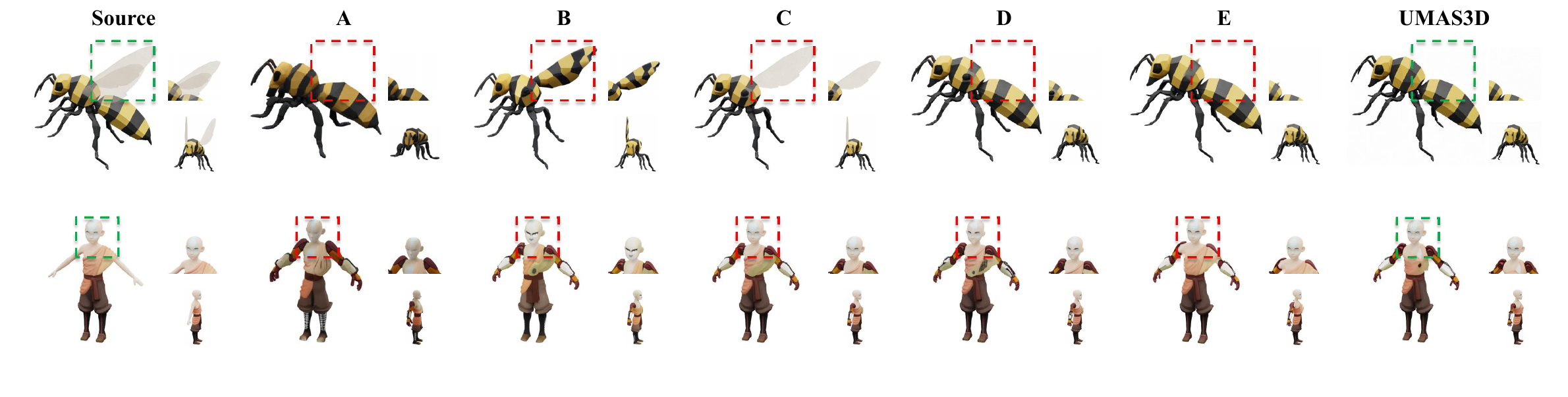}
\vspace{-1 em}
\caption{
\textbf{Qualitative ablation of UMAS3D.}
Variants follow Table~\ref{tab:ablation_studies}.
The first-row instruction is ``Remove both bee wings.'',
and the second-row instruction is ``Replace the human arm structure with robotic mechanical arms.''
}
\vspace{-.5 em}
\label{fig:ablation_experiment}
\end{figure}

As shown in Table~\ref{tab:ablation_studies}, we construct five ablation variants, \textbf{A--E}. \textbf{A} denotes the original TRELLIS; \textbf{B} introduces the structure-space supervision of \textbf{SGPS}, while \textbf{C} further introduces SLAT-space supervision. \textbf{D} additionally enables adaptive Edit/Keep adjustment. \textbf{E} replaces the external 3D masks with mask-free initialization while retaining the two-stage \textbf{SGPS}.

The ablation results show that introducing Stage~1 \textbf{SGPS} noticeably improves the preservation of unedited structures and overall geometric quality, indicating that source-consistency constraints in structure space effectively suppress unintended geometric changes during generation. Building upon this, adding Stage~2 \textbf{SGPS} further improves the preservation of appearance and local details in Keep regions, while also enhancing overall visual consistency. This demonstrates that \textbf{SGPS} provides complementary source-preservation capabilities in structure space and SLAT feature space.

Further introducing adaptive Edit/Keep adjustment improves both \emph{Unedited-region Preservation} and \emph{Overall Target Quality}, showing that dynamically adjusting Edit/Keep regions during generation provides more accurate control over the Edit/Keep areas than using a fixed initial partition, thereby reducing unintended modifications to unrelated source content. Meanwhile, replacing the external 3D mask with mask-free initialization still maintains strong editing quality and source-preservation capability, validating the effectiveness of \textbf{MARM} in constructing a reliable Edit/Keep prior without requiring an external 3D mask.

As shown in Figure~\ref{fig:ablation_experiment}, the qualitative results exhibit a trend consistent with the quantitative evaluation. With the progressive introduction of the two-stage \textbf{SGPS} and adaptive Edit/Keep adjustment, source structure and local appearance are preserved more consistently, while the editing regions are controlled more accurately. The full \textbf{UMAS3D} therefore achieves reliable target editing and source-content preservation without relying on an external 3D mask, further validating the effectiveness and complementary roles of \textbf{MARM} and \textbf{SGPS}.

\section{Conclusion}

This paper presents UMAS3D, a unified 3D editing framework that is training-free, inversion-free, and free from external 3D masks. UMAS3D unifies diverse multimodal editing instructions into target-view guidance and handles multiple editing tasks through a shared 3D editing pipeline. We introduce Mask-Free Adaptive Region Modeling (MARM) to initialize and progressively refine the Edit/Keep regions during generation, and Source-Guided Preservation Steering (SGPS) to preserve source-consistent geometry and appearance in the Keep regions while leaving the Edit regions target-driven. Experiments across diverse multimodal conditions and editing tasks demonstrate effective target editing together with strong preservation of unedited source content.

\textbf{Limitations.}
UMAS3D does not explicitly resolve conflicts among multimodal editing conditions. For example, if the text instruction asks to recolor a cup red while the reference image shows a blue cup, the framework lacks a dedicated mechanism to determine which condition should take priority.

\newpage

\subsection*{AI Use Statement}

Generative AI tools were used to assist with parts of the research and writing process. In the experimental pipeline, we used Gemini models to generate or supplement multimodal editing conditions, including bounding boxes, textual conditions, and reference images, and Gemini 3 Pro Image was further used to generate the target-view guidance employed by UMAS3D; the corresponding usage is described in the main paper and Appendix. Generative AI tools were also used to assist with code development and debugging, as well as language editing of the manuscript. All AI-assisted code was executed and checked for correctness, the generated multimodal conditions and target views were manually reviewed by the authors, and the manuscript text was carefully inspected and verified by the authors.

\subsection*{Reproducibility Statement}

To improve the reproducibility of our work, the main paper presents the complete UMAS3D pipeline together with the core methodology and mathematical formulations of MARM and SGPS, and details the datasets, baselines, and evaluation metrics used in our experiments. The Appendix further provides the construction of multimodal data, implementation settings, key hyperparameters, and additional evaluation details that are omitted from the main text due to space limitations. Together, these methodological and experimental specifications provide the key information needed to reproduce the main results reported in this work.

\bibliography{conference}
\bibliographystyle{conference}

\appendix
\section{Appendix}

\subsection{Dataset Construction}

Our quantitative benchmark is constructed from Pxform and Omni3DEdit and contains 150 source 3D assets spanning five editing tasks: Addition, Deletion, Modification, Appearance, and Pose, with 32, 29, 30, 29, and 30 samples, respectively. In addition, a small number of 3D assets from Objaverse are used for qualitative demonstrations on the project page and in the supplementary material to examine the applicability of UMAS3D to a broader range of object categories. These Objaverse samples are used only for additional qualitative visualization and are excluded from the quantitative benchmark, metric computation, and experimental statistics.

\begin{table}[H]
\centering
\footnotesize
\setlength{\tabcolsep}{2.5pt}
\renewcommand{\arraystretch}{1.06}

\caption{
Construction of the five multimodal editing conditions used in our benchmark.
Bracketed terms denote sample-specific content.
}
\label{tab:dataset_prompt_construction}

\begin{tabular}{
@{}p{0.09\textwidth}
p{0.25\textwidth}
p{0.31\textwidth}
p{0.30\textwidth}@{}
}
\toprule

\textbf{Input} &
\textbf{Text} &
\textbf{Bounding Box} &
\textbf{Reference Image}
\\
\midrule

$T$
&
\emph{[Original Instruction]}
\par\smallskip
The original editing instruction is retained verbatim without rewriting.
&
--
&
--
\\
\midrule

$T+B$
&
\emph{[Modality-Specific Text]}
\par\smallskip
Since the Bounding Box provides \textbf{where} the edit occurs,
redundant spatial descriptions already conveyed by the box are removed,
while preserving the editing operation and target content or appearance.
The original editing intent remains unchanged.
&
\emph{[Four Source Views]} +
\emph{[Original Instruction]}
\par\smallskip
Select the source view that best exposes the edit-relevant region,
identify the \textbf{source-side edit anchor}, and return a tight
2D box $[x_1,y_1,x_2,y_2]$ normalized to $[0,1000]$.
For Addition, the box localizes the source-side attachment or placement
region rather than future target content absent from the source.
&
--
\\
\midrule

$T+R$
&
\emph{[Modality-Specific Text]}
\par\smallskip
Since the Reference Image provides \textbf{what / appearance},
the text retains the editing operation, spatial location, and necessary
relation while removing target visual details already represented
by the reference.
&
--
&
\emph{[Selected Source View]} +
\emph{[Original Instruction]}
\par\smallskip
Generate a standalone visual reference that represents only the target
visual information required by the edit.
For Addition, the reference depicts the object to be added;
for Modification, the desired replacement or modified part;
and for Appearance, the target color, material, texture, or pattern.
\\
\midrule

$B+R$
&
--
&
Uses the same source-side Bounding Box as $T+B$.
&
Uses the same Reference Image as $T+R$.
\\
\midrule

$T+B+R$
&
\emph{[Modality-Specific Text]}
\par\smallskip
The Bounding Box provides \textbf{where}, while the Reference Image
provides \textbf{what / appearance}.
The text therefore retains only the editing operation and any relation
still required to specify the intended edit.
&
Uses the same source-side Bounding Box as $T+B$.
&
Uses the same Reference Image as $T+R$.
\\

\bottomrule
\end{tabular}
\end{table}

Since the original datasets do not jointly provide complete Text, Bounding Box, and Reference Image conditions, we construct the missing modalities while preserving the original editing intent. For each sample, the source 3D asset is first rendered from four canonical views, i.e., front, right, back, and left, and the source view that best exposes the region relevant to the requested edit is selected according to the original instruction. The selection considers the visibility of the target part, occlusion, and its spatial relationship with surrounding structures. Based on the selected source view and the original instruction, we then construct the corresponding Bounding Box, Reference Image, and text variants required by different multimodal settings. These components are finally organized into five unified input combinations: $T$, $T+B$, $T+R$, $B+R$, and $T+B+R$.

\begin{figure}[t]
\centering

\setlength{\tabcolsep}{2.5pt}
\renewcommand{\arraystretch}{1.0}

\newcommand{\conditionexample}[3]{%
\begin{tikzpicture}[baseline=(current bounding box.center)]
\node[
    draw=black!70,
    dashed,
    line width=0.55pt,
    rounded corners=5pt,
    inner xsep=4pt,
    inner ysep=4pt
] {%
    \begin{minipage}{0.295\textwidth}
        \centering

        \begin{minipage}[c]{0.485\linewidth}
            \centering
            \includegraphics[
                width=\linewidth,
                height=1.65cm,
                keepaspectratio
            ]{#1}
        \end{minipage}%
        \hfill
        \begin{minipage}[c]{0.485\linewidth}
            \centering
            \includegraphics[
                width=\linewidth,
                height=1.65cm,
                keepaspectratio
            ]{#2}
        \end{minipage}

        \vspace{1.4mm}

        \parbox{0.94\linewidth}{%
            \centering
            \scriptsize
            \textbf{Text:} #3
        }

    \end{minipage}%
};
\end{tikzpicture}%
}

\newcommand{\conditionheader}{%
\begin{minipage}{0.295\textwidth}
    \centering
    \scriptsize

    \begin{minipage}{0.485\linewidth}
        \centering
        \textbf{Bounding Box}
    \end{minipage}%
    \hfill
    \begin{minipage}{0.485\linewidth}
        \centering
        \textbf{Reference Image}
    \end{minipage}
\end{minipage}%
}

\begin{tabular}{@{}ccc@{}}

\conditionheader
&
\conditionheader
&
\conditionheader
\\[1.5mm]

\conditionexample
    {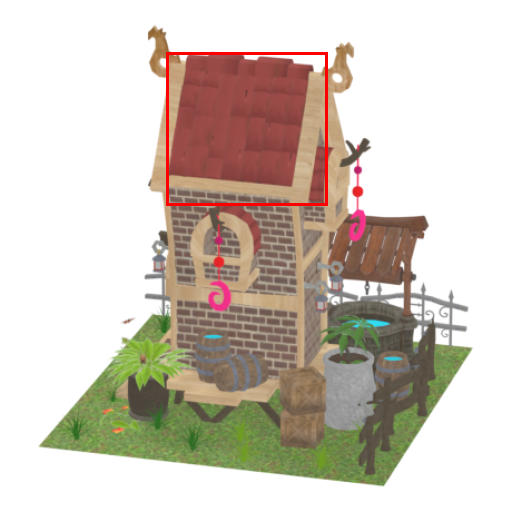}
    {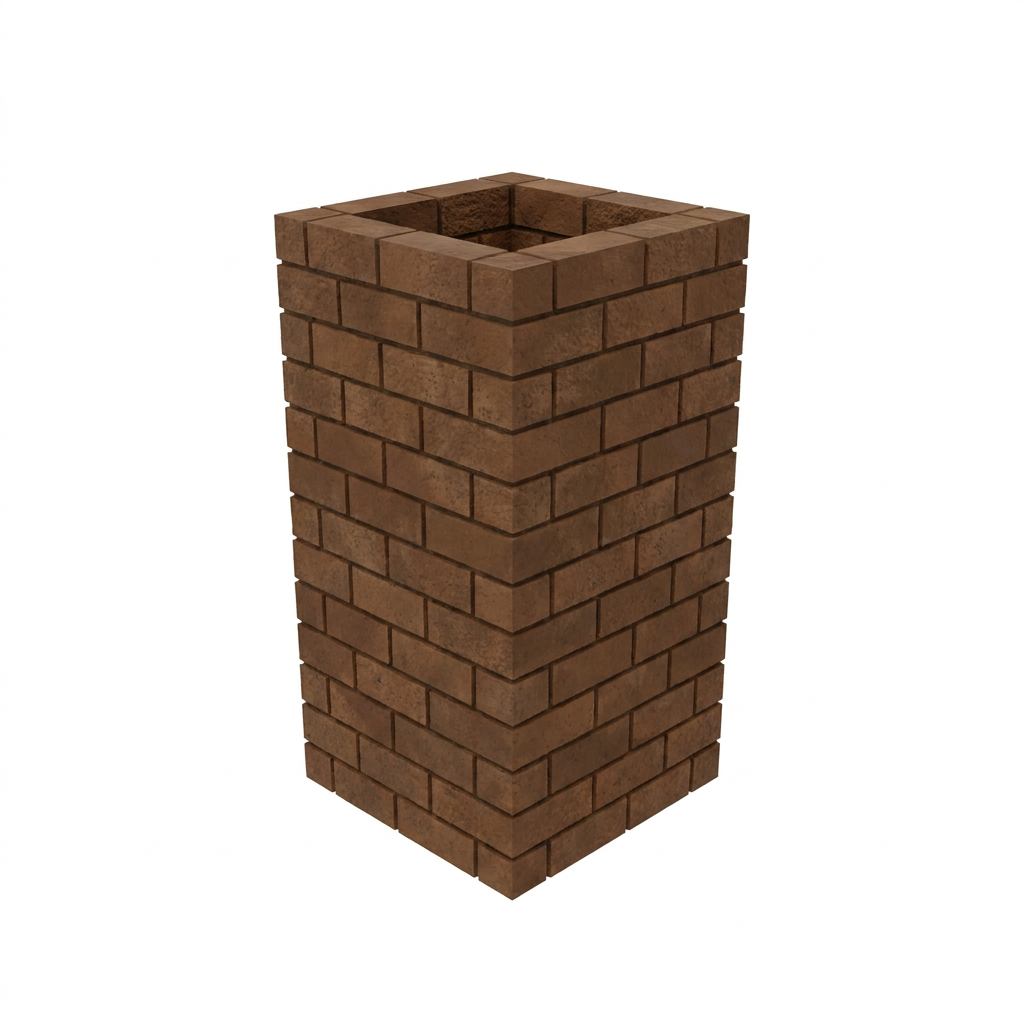}
    {Add a chimney to the roof.}
&
\conditionexample
    {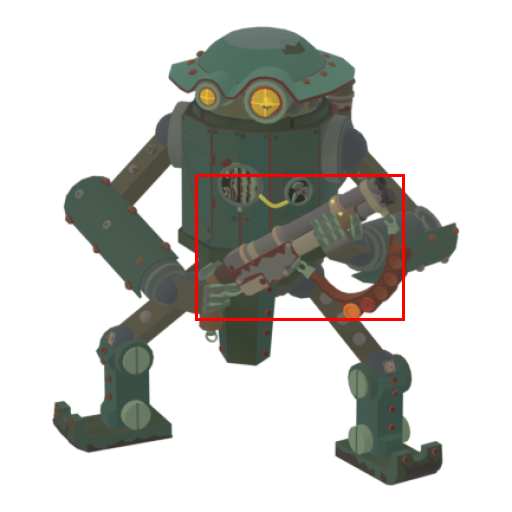}
    {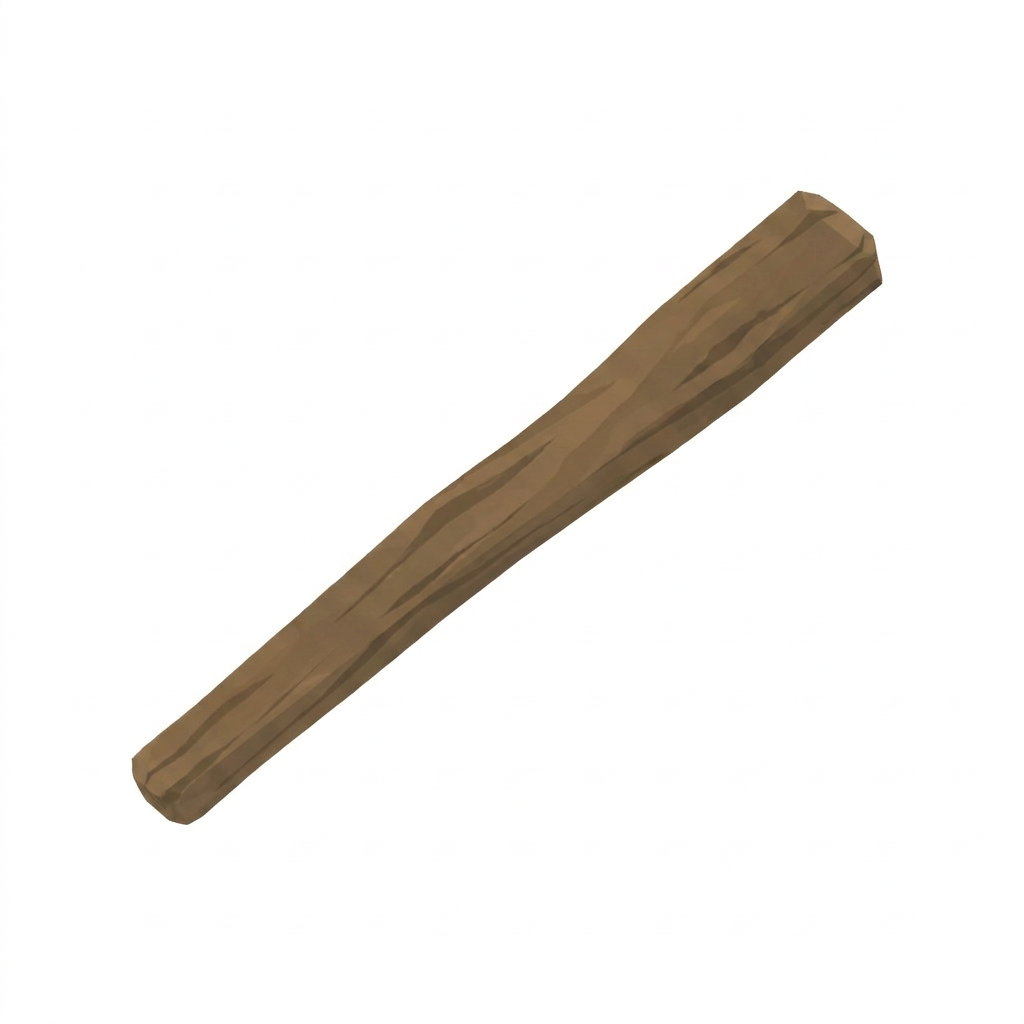}
    {Replace the pistol with a stick.}
&
\conditionexample
    {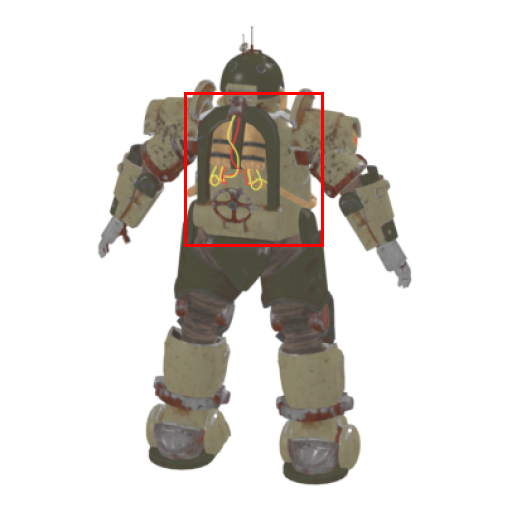}
    {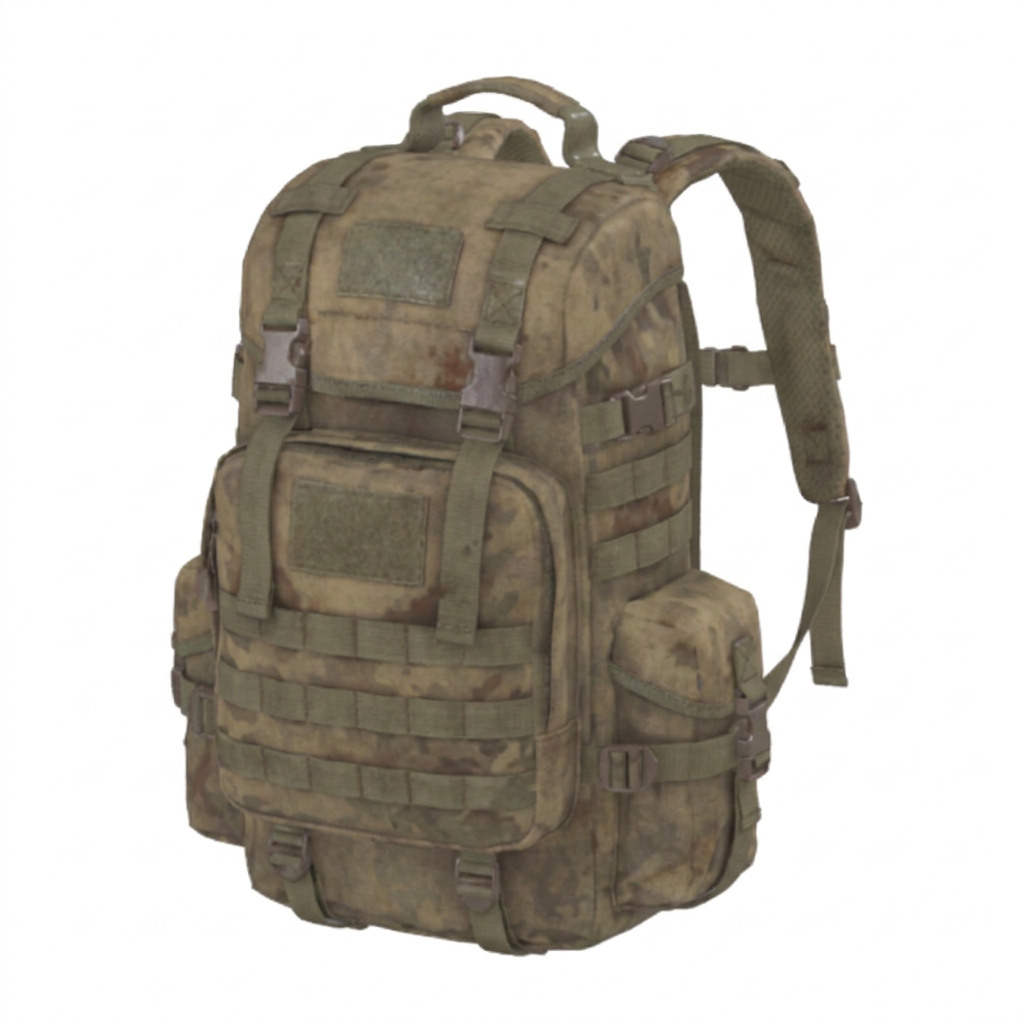}
    {Add a backpack.}
\\[12mm]

\conditionexample
    {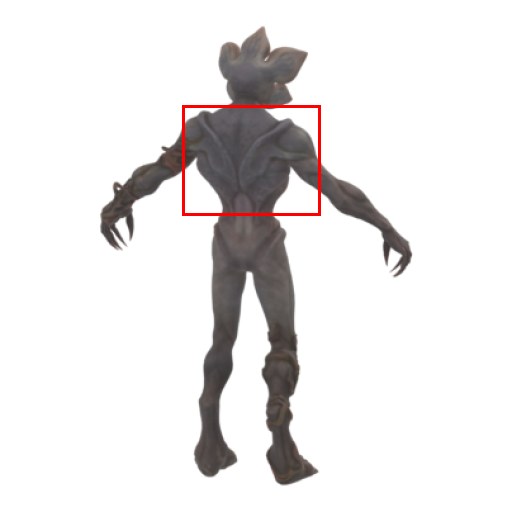}
    {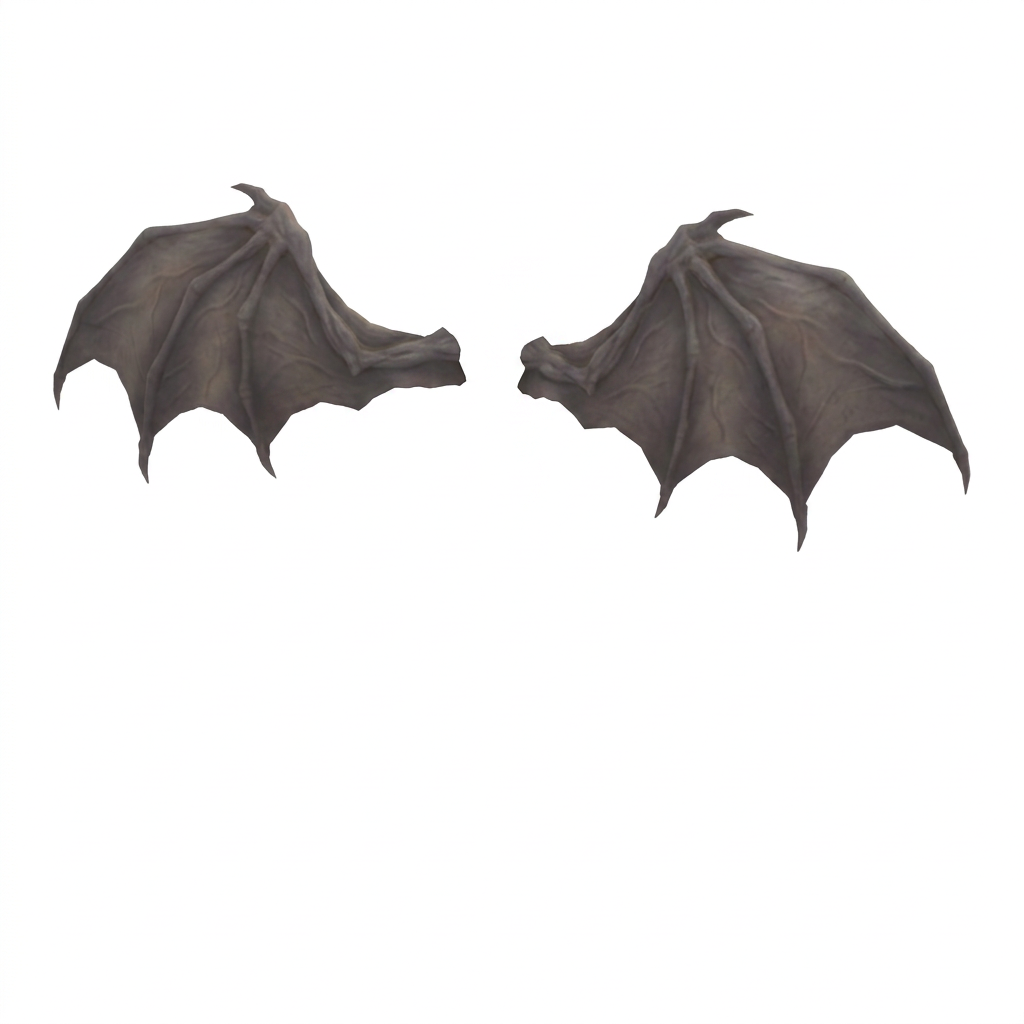}
    {Add a pair of wings.}
&
\conditionexample
    {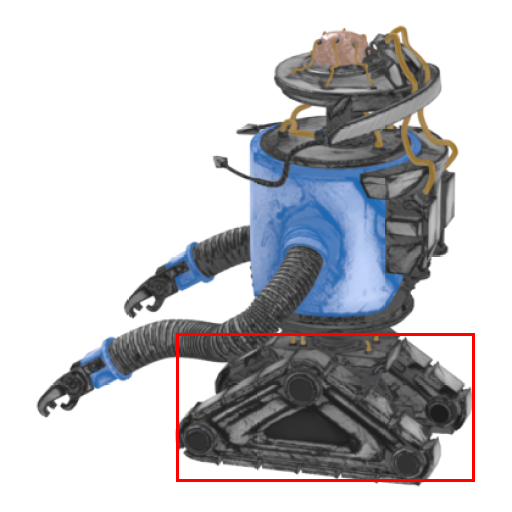}
    {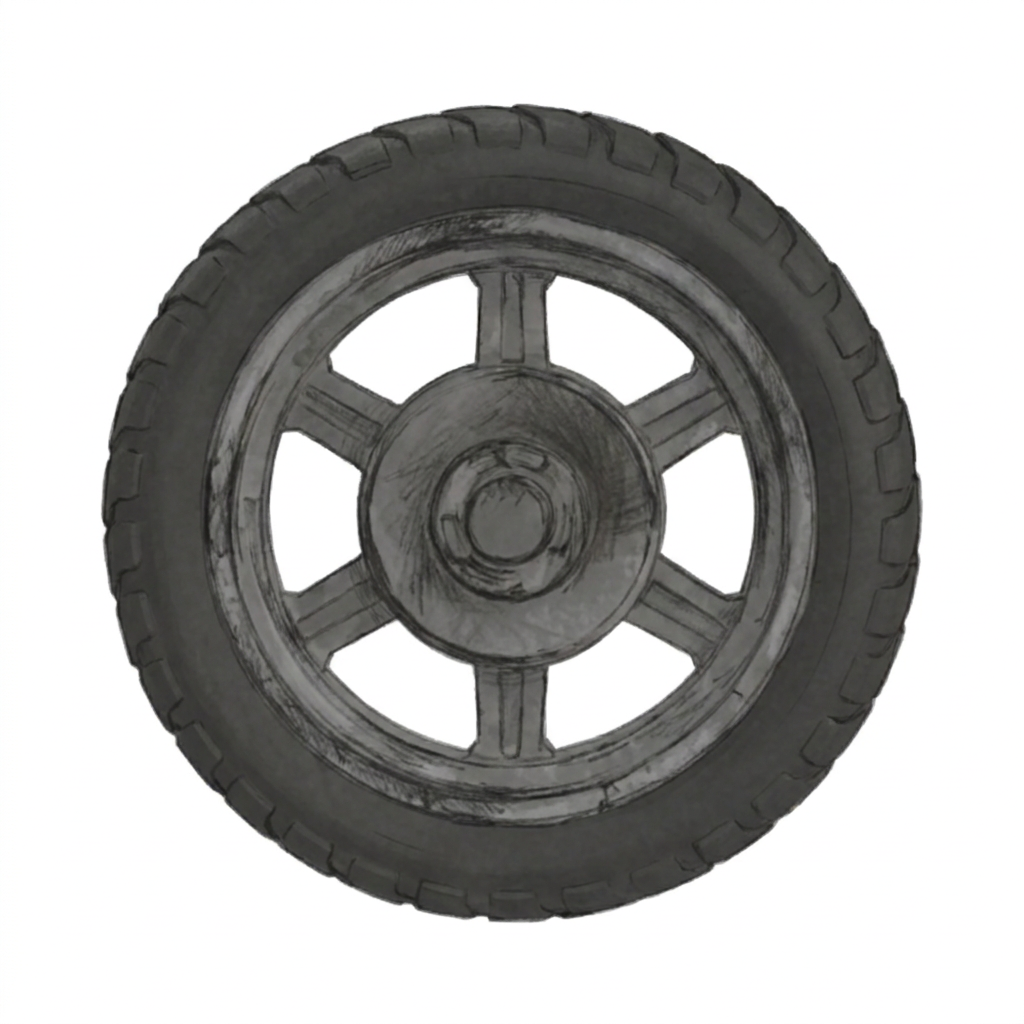}
    {Replace with a round wheel.}
&
\conditionexample
    {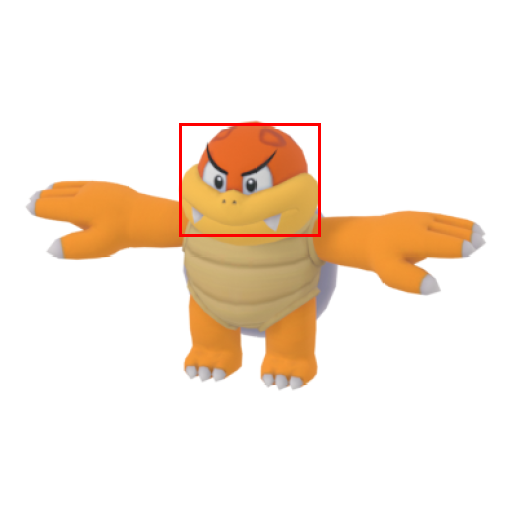}
    {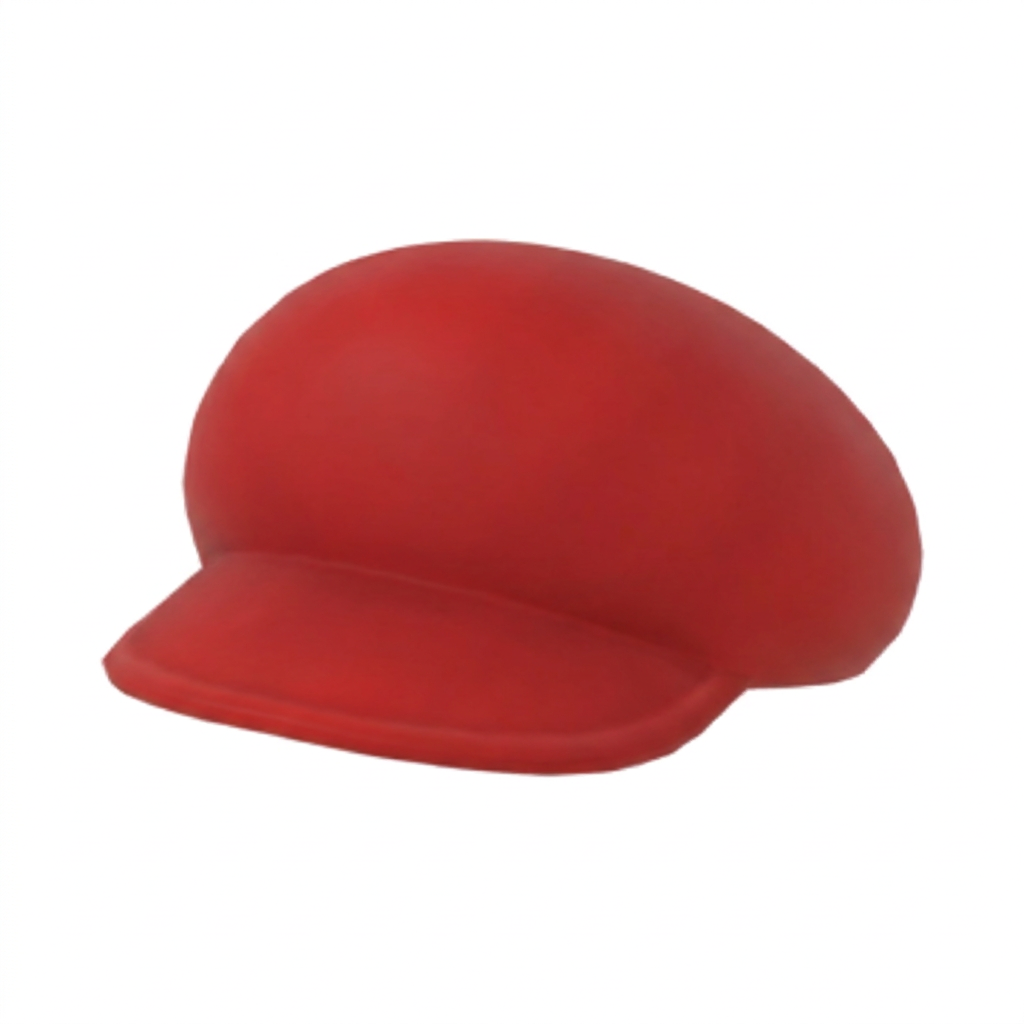}
    {Wear a red hat.}
\\[12mm]

\conditionexample
    {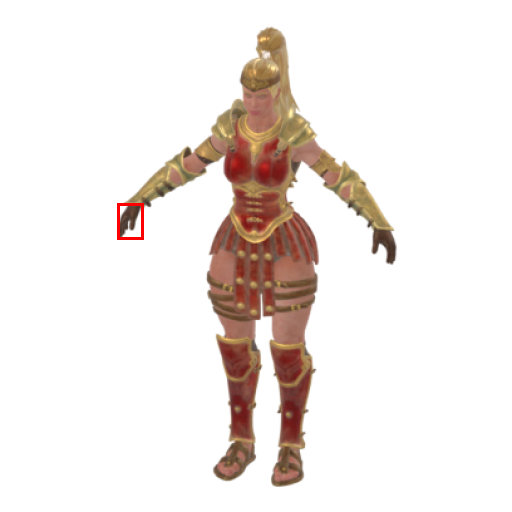}
    {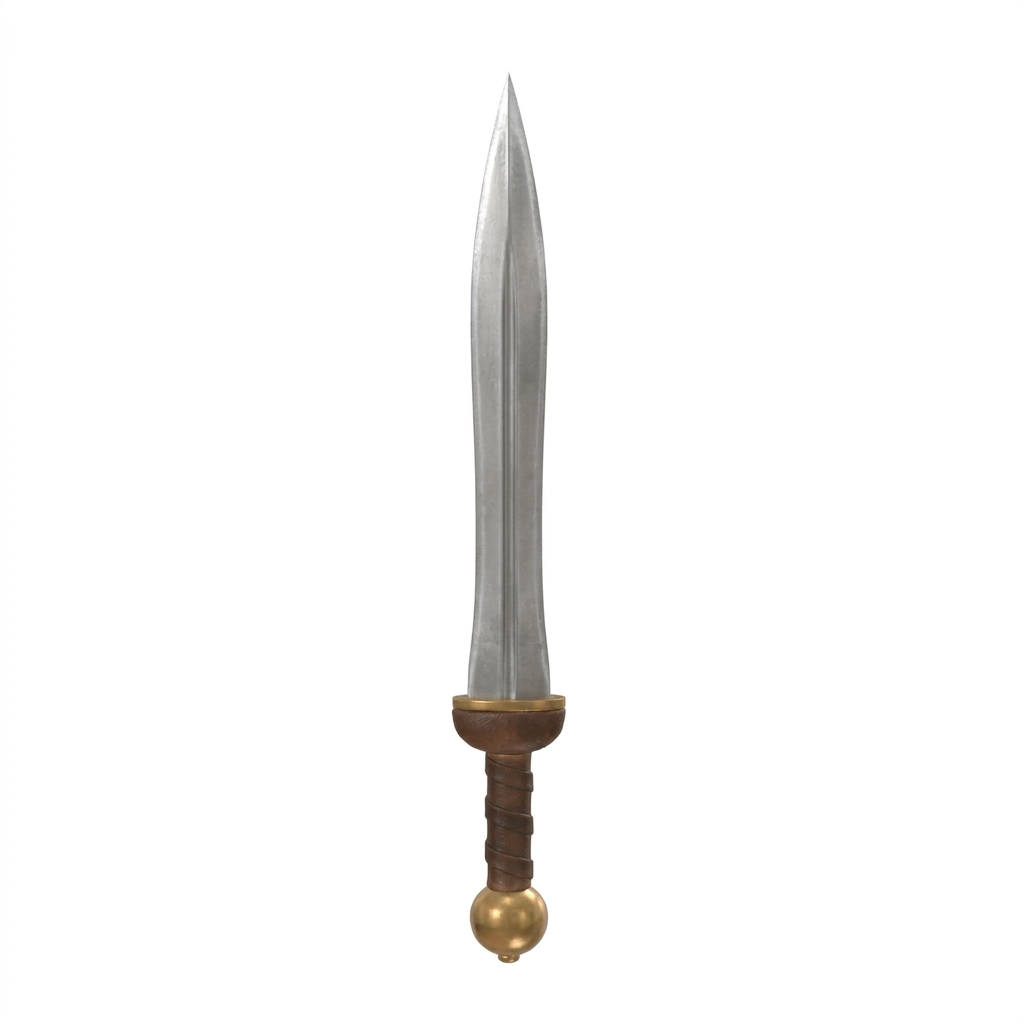}
    {Hold a sword \\ in one's hand.}
&
\conditionexample
    {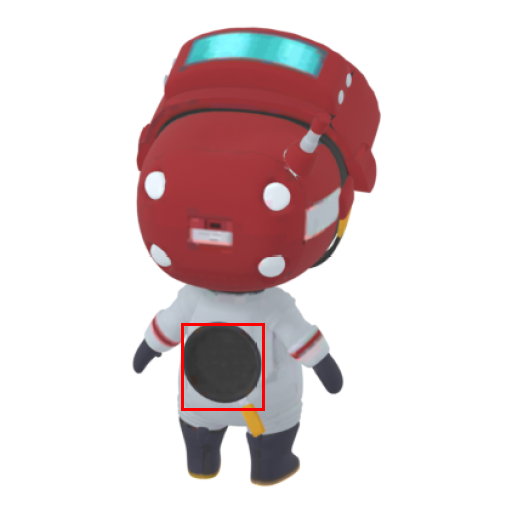}
    {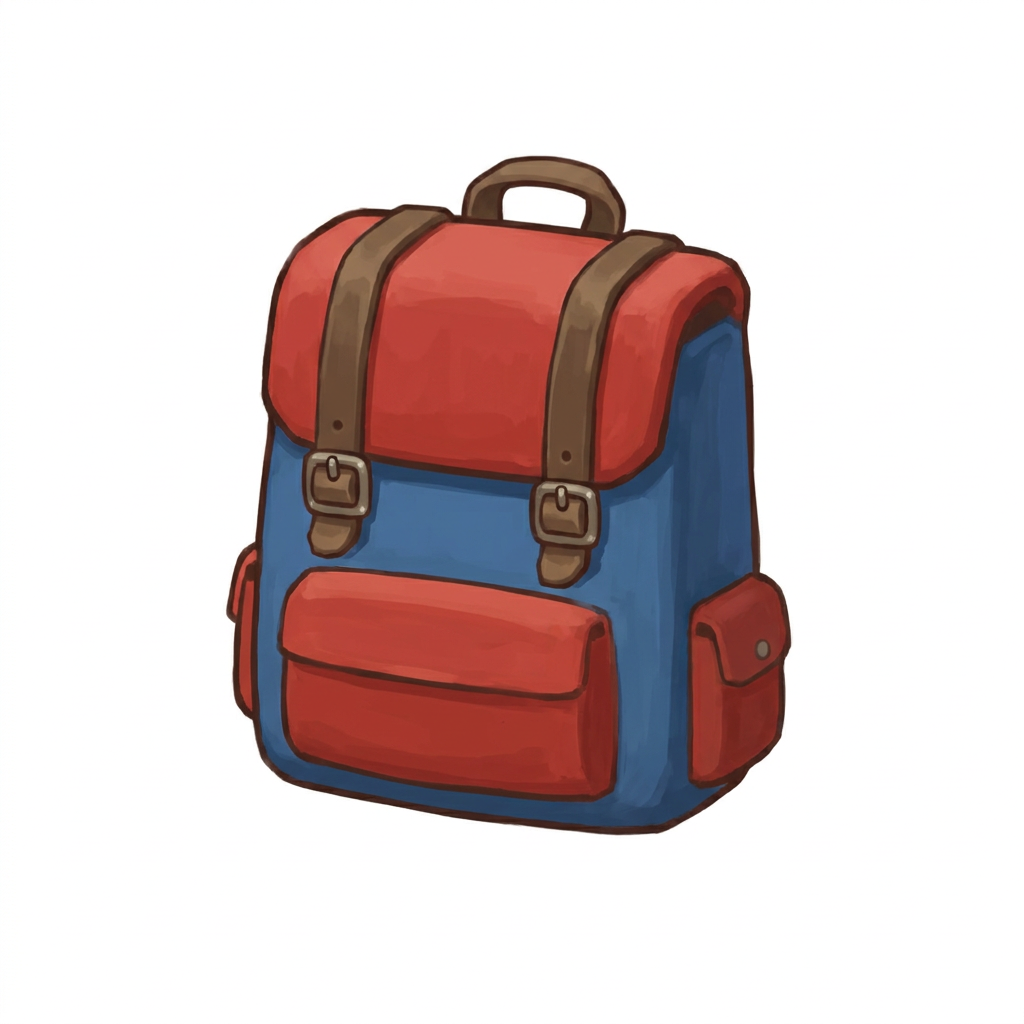}
    {Replace the heavy backpack with a small schoolbag on the back.}
&
\conditionexample
    {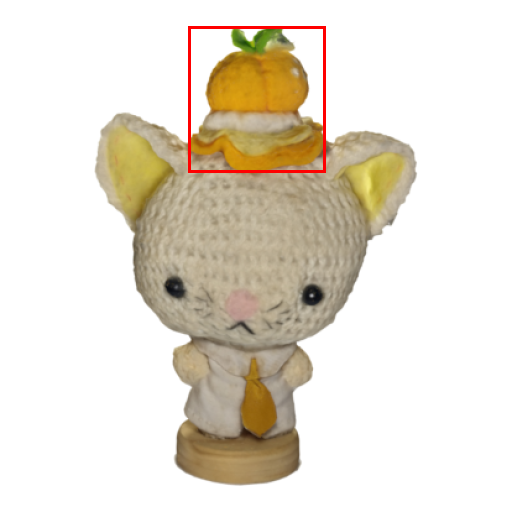}
    {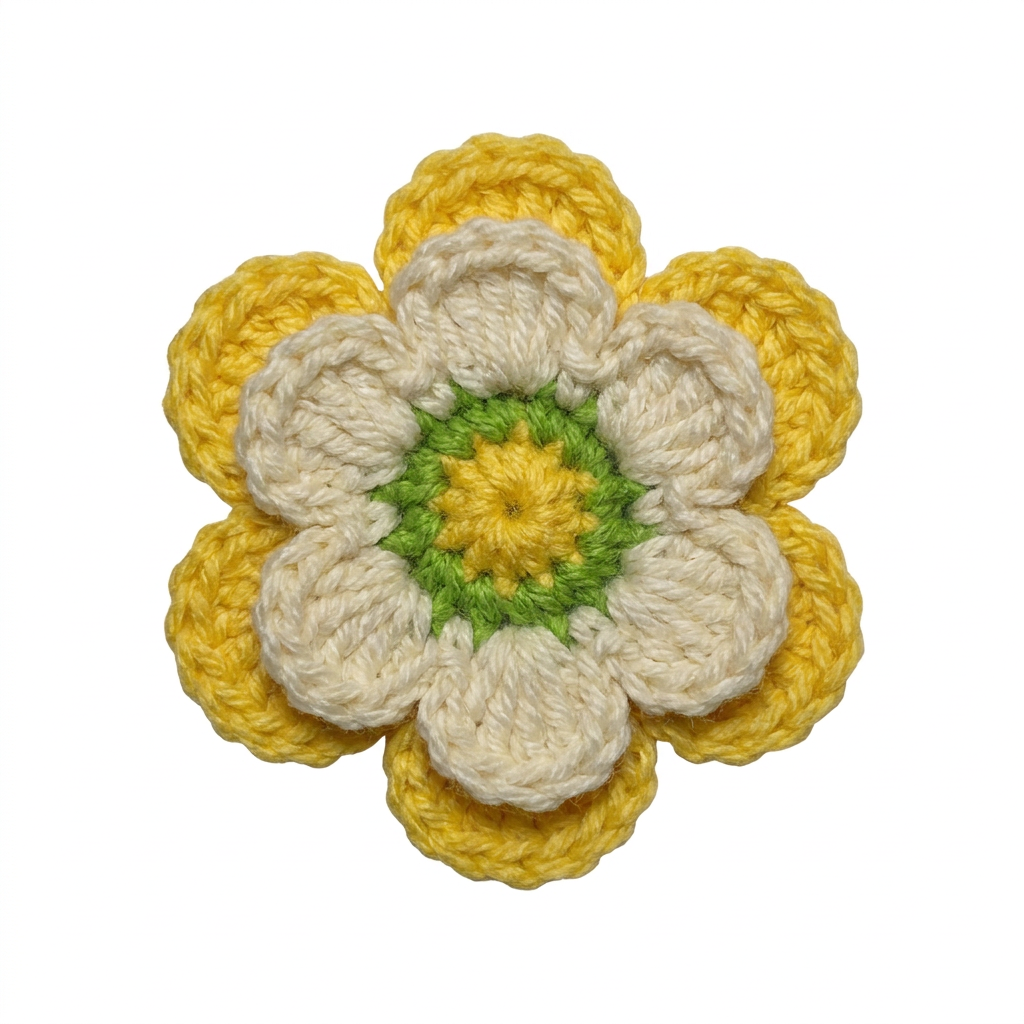}
    {Replace the orange on your head with a flower.}
\\[13mm]

\conditionexample
    {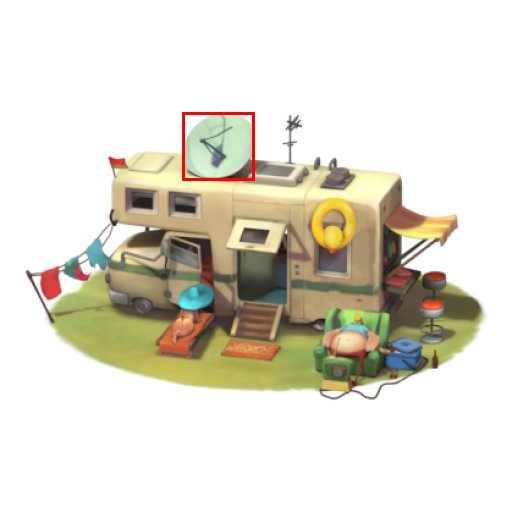}
    {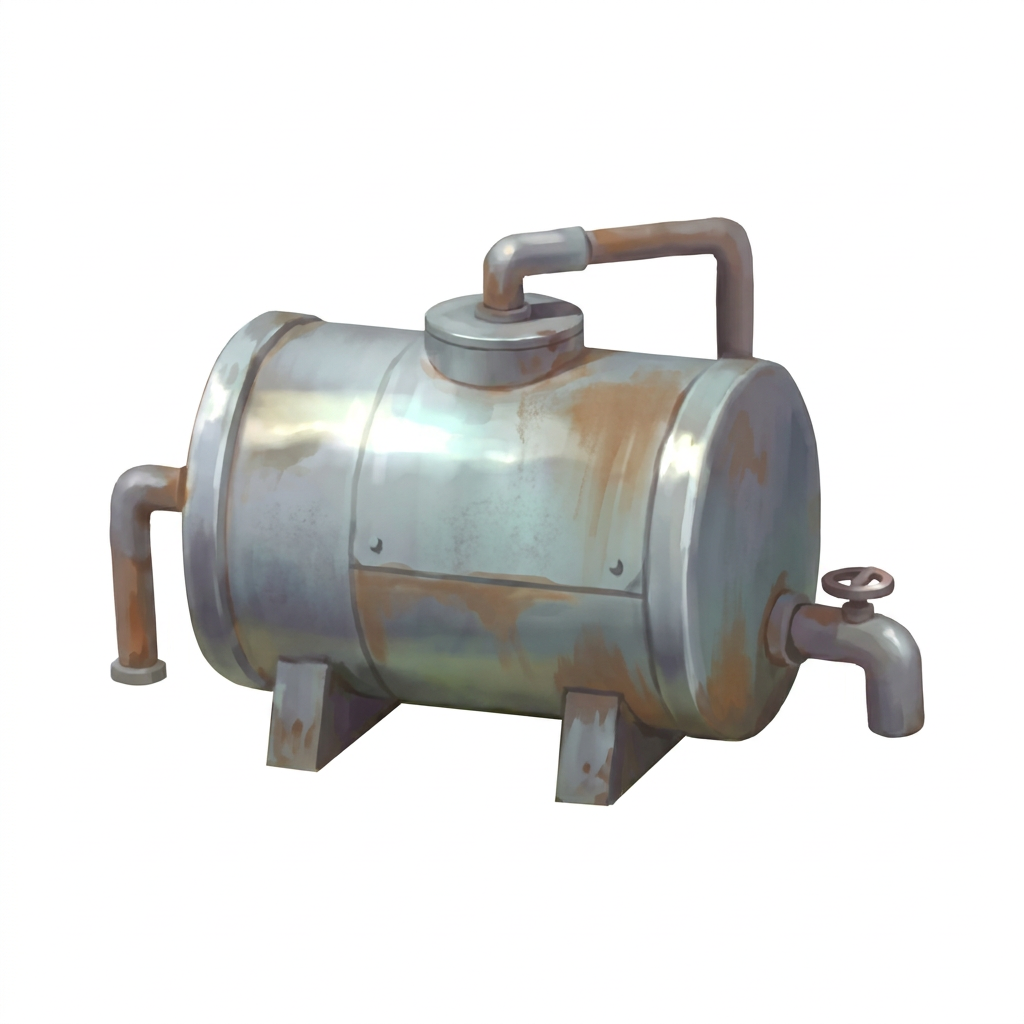}
    {Replace the remote sensing device on the roof with a water tank.}
&
\conditionexample
    {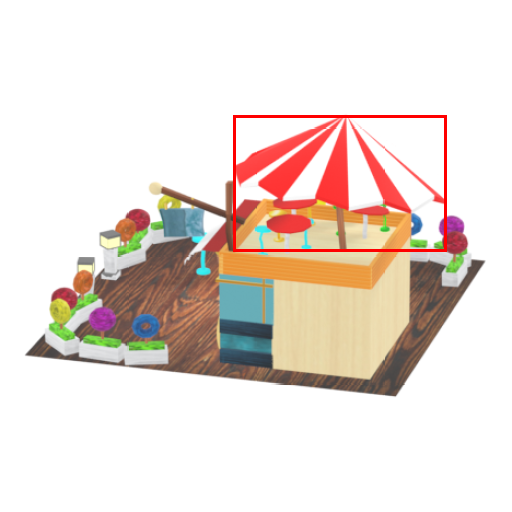}
    {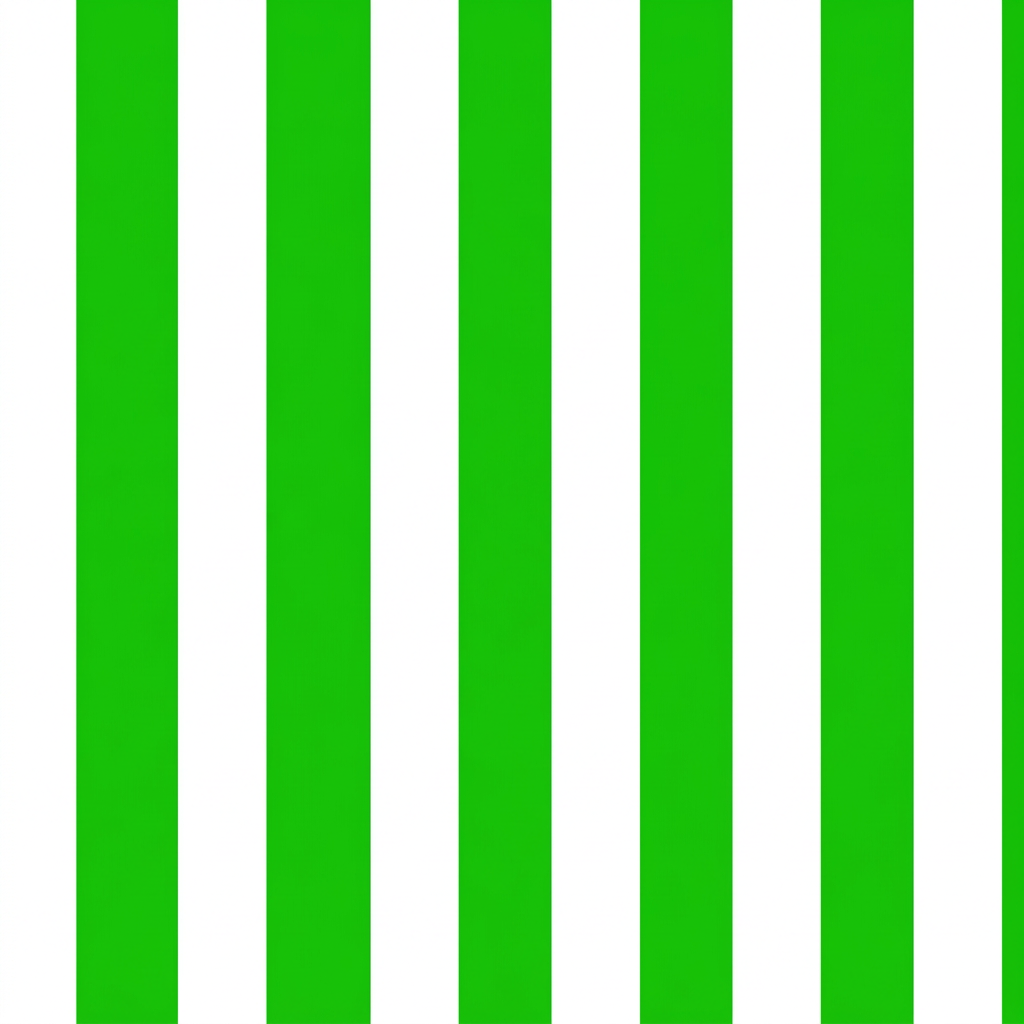}
    {Change the umbrella to a green and white color combination.}
&
\conditionexample
    {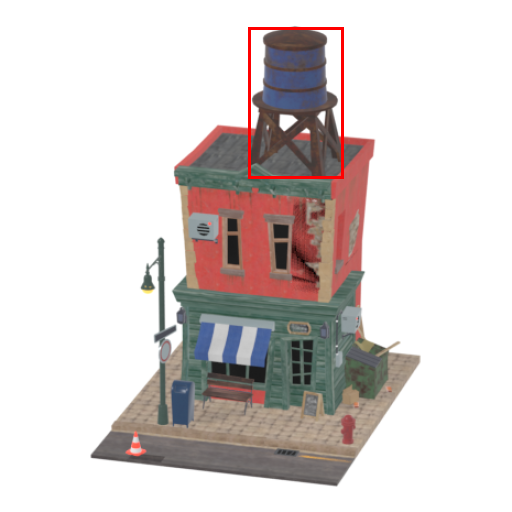}
    {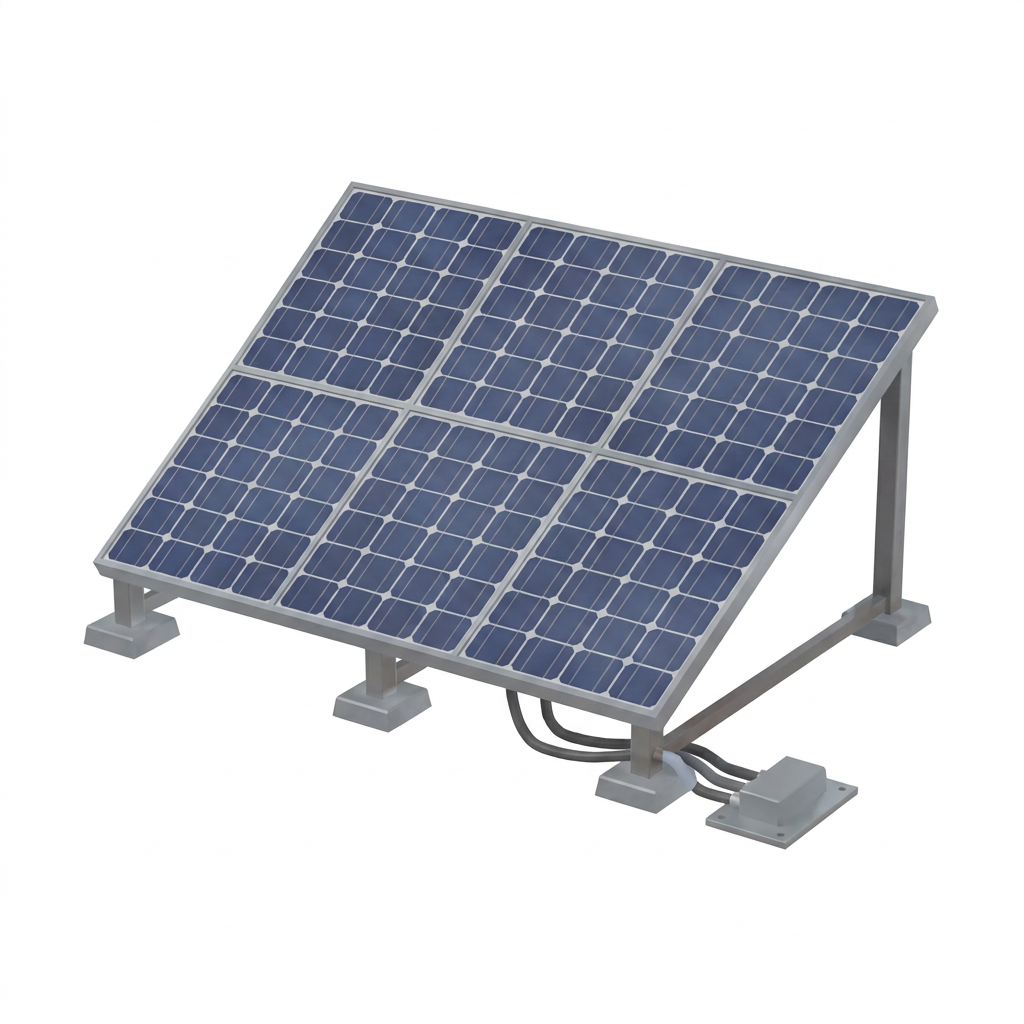}
    {Replace the water bucket on the roof with a solar photovoltaic panel.}
\\[13mm]

\conditionexample
    {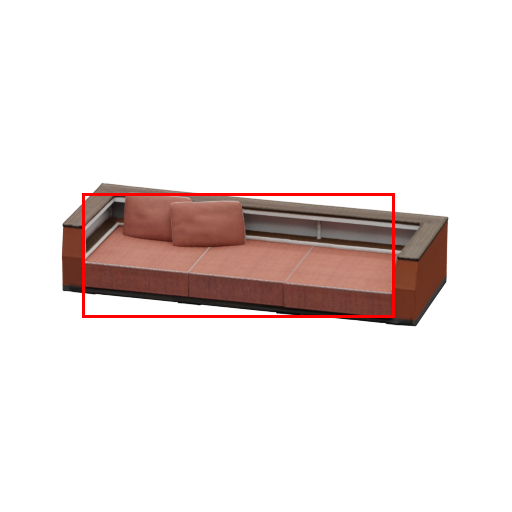}
    {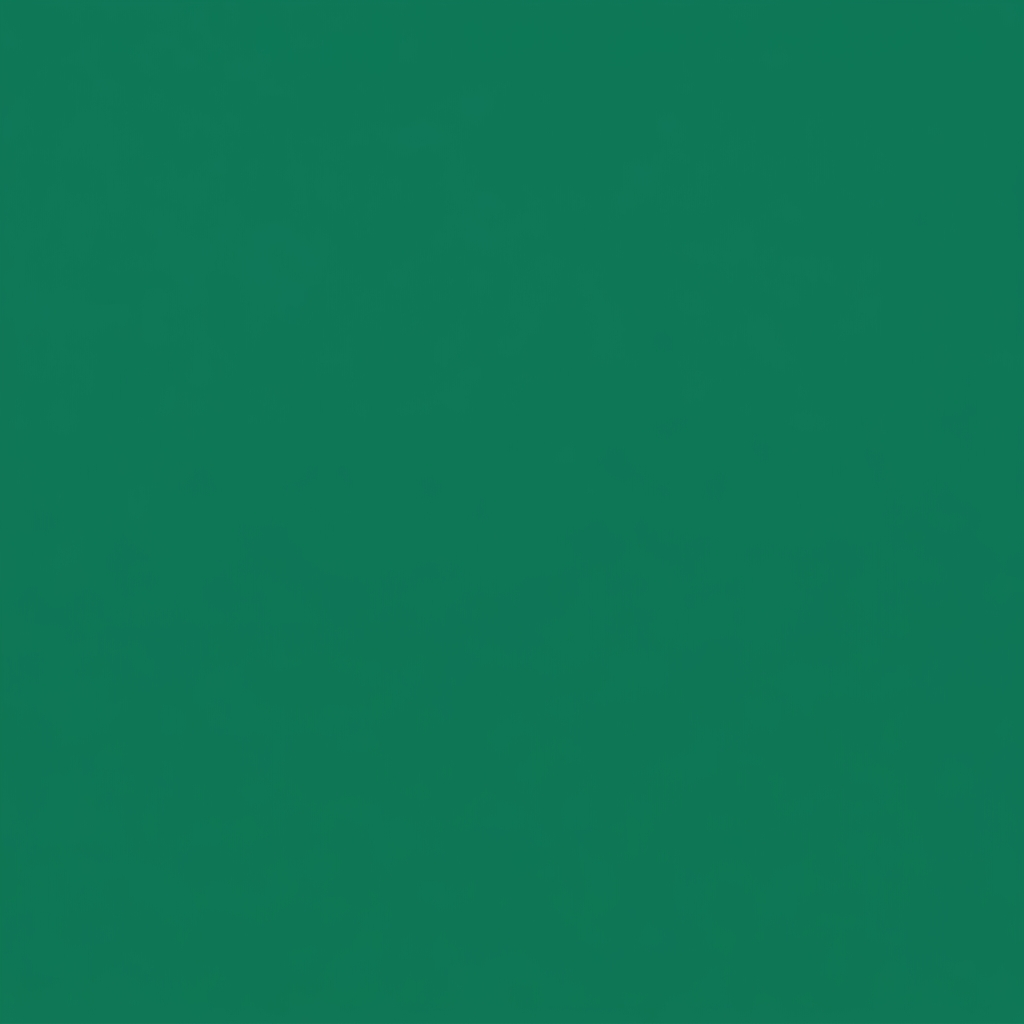}
    {Repaint all red cushions to emerald green.}
&
\conditionexample
    {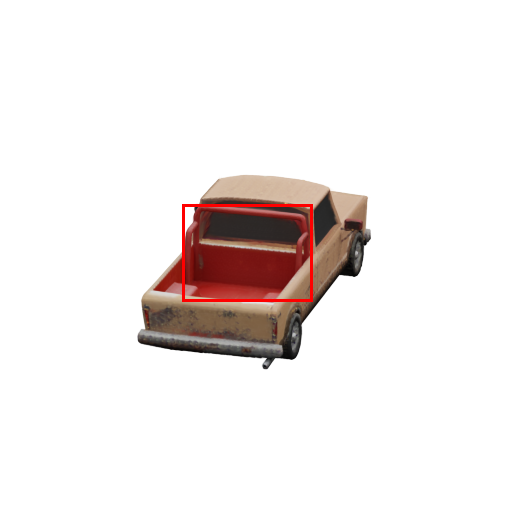}
    {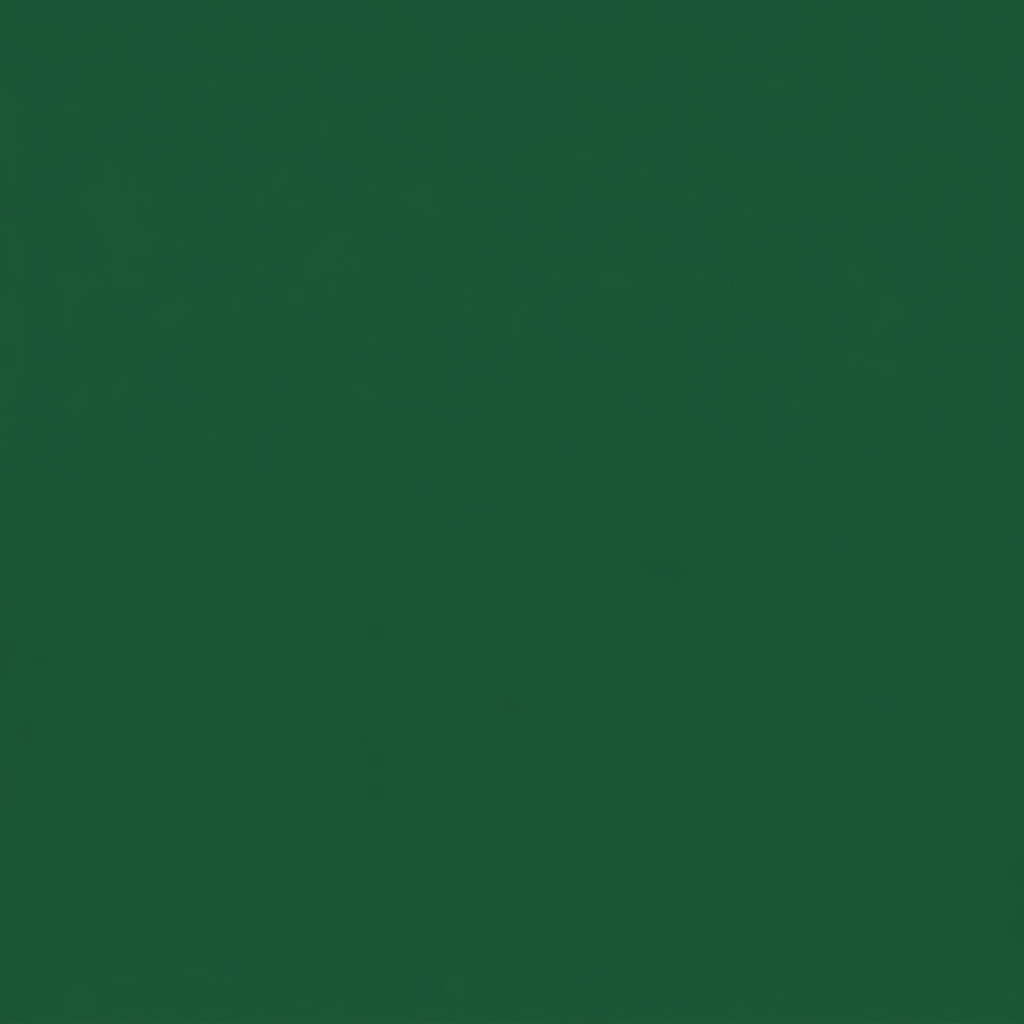}
    {Repaint the truck bed and rail in deep forest green.}
&
\conditionexample
    {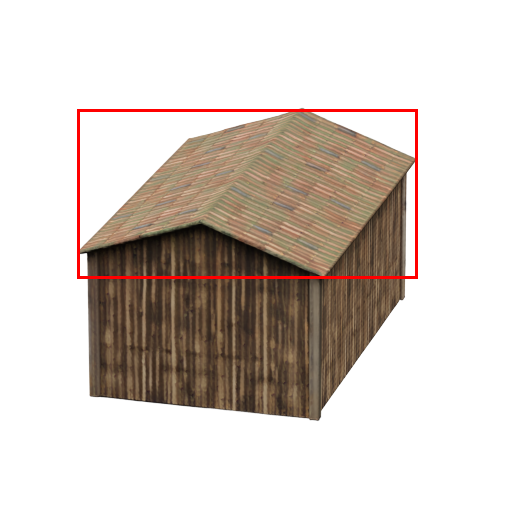}
    {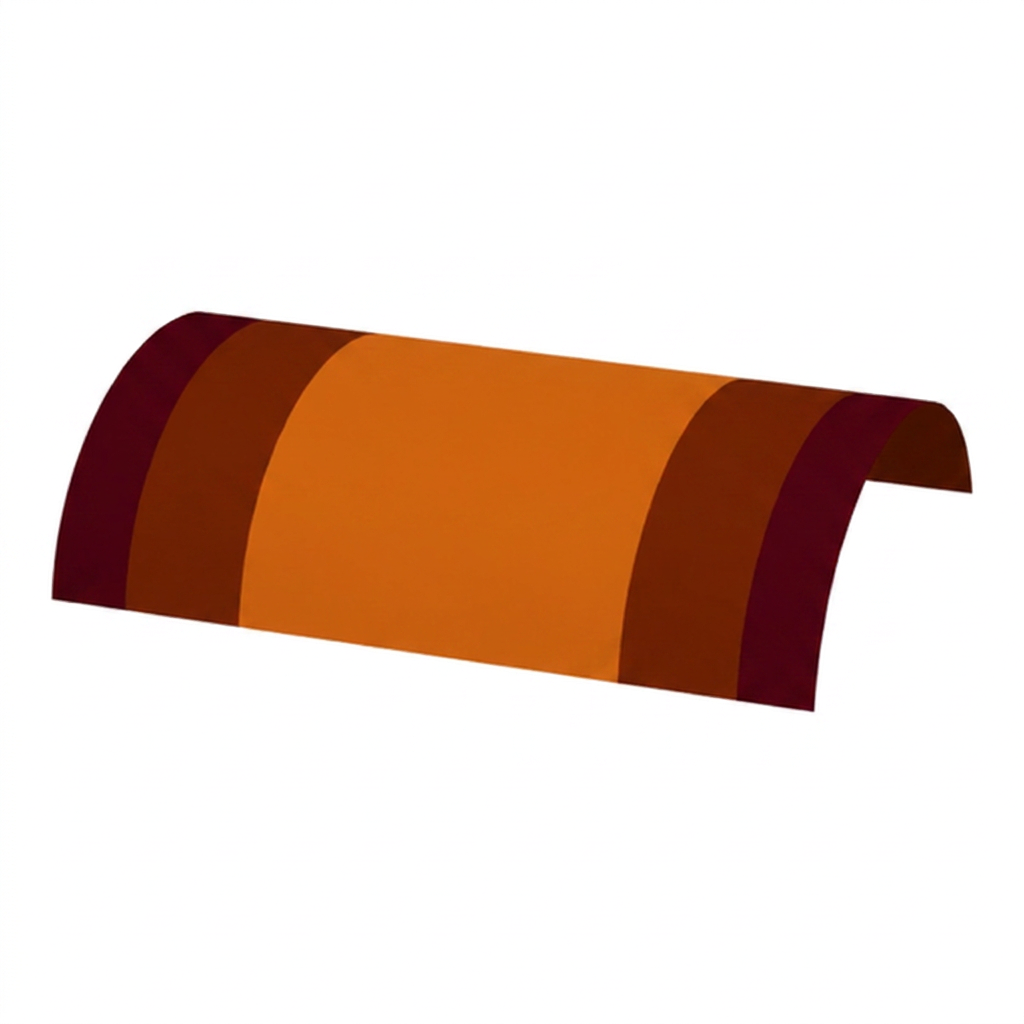}
    {Replace the wooden roof with a curved metal dome.}
\\[13mm]

\conditionexample
    {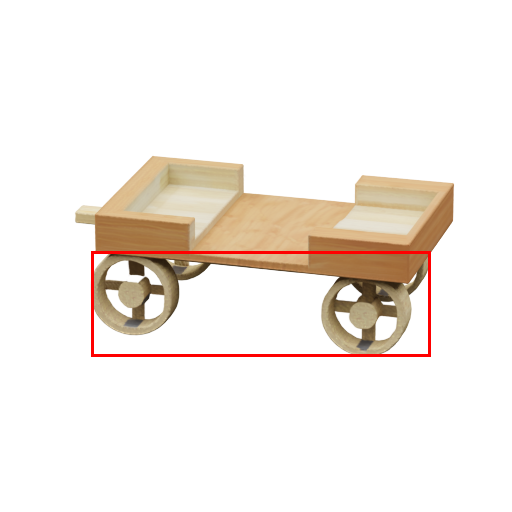}
    {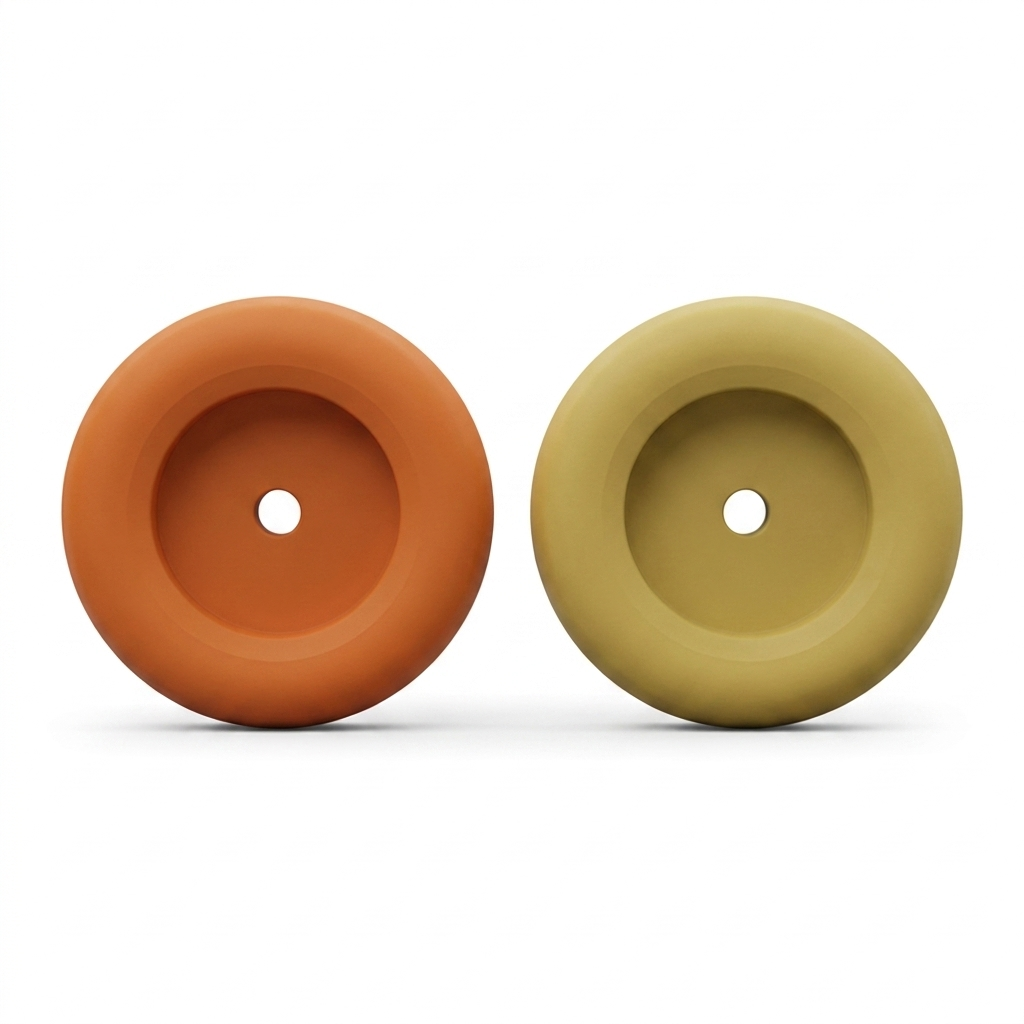}
    {Replace the wooden wheels with solid rubber tires.}
&
\conditionexample
    {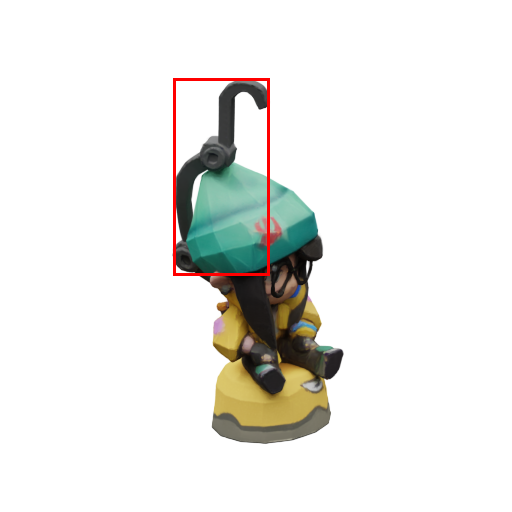}
    {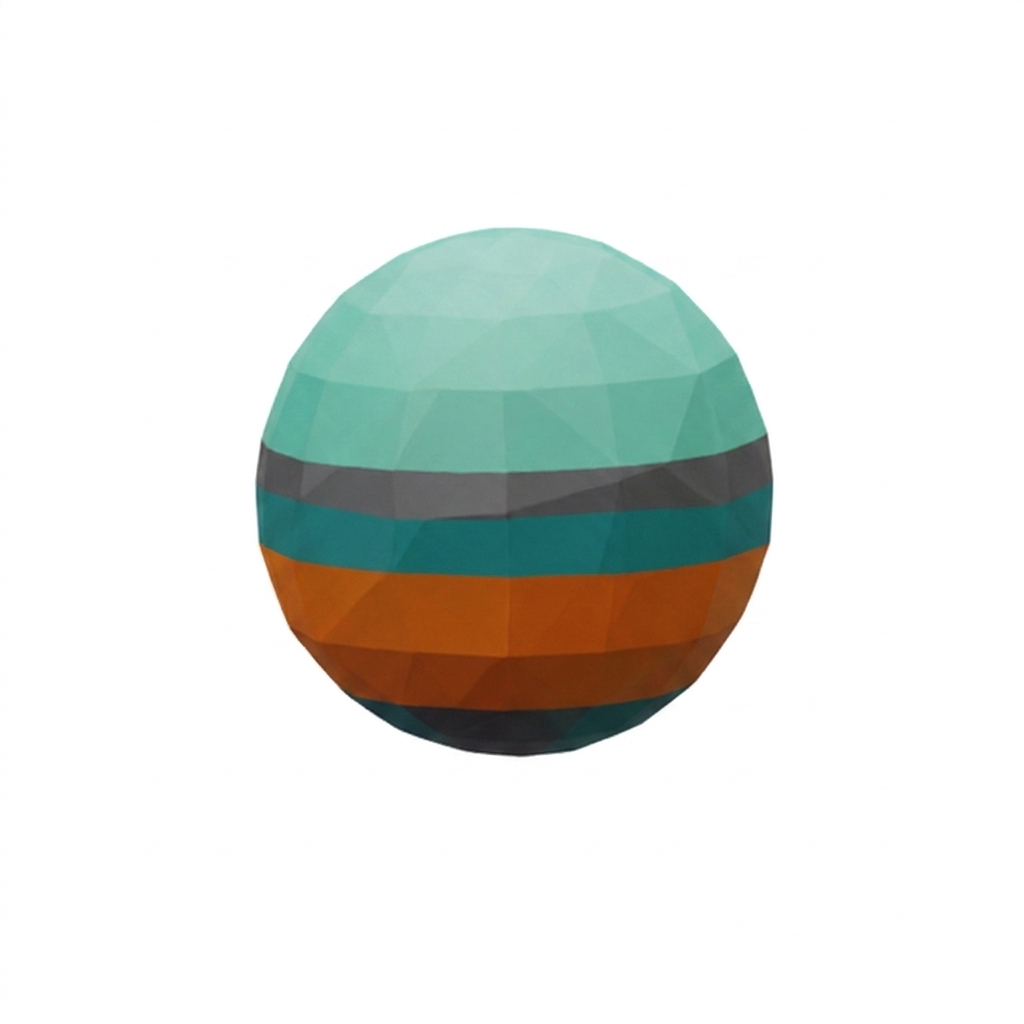}
    {Replace the mechanical hook with a spherical manipulator ball.}
&
\conditionexample
    {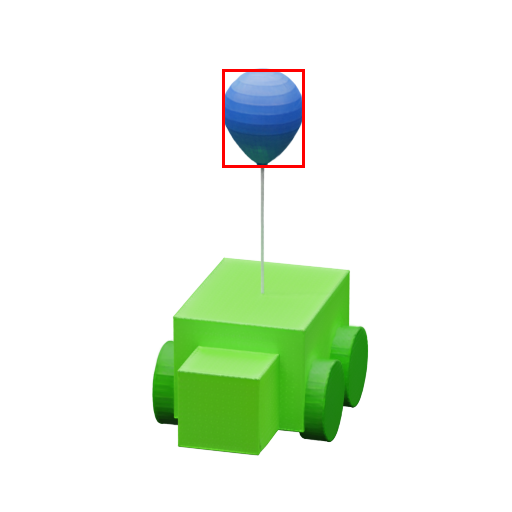}
    {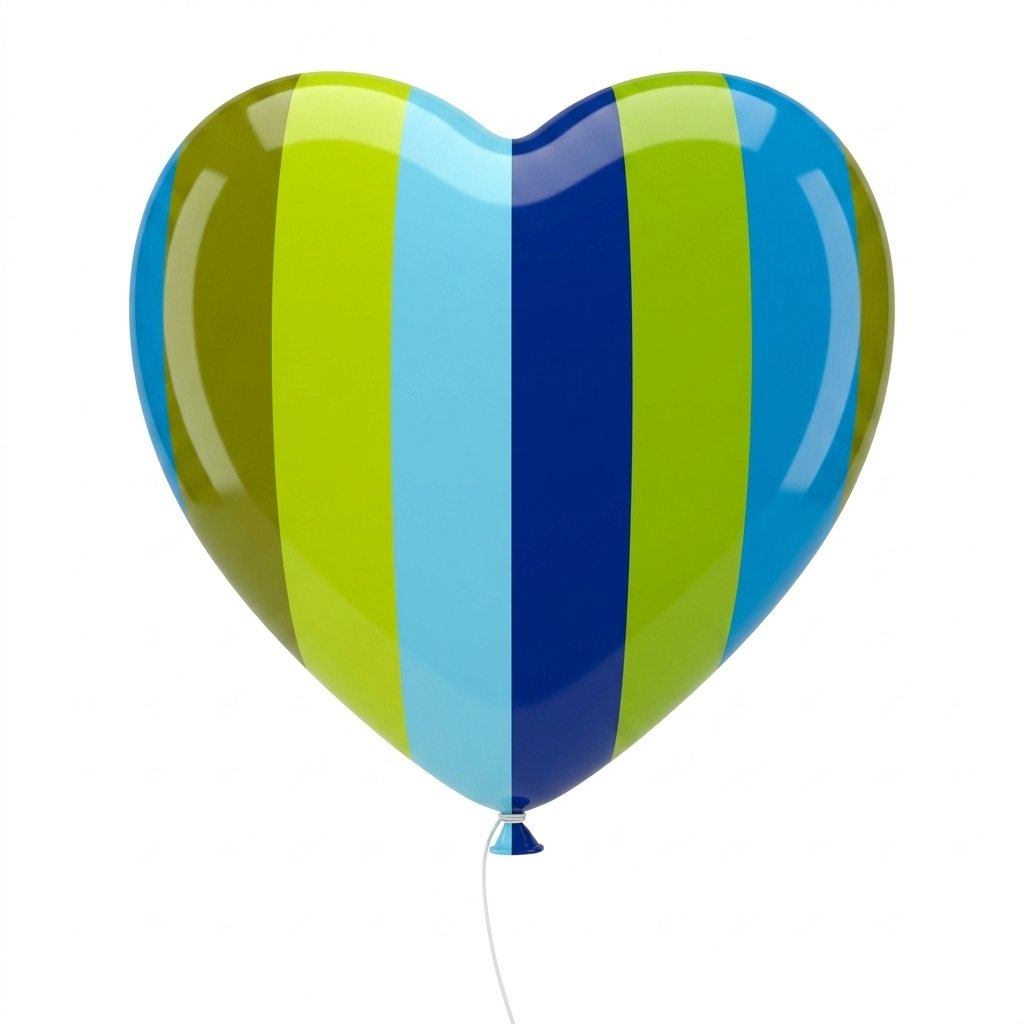}
    {Replace the blue balloon with a red heart-shaped balloon.}

\end{tabular}

\caption{Examples of the constructed multimodal editing conditions. Each dashed box shows the Bounding Box visualization, the generated Reference Image, and the original textual editing instruction. Modality-specific text variants are constructed according to Table~\ref{tab:dataset_prompt_construction}.}
\label{fig:A1}

\end{figure}

The Text-only condition always retains the original editing instruction verbatim. For Bounding-Box-based conditions, Gemini 3.6 Flash analyzes the four source views together with the original editing semantics, selects the most suitable source view, and localizes the source-side edit anchor within that view to construct a normalized 2D box. The box therefore represents an edit-relevant spatial anchor already present in the source rather than future target content. For reference-based conditions, Gemini 3 Pro Image generates a standalone visual reference from the selected source view and the original editing instruction. The Reference Image represents the object to be added for Addition, the desired replacement or modified part for Modification, and the target color, material, texture, or pattern for Appearance. Reference-based conditions are used only for Addition, Modification, and Appearance. For the same sample, all Bounding-Box-based combinations share the same Bounding Box, while all reference-based combinations share the same Reference Image, ensuring consistent editing semantics across different multimodal settings.

The construction rules for the five multimodal input combinations are summarized in Table~\ref{tab:dataset_prompt_construction}. Bracketed terms denote sample-specific content inserted into the corresponding prompt.

The source-side anchor is defined according to the editing task: the attachment or placement region for Addition, the source part to be removed for Deletion, the part to be replaced or modified for Modification, the region whose appearance changes for Appearance, and the articulated component involved in the pose change for Pose. The Reference Image encodes only the target visual information and does not provide spatial localization.

Figure~\ref{fig:A1} presents representative examples of the constructed multimodal editing conditions. Each dashed box contains the Bounding Box visualization, the corresponding Reference Image, and the original textual editing instruction. For clarity, the figure shows the original instruction rather than the modality-specific text variants, which are constructed according to Table~\ref{tab:dataset_prompt_construction}.

\subsection{Target-View Generation}
\label{A2}

For each sample, we first load the four canonical renderings of the source 3D
asset, i.e., the front, right, back, and left views, together with the current
multimodal editing condition. We use Gemini 3.6 Flash as a unified planning
module to interpret the editing intent solely from the available input
modalities and formulate a unified editing description $E$, including the
relevant source part, target content, requested operation, spatial relationship,
target geometry or appearance, and preservation requirements. No predefined
edit-type or task-category label is provided during target-view generation;
the required editing operation is inferred entirely from the current
multimodal condition.

\begin{table}[t]
\centering
\footnotesize
\setlength{\tabcolsep}{3pt}
\renewcommand{\arraystretch}{1.10}

\caption{
Prompt construction for target-view generation under the five multimodal input combinations.
The five settings share the same generation pipeline but differ in the roles assigned to Text, Bounding Box, and Reference Image.
}
\label{tab:target_view_prompt_construction}

\begin{tabular}{
@{}p{0.10\textwidth}
p{0.27\textwidth}
p{0.58\textwidth}@{}
}
\toprule

\textbf{Input} &
\textbf{Gemini 3 Pro Image Input} &
\textbf{Prompt Construction and Modality Roles}
\\
\midrule

$T$
&
\textit{[Source View]}
&
Text is the only explicit editing condition.
The textual instruction defines the requested operation and target semantics, while the spatial location is inferred jointly from the instruction and the source image.
The prompt explicitly prohibits introducing content that is not requested by the instruction.
\\
\midrule

$T+B$
&
\textit{[Source View]} +
\textit{[BBox Visualization]}
&
Text defines \textbf{what operation} should be performed, whereas the Bounding Box provides only a soft cue for \textbf{where} the relevant source region is located.
The box is not treated as a hard mask, exact boundary, crop, or target silhouette, and the actual edited area may extend slightly beyond or remain within the box when required for coherent geometry and appearance.
The BBox itself must not appear in the generated image.
\\
\midrule

$T+R$
&
\textit{[Source View]} +
\textit{[Reference Image]}
&
Text is authoritative for the editing operation, while the Reference Image provides the requested visual target information.
Depending on the editing intent inferred from the available conditions, the reference may provide target identity, shape, parts, color, material, or texture.
Its framing, pose, background, shadows, and unrelated content are not directly copied; instead, the relevant visual attributes are adapted to the source image.
\\
\midrule

$B+R$
&
\textit{[Source View]} +
\textit{[BBox Visualization]} +
\textit{[Reference Image]}
&
No user-provided Text condition, original textual instruction, or edit-type label is supplied.
The Bounding Box provides approximate \textbf{where} information, while the Reference Image provides \textbf{what / appearance} information.
The editing operation is inferred from the source image, the localized source region, and the visual relationship between the source and the reference, and is represented internally by the unified editing description $E$.
\\
\midrule

$T+B+R$
&
\textit{[Source View]} +
\textit{[BBox Visualization]} +
\textit{[Reference Image]}
&
All three modalities jointly constrain the edit.
Text is authoritative for the requested \textbf{operation / semantics}, the Bounding Box provides soft \textbf{where} guidance, and the Reference Image provides the desired \textbf{what / appearance}.
The spatial and visual conditions complement the textual instruction without overriding its editing intent.
\\

\bottomrule
\end{tabular}
\end{table}

The source-view selection strategy depends on the available modalities. For
$T$ and $T+R$, Gemini 3.6 Flash selects the source view that is expected to
provide the clearest evidence for the requested edit, considering the
visibility of the target-relevant region, occlusion, local attachment or
context, and preservation of unrelated geometry. For $T+B$, $B+R$, and
$T+B+R$, the Bounding Box is already associated with a particular source view
during multimodal condition construction. We therefore directly recover and
lock this corresponding view rather than performing semantic view reselection.

\begin{figure}[t]
\centering

\setlength{\tabcolsep}{2.5pt}
\renewcommand{\arraystretch}{1.0}

%
\newcommand{\targetviewpair}[2]{%
\begin{tikzpicture}[baseline=(current bounding box.center)]
\node[
    draw=black!70,
    dashed,
    line width=0.55pt,
    rounded corners=5pt,
    inner xsep=4pt,
    inner ysep=4pt
] {%
    \begin{minipage}{0.295\textwidth}
        \centering

        \begin{minipage}[c]{0.485\linewidth}
            \centering
            \includegraphics[
                width=\linewidth,
                height=1.75cm,
                keepaspectratio
            ]{#1}
        \end{minipage}%
        \hfill
        \begin{minipage}[c]{0.485\linewidth}
            \centering
            \includegraphics[
                width=\linewidth,
                height=1.75cm,
                keepaspectratio
            ]{#2}
        \end{minipage}

    \end{minipage}%
};
\end{tikzpicture}%
}

\newcommand{\targetviewheader}{%
\begin{minipage}{0.295\textwidth}
    \centering
    \scriptsize
    \begin{minipage}{0.485\linewidth}
        \centering
        \textbf{Source View}
    \end{minipage}%
    \hfill
    \begin{minipage}{0.485\linewidth}
        \centering
        \textbf{Target View}
    \end{minipage}
\end{minipage}%
}

\begin{tabular}{@{}ccc@{}}

\targetviewheader
&
\targetviewheader
&
\targetviewheader
\\[1.2mm]

\targetviewpair
    {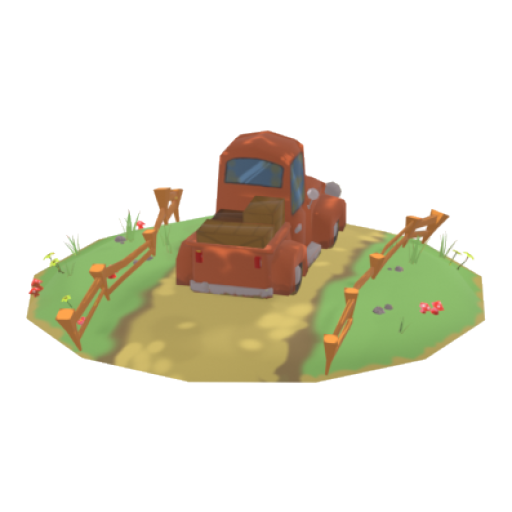}
    {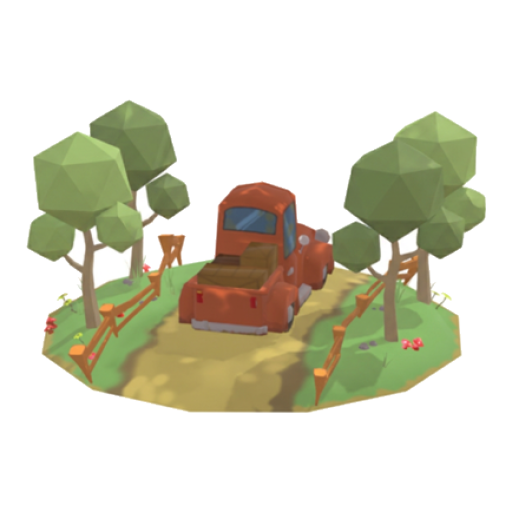}
&
\targetviewpair
    {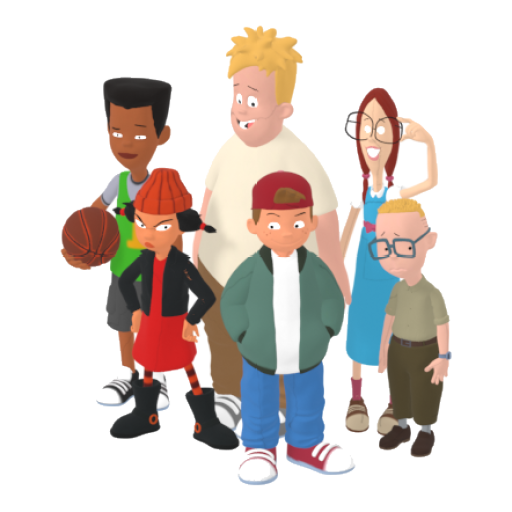}
    {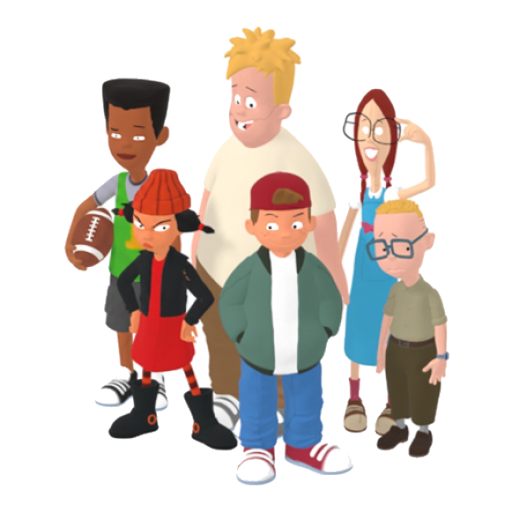}
&
\targetviewpair
    {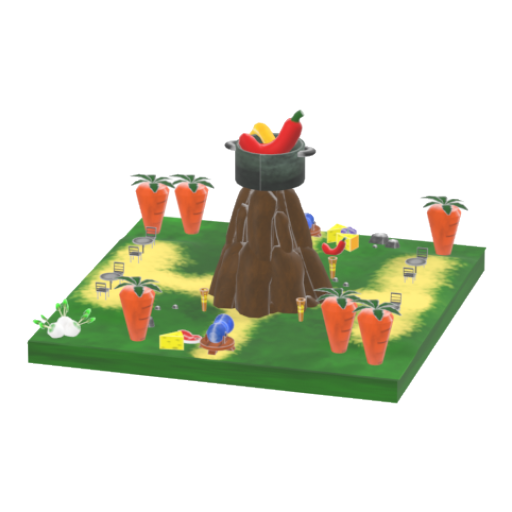}
    {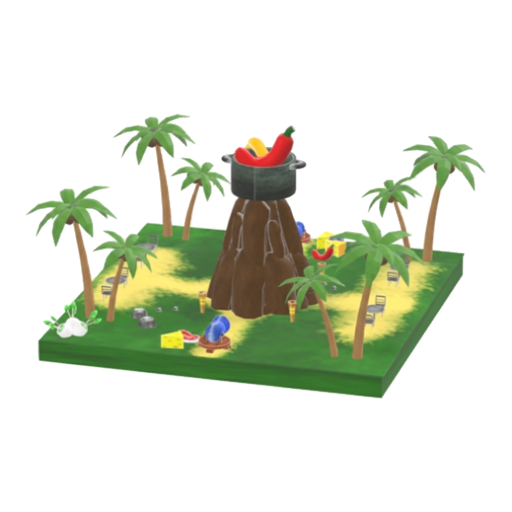}
\\[10mm]

\targetviewpair
    {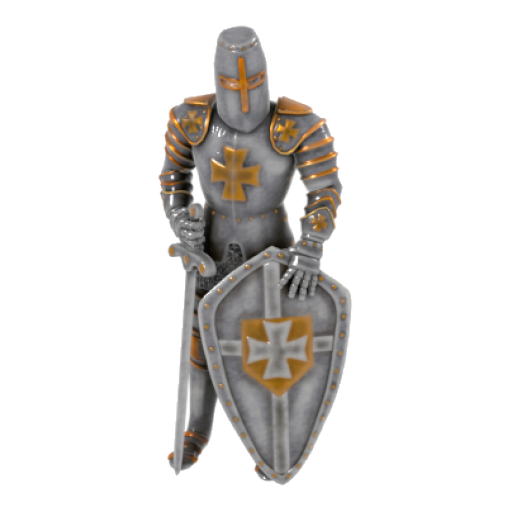}
    {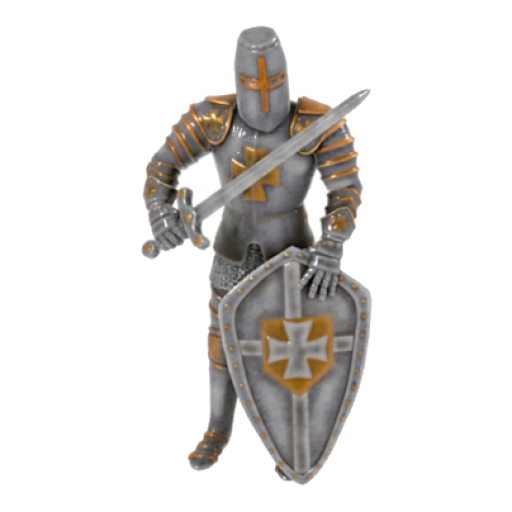}
&
\targetviewpair
    {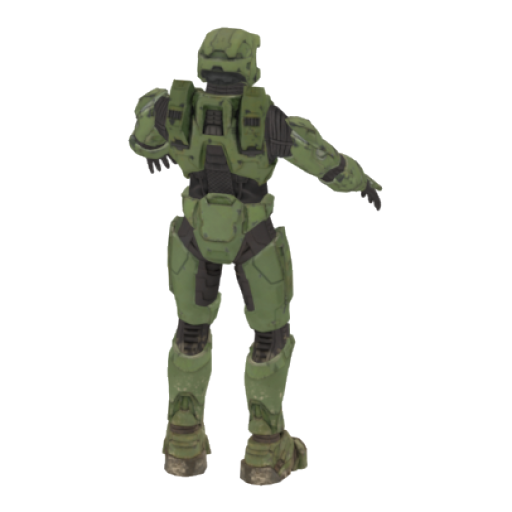}
    {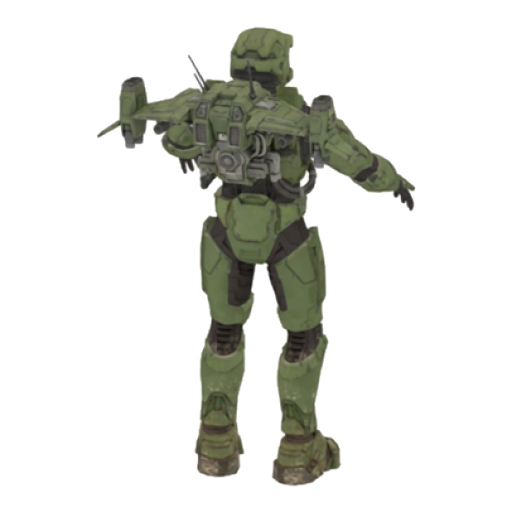}
&
\targetviewpair
    {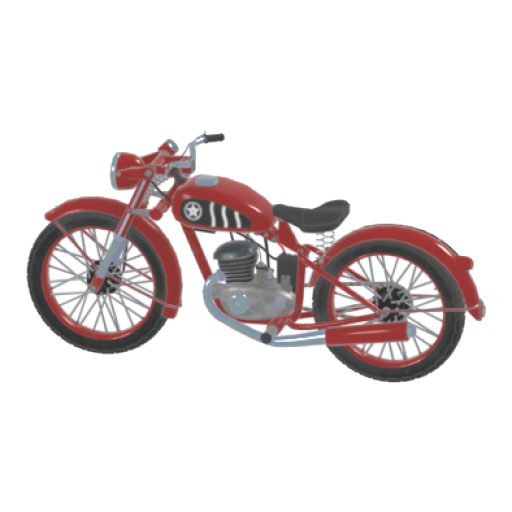}
    {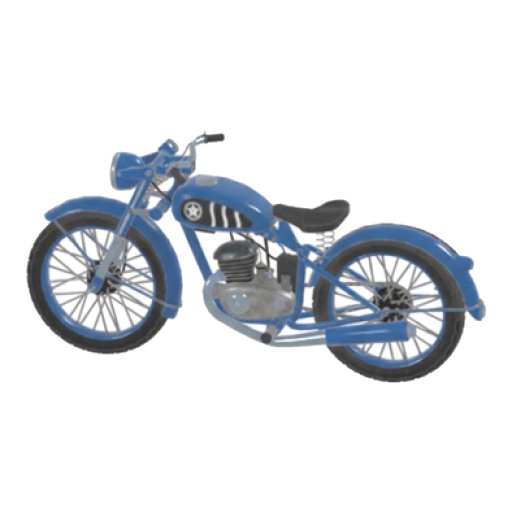}
\\[10mm]

\targetviewpair
    {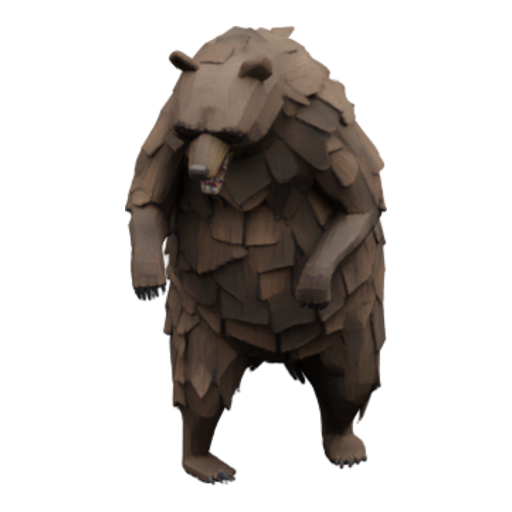}
    {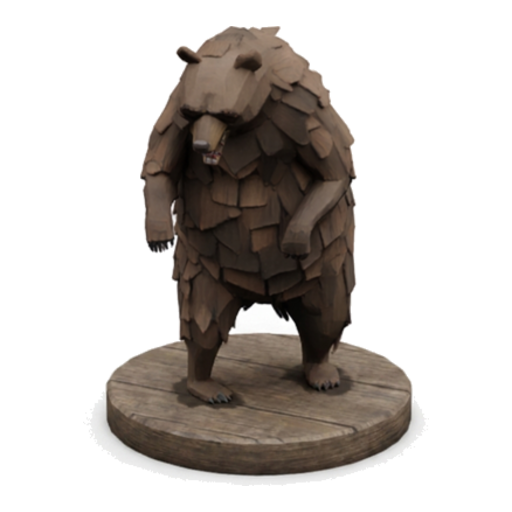}
&
\targetviewpair
    {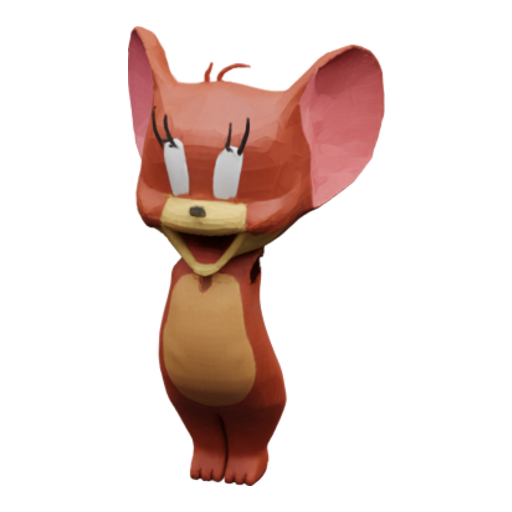}
    {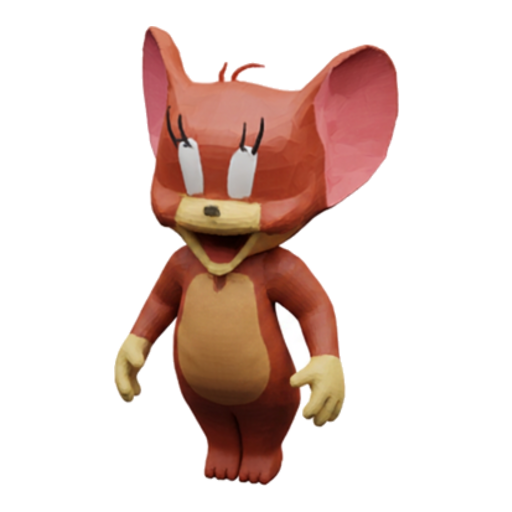}
&
\targetviewpair
    {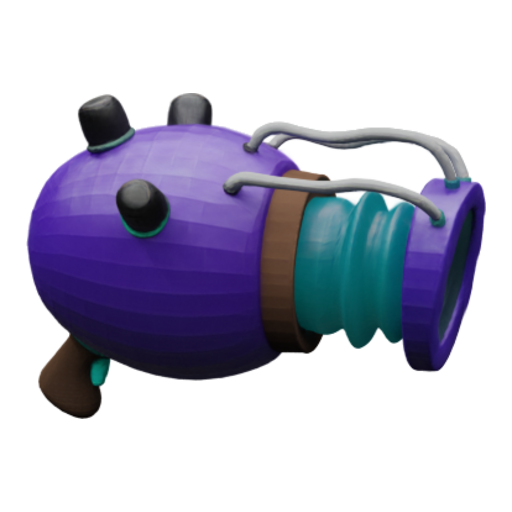}
    {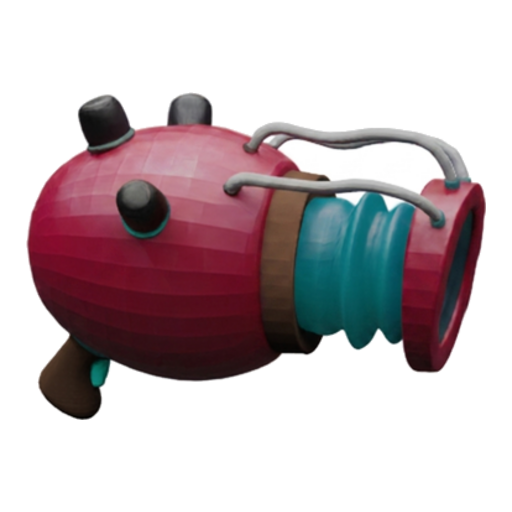}
\\[10mm]

\targetviewpair
    {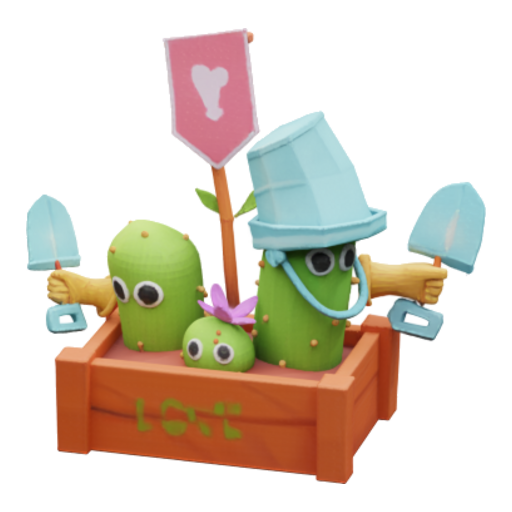}
    {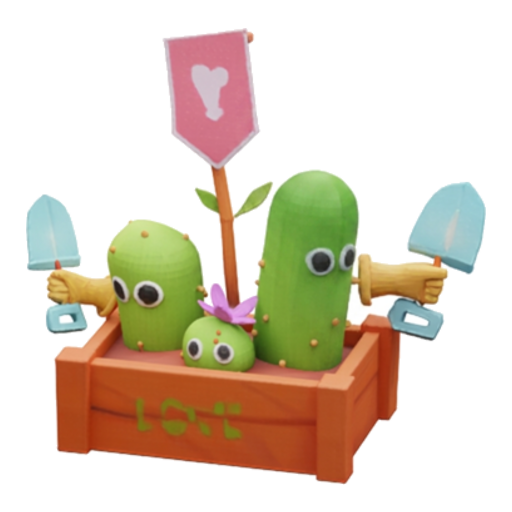}
&
\targetviewpair
    {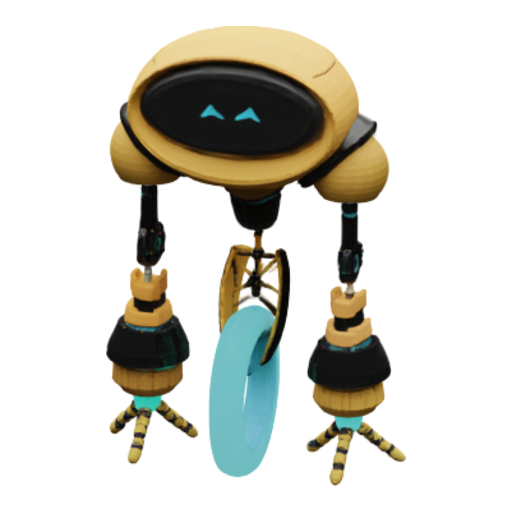}
    {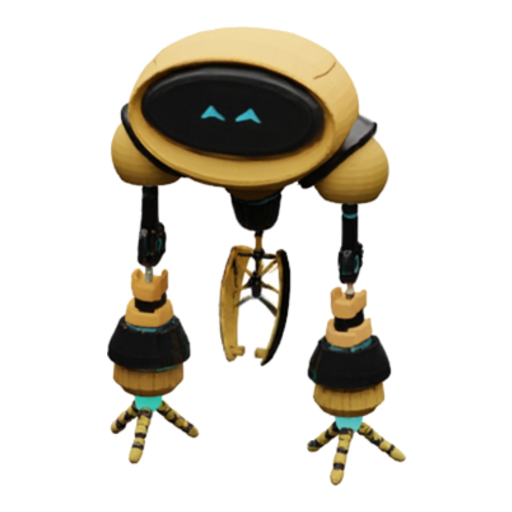}
&
\targetviewpair
    {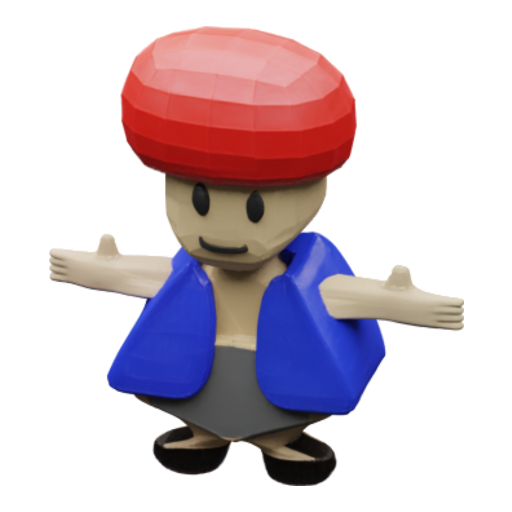}
    {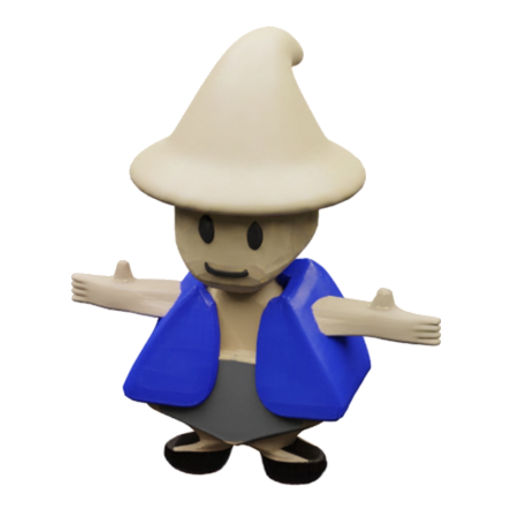}
\\[10mm]

\targetviewpair
    {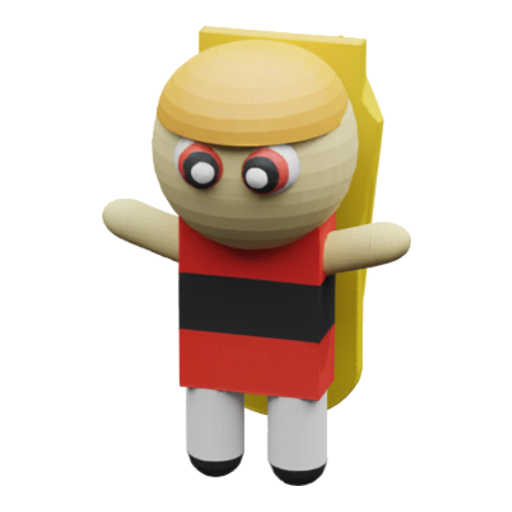}
    {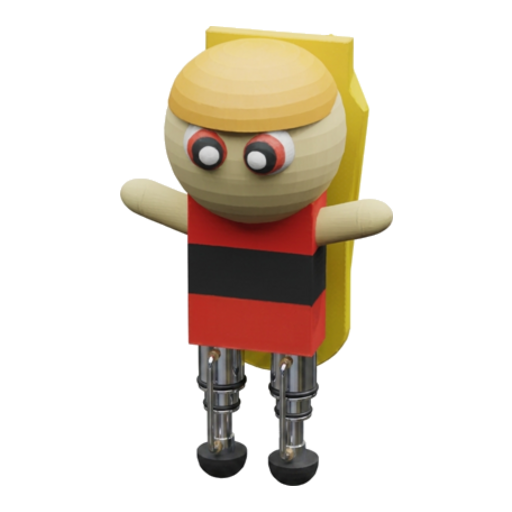}
&
\targetviewpair
    {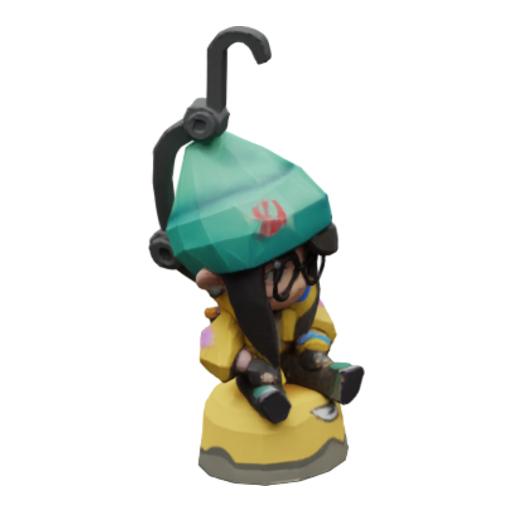}
    {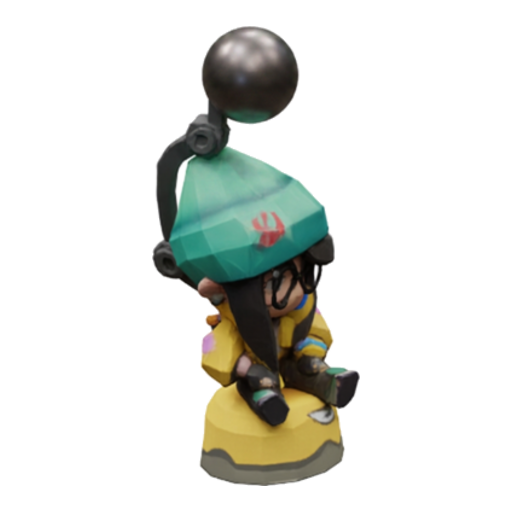}
&
\targetviewpair
    {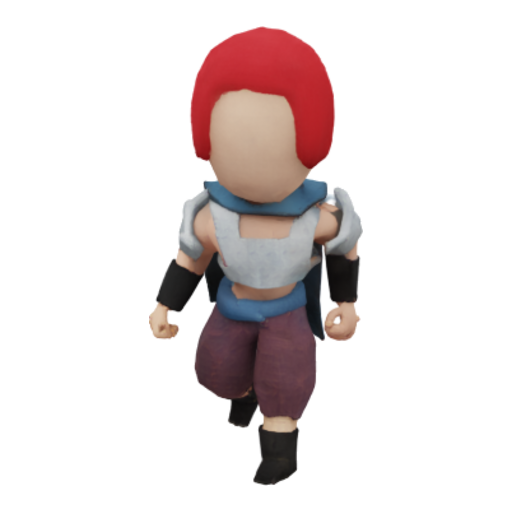}
    {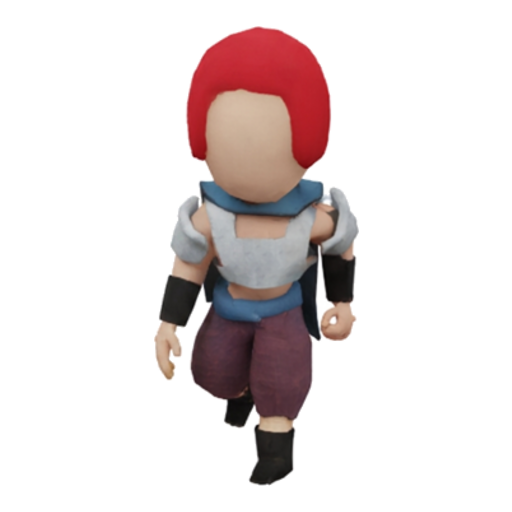}
\\[10mm]

\targetviewpair
    {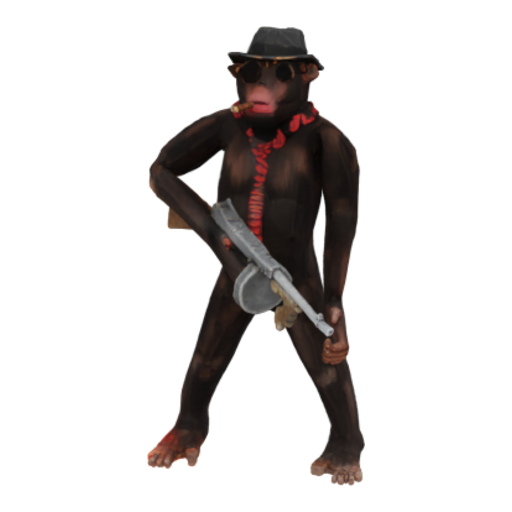}
    {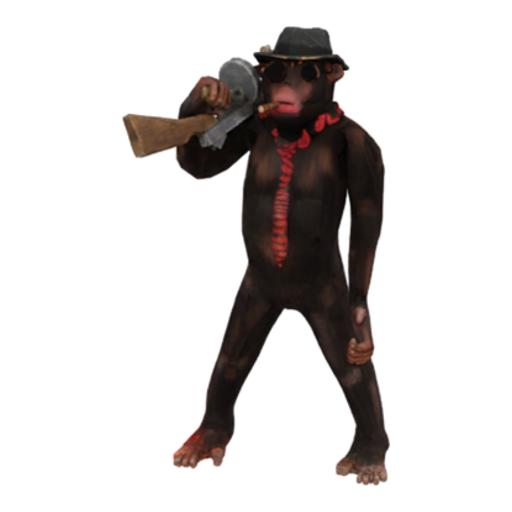}
&
\targetviewpair
    {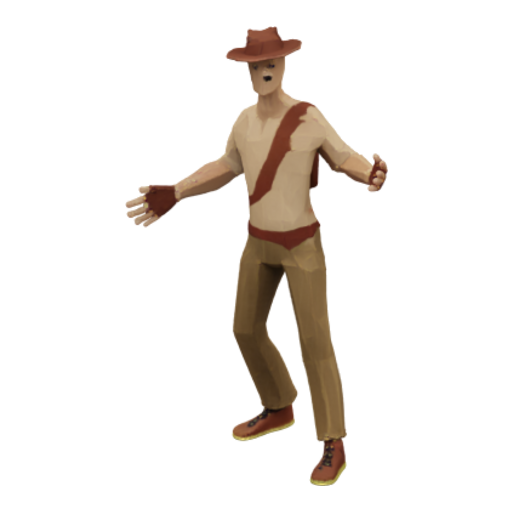}
    {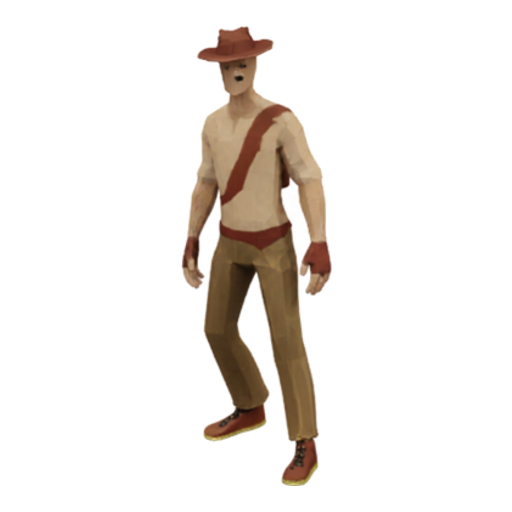}
&
\targetviewpair
    {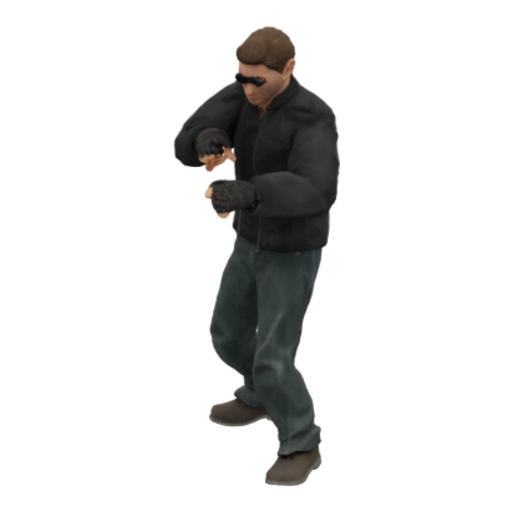}
    {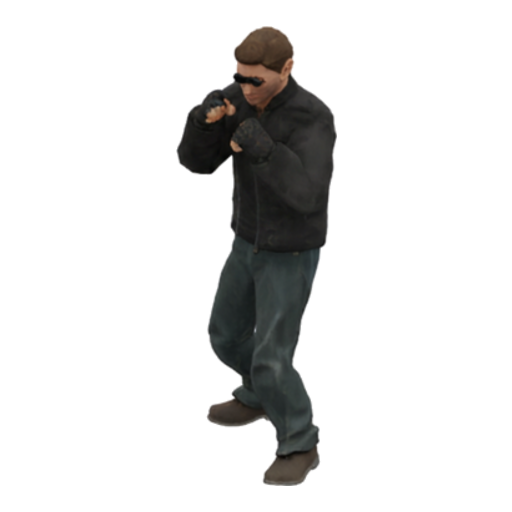}

\end{tabular}

\caption{
Examples of target-view generation.
Each dashed box groups a selected source view and its corresponding target view generated by Gemini 3 Pro Image.
The examples cover diverse objects and editing operations.
}
\label{fig:A2}

\end{figure}

Given the unified editing description $E$ and the selected source view, we
construct a modality-specific image-editing prompt and use Gemini 3 Pro Image
to generate the target view $I_t$. The selected source view always serves as
the editing canvas, whereas Text, Bounding Box, and Reference Image provide
complementary editing guidance. Text primarily specifies the requested
operation and target semantics; the Bounding Box provides only a soft spatial
cue for the relevant source region; and the Reference Image provides the
desired target identity, shape, part, color, material, or texture when
applicable. For $B+R$, no user-provided Text condition, original textual
instruction, or edit-type label is supplied; the required operation is inferred
from the available visual and spatial cues and represented internally by $E$.
In particular, the Bounding Box is explicitly treated as an approximate
semantic localization cue rather than a pixel-level mask, exact editing
boundary, crop, or target silhouette. Similarly, visual information from the
Reference Image is adapted to the viewpoint, scale, pose, lighting, and
structural context of the source image rather than being directly copied.

Across all input settings, the generation prompt consistently requires
preservation of unrelated object identity, unedited geometry and proportions,
viewpoint, orientation, framing, lighting, shadows, and the original clean
white background. The modality-specific prompt construction is summarized in
Table~\ref{tab:target_view_prompt_construction}.

Although no explicit task-category label is provided, the unified editing
description $E$ induces operation-specific constraints according to the
editing intent inferred from the current multimodal condition. For an inferred
addition operation, the requested content is inserted with appropriate scale
and natural physical connectivity. For an inferred deletion operation, the
target content is completely removed and the newly exposed surface is
naturally reconstructed. For a local modification, only the relevant source
part is modified while maintaining coherent geometry and intended connections.
For an appearance change, the requested visual attributes are modified while
preserving the geometry, size, pose, and placement of the corresponding part.
For a pose change, the relevant articulation or part configuration is changed
while preserving object identity and plausible structural connectivity.
These constraints are determined from the inferred editing semantics rather
than selected using an externally provided task label.

Figure~\ref{fig:A2} presents additional examples of target-view generation.
Each pair shows the selected source view and the corresponding target view
generated by Gemini 3 Pro Image. The examples cover diverse objects and
editing operations and illustrate that the generated target views primarily
modify the requested content while preserving the viewpoint, overall
structure, appearance style, and unrelated regions of the source image.

\subsection{Multimodal Condition--Task Coverage}
\label{A3}

UMAS3D supports five multimodal input combinations: $T$, $T+B$, $T+R$, $B+R$, and $T+B+R$. In our experimental setting, Addition, Modification, and Appearance cover all five combinations, whereas Deletion and Pose use $T$ and $T+B$. This configuration is not obtained by mechanically applying every modality to every task, but is designed according to the information provided by each modality for different editing intents. Text primarily specifies the editing operation and semantics, the Bounding Box provides spatial localization of the edit-relevant source region, and the Reference Image provides complementary visual information about the target object, part, or appearance.

For \textbf{Addition}, \textbf{Modification}, and \textbf{Appearance}, the Reference Image provides explicit target-side visual information. For Addition, $R$ can describe the identity, shape, or appearance of the object to be added. For Modification, $R$ provides a visual reference for the desired target part or structure. For Appearance, $R$ can specify target visual attributes such as color, material, texture, or pattern. These tasks therefore include the reference-based settings $T+R$, $B+R$, and $T+B+R$. Under $B+R$, no additional Text condition or task-category label is supplied; the editing semantics are inferred jointly from the source view, the spatial information provided by the Bounding Box, and the target visual information conveyed by the Reference Image.

For \textbf{Deletion} and \textbf{Pose}, we do not construct Reference-Image-based conditions and instead use $T$ and $T+B$. Deletion is defined by removing existing source content and therefore primarily requires specifying what should be removed and where it is located. Pose changes the articulation, pose, or spatial configuration of existing parts, with the condition primarily describing motion semantics and spatial relations. Based on this condition design, we use Reference Images for editing scenarios that require explicit visual specification of newly added, replaced, or target appearance content, while not constructing Reference-Image-based conditions for Deletion and Pose. This setting also reflects how different editing intents can be expressed through complementary input modalities in practical editing scenarios.

\begin{figure}[H]
\centering
\includegraphics[width=0.99\linewidth]{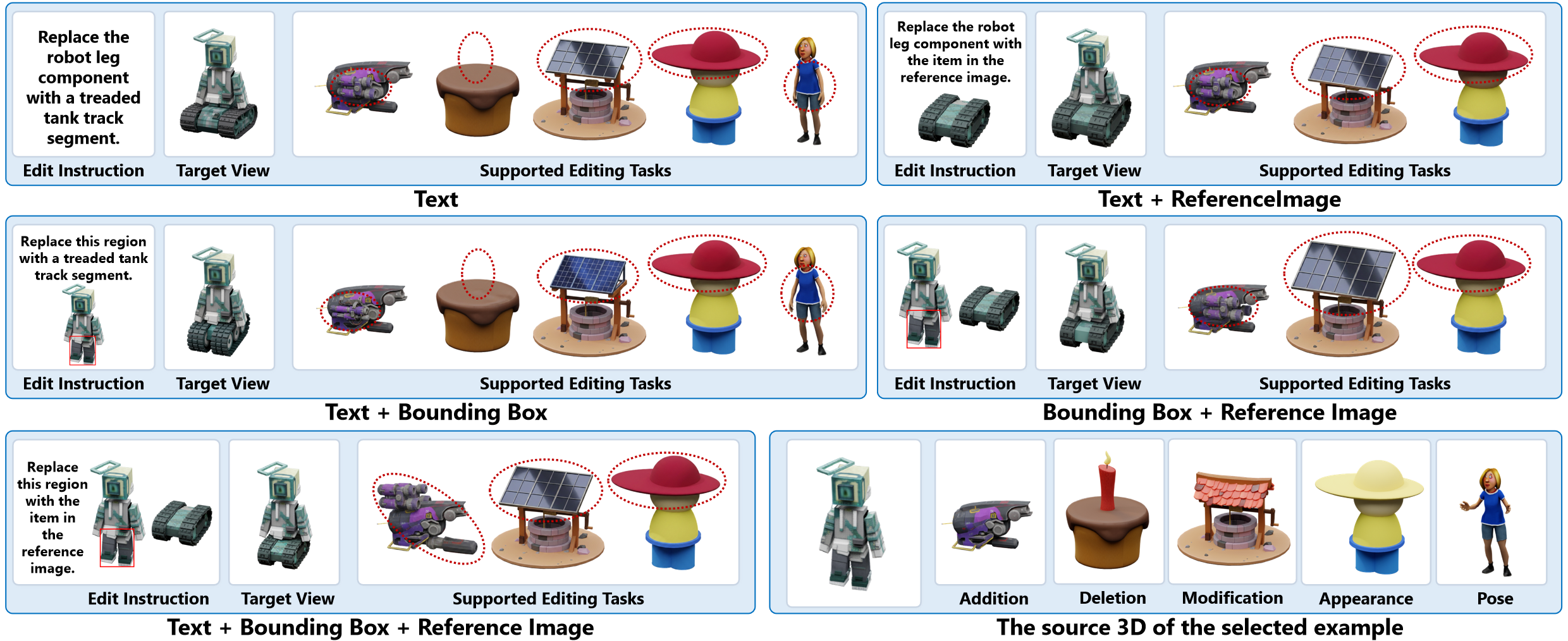}
\caption{\textbf{Unified 3D editing across multimodal instruction combinations and editing tasks.}
UMAS3D converts heterogeneous multimodal conditions into target-view guidance and handles Addition, Deletion, Modification, Appearance and Pose through the same 3D editing pipeline.}
\label{fig:unified_editing_appendix}
\end{figure}

Despite these task-dependent input combinations, the downstream processing remains unified. All available conditions are first converted into target-view guidance through the multimodal interface described in Section~\ref{A2}. Once the target view is obtained, all five editing tasks share the same downstream 3D editing process without task-specific 3D editing pipelines. As illustrated in Figure~\ref{fig:unified_editing_appendix}, UMAS3D therefore achieves unification at two complementary levels: heterogeneous multimodal conditions are converted into target-view guidance at the \emph{condition level}, while the five editing tasks share the same downstream 3D editing process at the \emph{task level}.

\subsection{Visualization of MARM Initialization and Adaptive Region Refinement}
\label{sec:appendix_marm_visualization}

UMAS3D does not require any externally provided or manually annotated 3D
editing mask. Given the source 3D asset, its multi-view observations, and the
generated target image, MARM first compares patch-level DINOv2 correspondences
between the selected source view and the target image at the same and neighboring
locations, together with foreground, appearance, and edge differences, to obtain
2D Edit/Keep evidence. Target-only content without reliable source correspondence
provides additional target-side Edit evidence for opening editable space in
source-empty regions.

To construct complete 3D regions, we do not directly back-project a single 2D
mask. Instead, source-occupied Grid64 voxels are projected into the four source
views and aggregate visible features from the frozen DINOv2 encoder. These voxel
features are grouped according to DINOv2 feature similarity, 3D spatial distance,
and connectivity to form complete 3D macro regions. Each 3D region is then
projected to the selected-view Edit/Keep evidence map, where region-level evidence
aggregation determines its initial state, yielding
$\mathbf{M}_0=(\mathbf{M}_0^E,\mathbf{M}_0^K)$.

The formal MARM partition contains only Edit and Keep states. Unknown and
Target-New are used only as intermediate initialization evidence: Unknown denotes
insufficient source-side evidence before region-level assignment, whereas
Target-New denotes target-side content without reliable source correspondence and
is ultimately incorporated into the editable support. They therefore do not
introduce additional MARM region states.

Importantly, $\mathbf{M}_0$ serves only as the initial region prior rather than a
fixed editing boundary. After an initial warm-up period, MARM periodically
re-evaluates the current 3D regions according to the evolving target structure.
A region is reassigned only when persistent evidence supports a transition. A
Keep region can be released to Edit when its target-driven structural deviation
remains consistently larger than that of comparable preserved regions.
Conversely, an Edit region can be reassigned to Keep when its state remains
consistently source-like. SGPS always follows the current region partition, so
its source-preservation support evolves together with MARM refinement.

\begin{figure}[t]
\centering

\setlength{\tabcolsep}{0pt}
\renewcommand{\arraystretch}{1.0}

%
\newcommand{\maskinitexample}[6]{%
\begin{tikzpicture}[baseline=(current bounding box.center)]
\node[
    draw=black!70,
    dashed,
    line width=0.55pt,
    rounded corners=5pt,
    inner xsep=3pt,
    inner ysep=4pt
] {%
    \begin{minipage}{0.985\textwidth}
        \centering

        \begin{tabular}{
            @{}
            >{\centering\arraybackslash}m{0.160\linewidth}
            >{\centering\arraybackslash}m{0.162\linewidth}
            >{\centering\arraybackslash}m{0.162\linewidth}
            >{\centering\arraybackslash}m{0.162\linewidth}
            >{\centering\arraybackslash}m{0.162\linewidth}
            >{\centering\arraybackslash}m{0.162\linewidth}
            @{}
        }

        \scriptsize #1
        &
        \includegraphics[
            width=0.98\linewidth,
            height=2.15cm,
            keepaspectratio
        ]{#2}
        &
        \includegraphics[
            width=0.98\linewidth,
            height=2.15cm,
            keepaspectratio
        ]{#3}
        &
        \includegraphics[
            width=0.98\linewidth,
            height=2.15cm,
            keepaspectratio
        ]{#4}
        &
        \includegraphics[
            width=0.98\linewidth,
            height=2.15cm,
            keepaspectratio
        ]{#5}
        &
        \includegraphics[
            width=0.98\linewidth,
            height=2.15cm,
            keepaspectratio
        ]{#6}

        \end{tabular}

    \end{minipage}%
};
\end{tikzpicture}%
}

\newcommand{\maskinitheader}{%
\begin{minipage}{0.985\textwidth}
    \centering
    \tiny

    \begin{tabular}{
        @{}
        >{\centering\arraybackslash}m{0.160\linewidth}
        >{\centering\arraybackslash}m{0.162\linewidth}
        >{\centering\arraybackslash}m{0.162\linewidth}
        >{\centering\arraybackslash}m{0.162\linewidth}
        >{\centering\arraybackslash}m{0.162\linewidth}
        >{\centering\arraybackslash}m{0.162\linewidth}
        @{}
    }

    \textbf{Editing Instruction}
    &
    \textbf{Source View}
    &
    \textbf{Target View}
    &
    \textbf{Coarse 3D Prior}
    &
    \shortstack{
        \textbf{Initial Edit/Keep}\\
        \textbf{Partition $\mathbf{M}_0$}
    }
    &
    \shortstack{
        \textbf{Final Edit/Keep}\\
        \textbf{Partition $\mathbf{M}_T$}
    }

    \end{tabular}

\end{minipage}%
}

\begin{tabular}{@{}c@{}}

\maskinitheader
\\[1.5mm]

\maskinitexample
    {\textit{Remove the two stools.}}
    {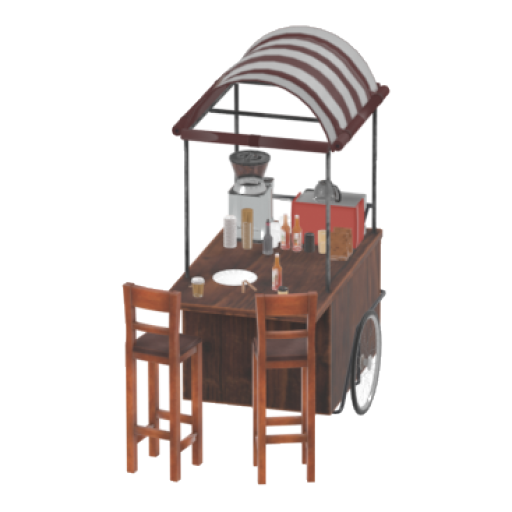}
    {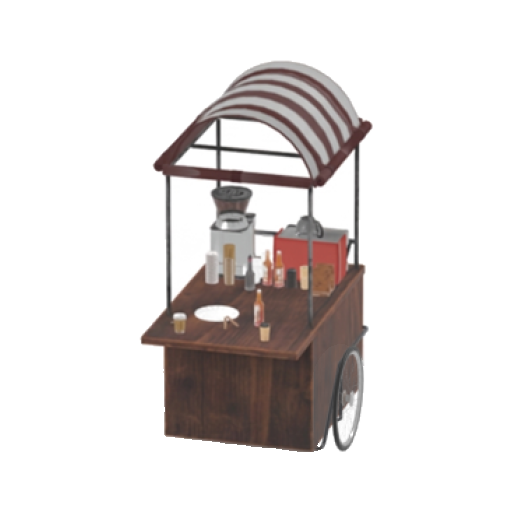}
    {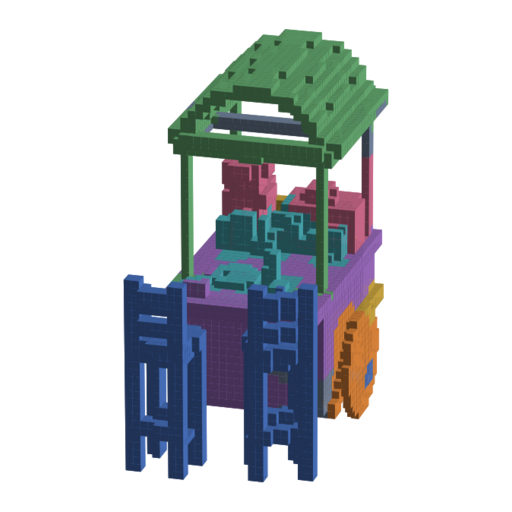}
    {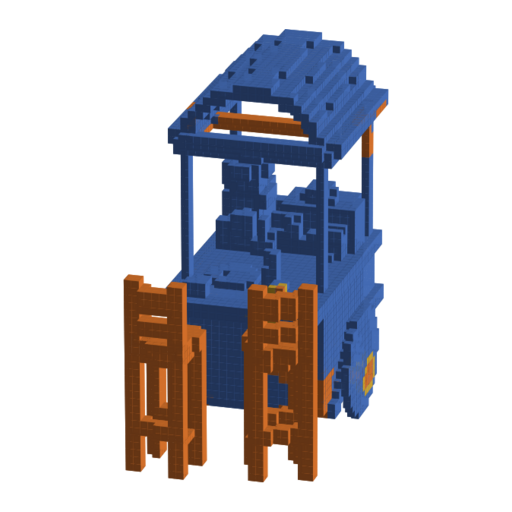}
    {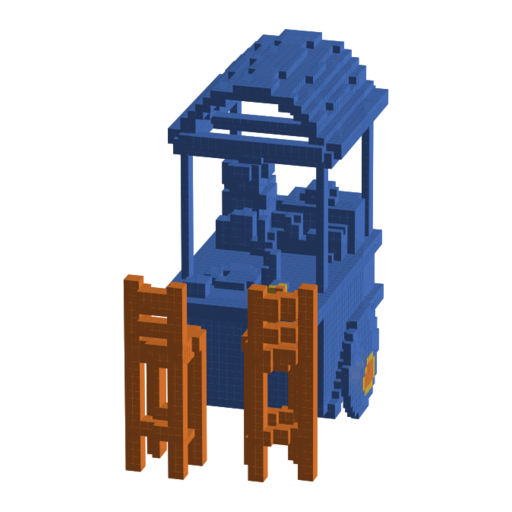}
\\[13mm]

\maskinitexample
    {\textit{Raise your hands.}}
    {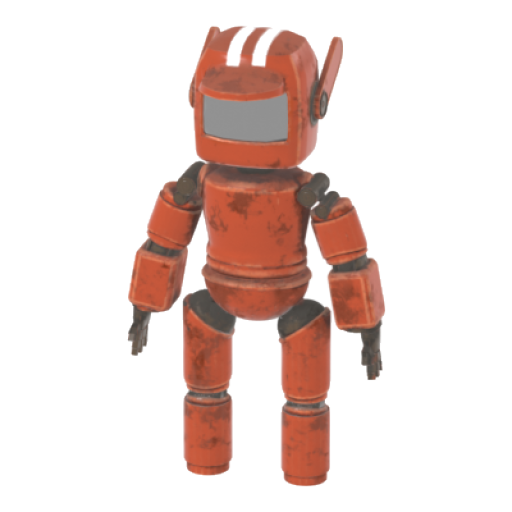}
    {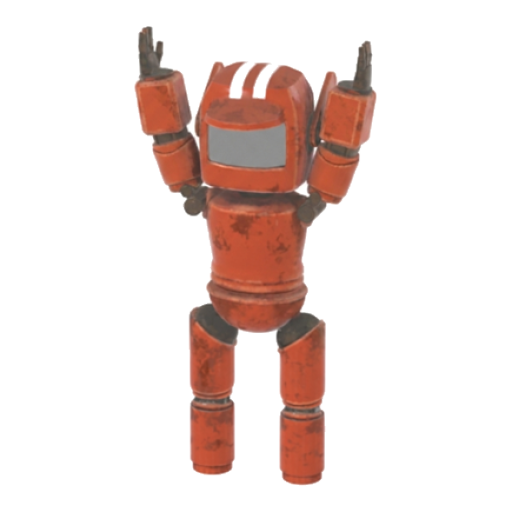}
    {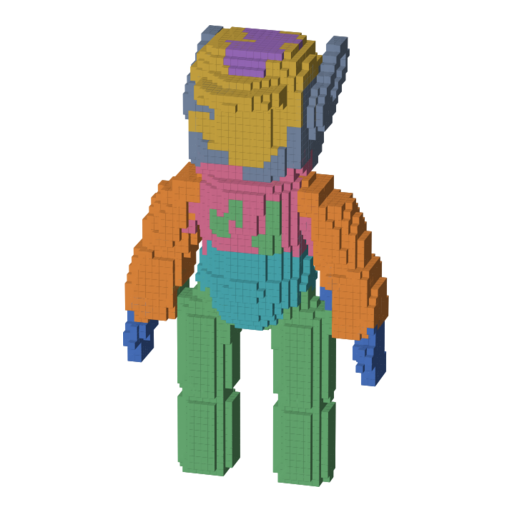}
    {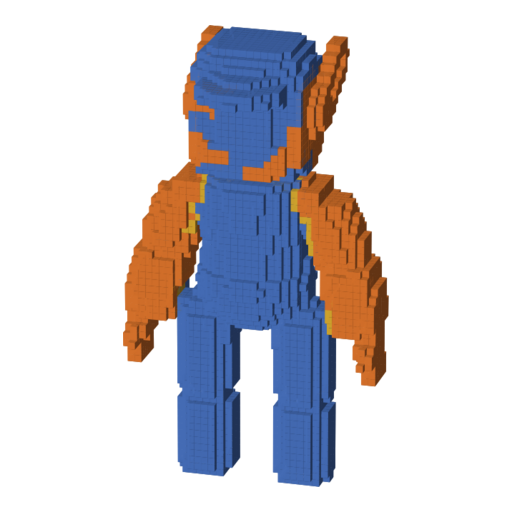}
    {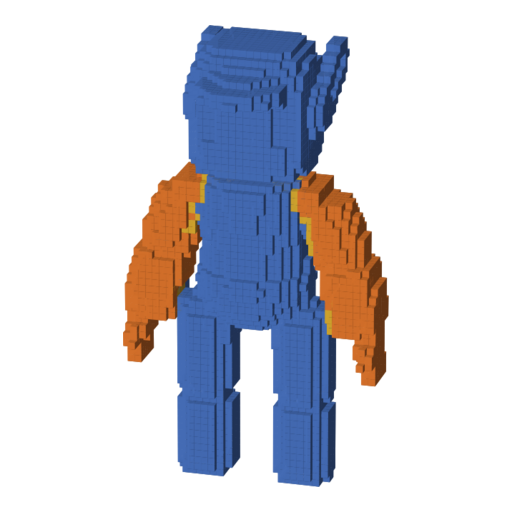}
\\[13mm]

\maskinitexample
    {\textit{Repaint the seat deep crimson red.}}
    {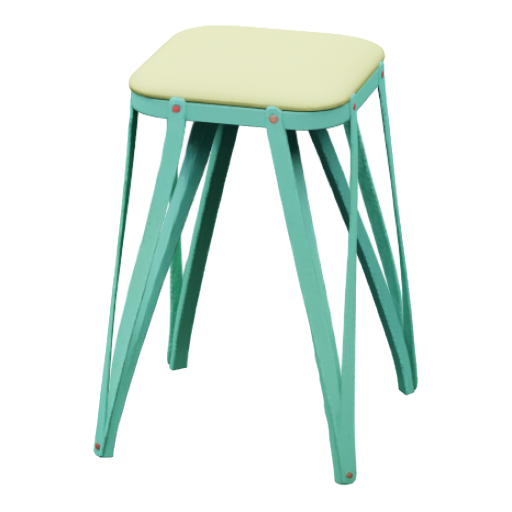}
    {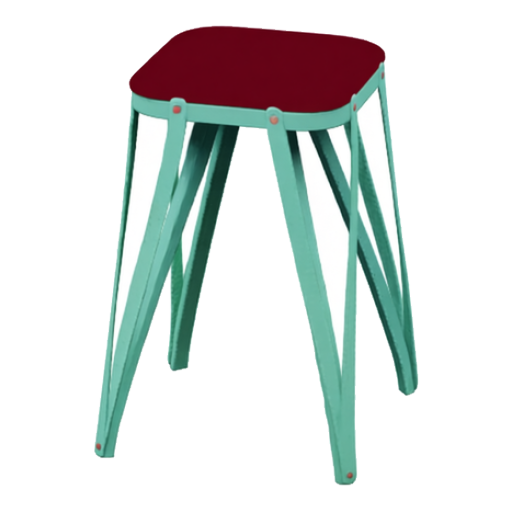}
    {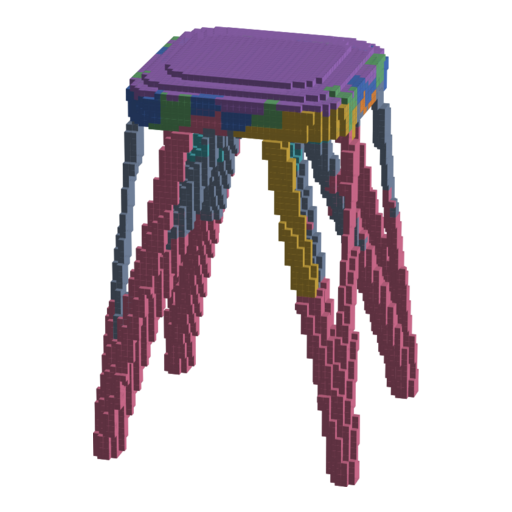}
    {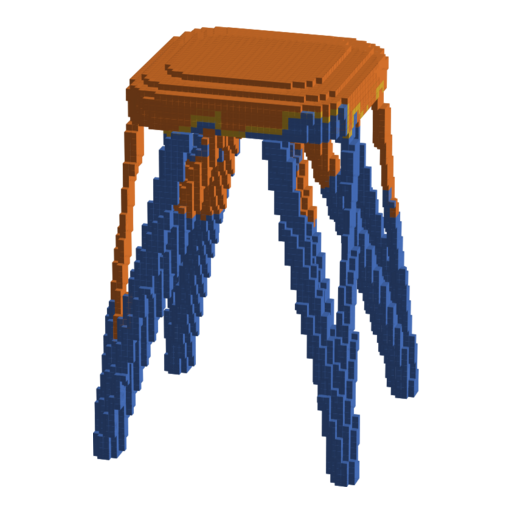}
    {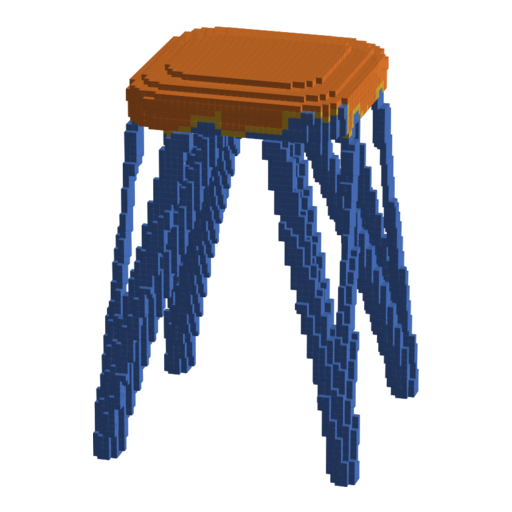}
\\[13mm]

\maskinitexample
    {\textit{Replace the knife with a baseball bat.}}
    {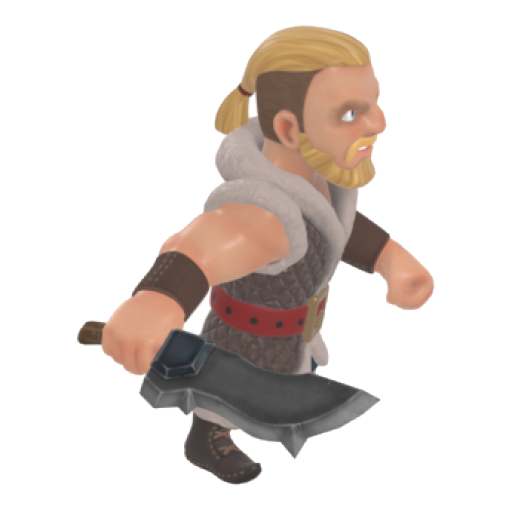}
    {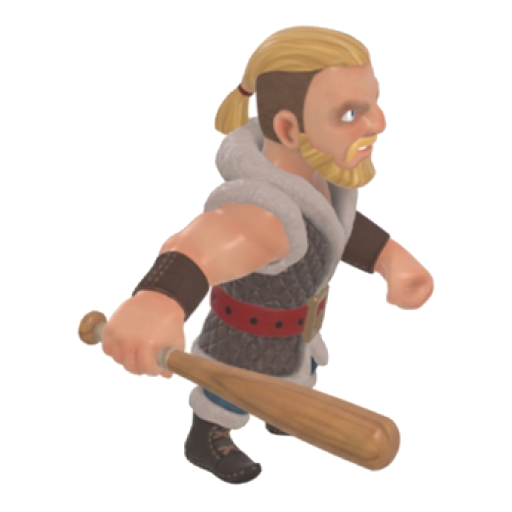}
    {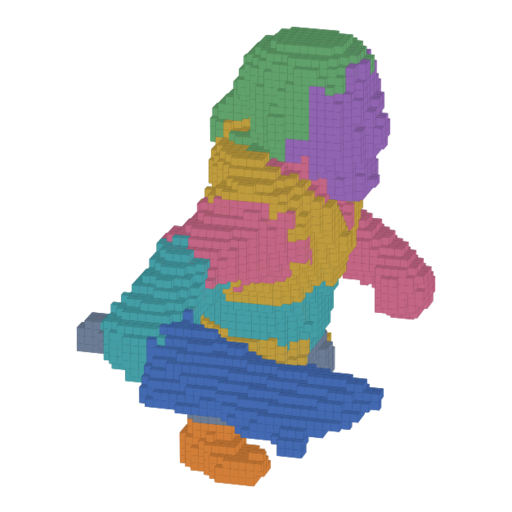}
    {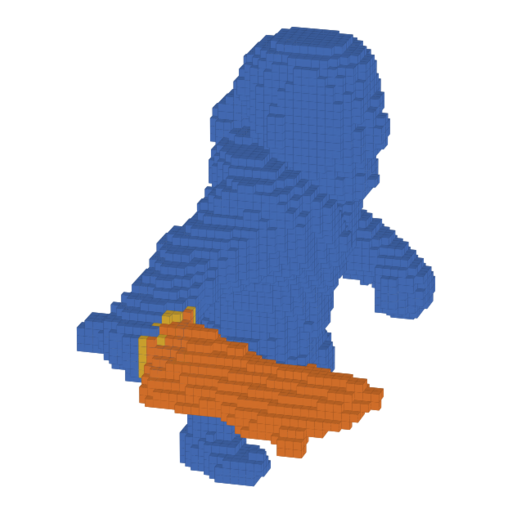}
    {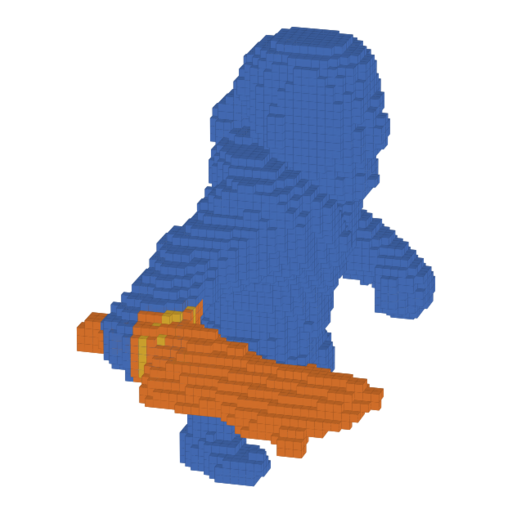}
\\[13mm]

\maskinitexample
    {\textit{Remove the detached bow and the quiver filled with arrows.}}
    {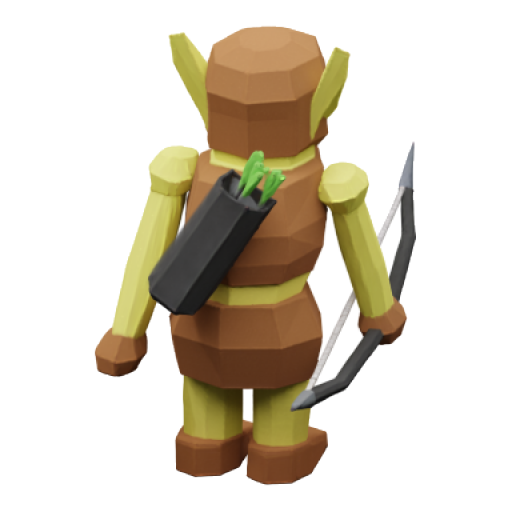}
    {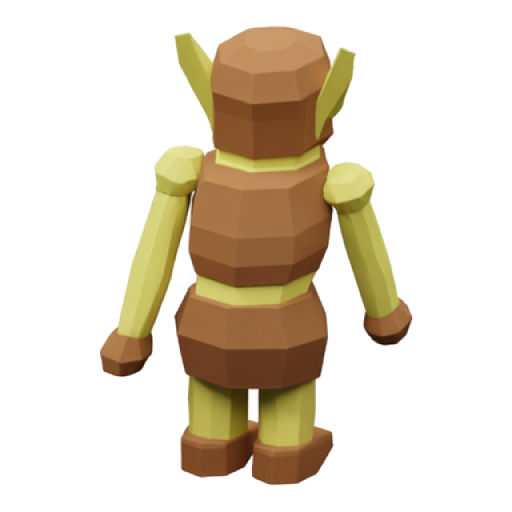}
    {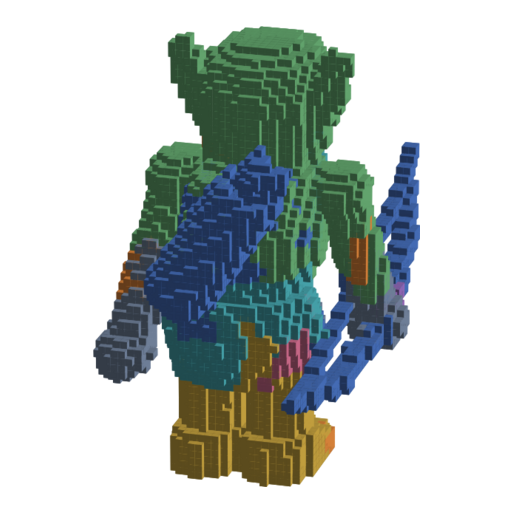}
    {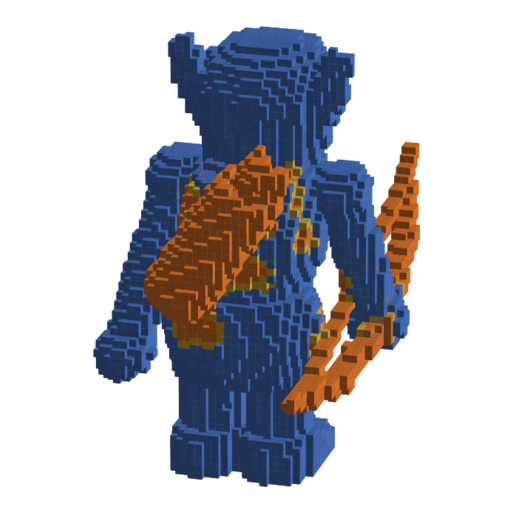}
    {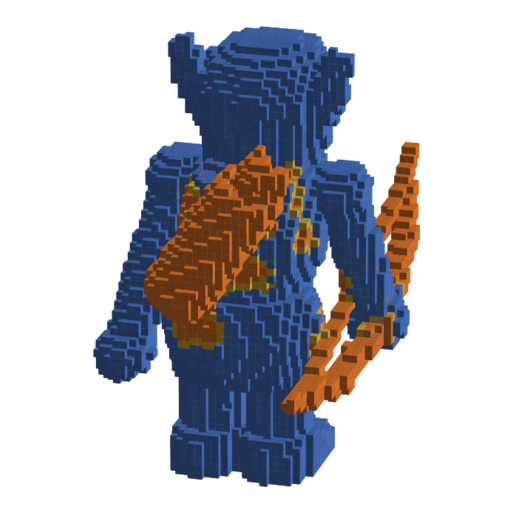}
\\[13mm]

\maskinitexample
    {\textit{Remove the carriage.}}
    {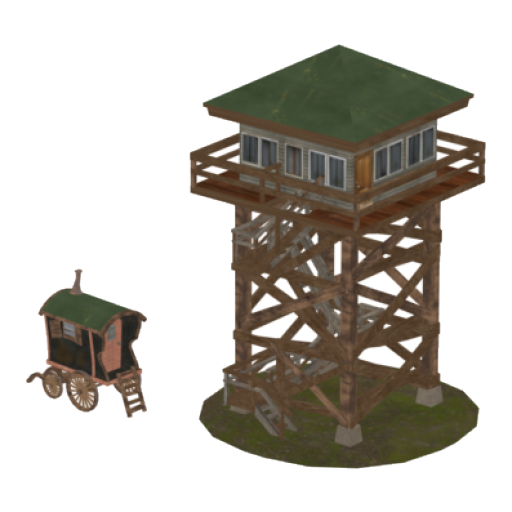}
    {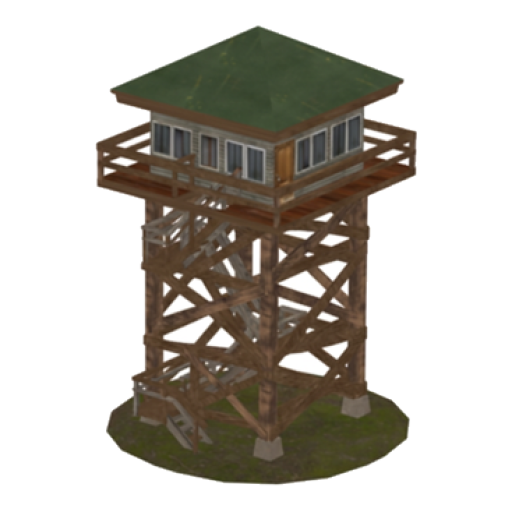}
    {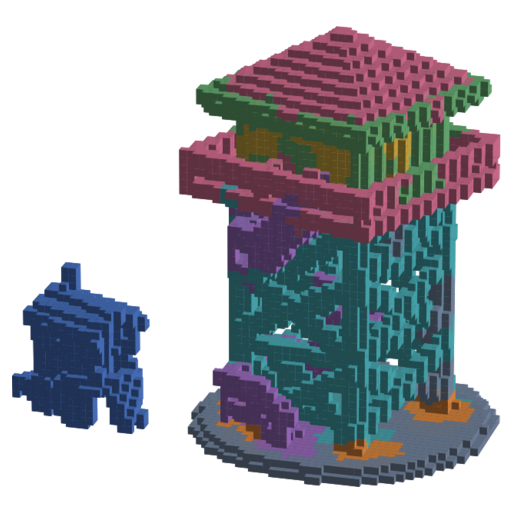}
    {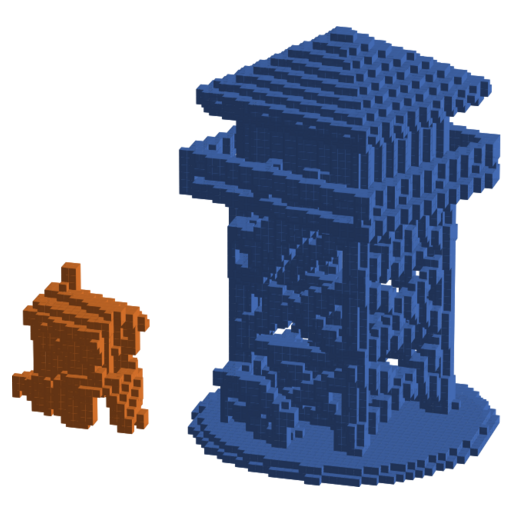}
    {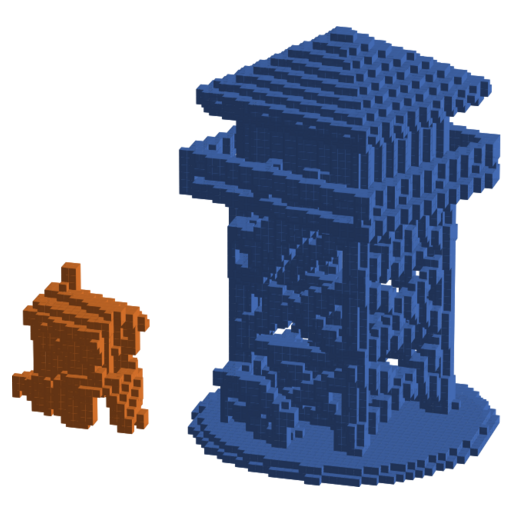}

\end{tabular}

\caption{
\textbf{Visualization of MARM initialization and adaptive region refinement.}
Each row shows the editing instruction, selected source view, generated target
view, coarse 3D region prior, initial Edit/Keep partition $\mathbf{M}_0$, and
final partition $\mathbf{M}_T$ after Stage~1 refinement. Differences between
$\mathbf{M}_0$ and $\mathbf{M}_T$ directly visualize the online reassignment of
Edit/Keep regions within the single Stage~1 trajectory.
}
\label{fig:A3}

\end{figure}

Figure~\ref{fig:A3} directly visualizes this adaptive process. For each example,
we show the editing instruction, selected source view, generated target view,
coarse 3D region prior, initial Edit/Keep partition $\mathbf{M}_0$, and final
partition $\mathbf{M}_T$ obtained after Stage~1 refinement. Differences between
$\mathbf{M}_0$ and $\mathbf{M}_T$ provide process-level evidence that the
Edit/Keep support is actively reassigned during generation rather than remaining
fixed after initialization.

At the end of Stage~1, the refined partition $\mathbf{M}_T$ is frozen and passed
to Stage~2. No further MARM region reassignment is performed in Stage~2; instead,
the final Keep support $\mathbf{M}_T^K$ is directly used by SGPS for source-feature
preservation in SLAT space. Thus, Figure~\ref{fig:A3} complements the quantitative
ablation in the main paper by explicitly visualizing how the initial partition
evolves within the single Stage~1 generation trajectory.

\subsection{Model and Inference Settings}
\label{app:model_settings}

\noindent\textbf{Training-Free Setting.}
UMAS3D is fully training-free: all pretrained model parameters remain frozen
throughout inference, with no additional training, fine-tuning, or test-time
parameter adaptation. The optimization steps in SGPS act only on the current
generation state and never update pretrained model parameters. TRELLIS
Stage~1 and Stage~2 use 100 and 25 inference steps, respectively, with the
same configuration for all editing tasks.

\noindent\textbf{MARM Initialization Details.}
To complement Section~\ref{4.3}, the selected source image and target image
are resized to $518\times518$ and encoded by the same frozen DINOv2 frontend
used by TRELLIS, yielding a $37\times37$ patch grid. MARM combines foreground
presence, DINOv2 correspondence, Lab appearance difference, and edge difference
into continuous Keep and Edit evidence. Target-only content additionally
produces Target-New evidence for opening editable space in source-empty regions.

A source-foreground patch provides reliable Keep evidence when its Keep score is
at least $0.55$ and exceeds its Edit score by at least $0.08$, and reliable Edit
evidence when its Edit score is at least $0.55$ and exceeds its Keep score by at
least $0.10$. Otherwise, the patch is marked Unknown at the evidence stage.
Unknown and Target-New are internal initialization labels only: Unknown denotes
insufficient evidence prior to region-level assignment, whereas Target-New
denotes target-side content without reliable source correspondence. They do not
introduce additional MARM region states; the formal MARM partition remains
binary Edit/Keep.

The complete 3D region partition is constructed from the four source views
rather than directly from single-target-image clustering. Source-occupied
Grid64 voxels aggregate visible DINOv2 features from the four rendered source
views. We first construct 16 provisional macro regions using deterministic
clustering over DINO cosine distance and 3D spatial distance, followed by
3D-connectivity cleanup, and consolidate them into $K=8$ adaptive regions.
Each adaptive region is projected to the selected-view Edit/Keep evidence map,
and region-level evidence aggregation determines the binary initial partition
$\mathbf{M}_0$.

For target-only content, Target-New evidence is grouped into 2D candidate
regions and projected through the calibrated selected-view camera into
source-empty Grid64 candidate space. This candidate space permits new structure
to emerge but does not directly enforce occupancy. Source-empty locations that
are explicitly retained as non-editable support form the valid empty set
$\mathbf{S}_e$ used by the Stage~1 SGPS empty-preservation branch.

\noindent\textbf{Inference Parameters.}
The main inference parameters are summarized in
Table~\ref{tab:inference_parameters}. All settings are fixed across datasets
and editing tasks.

\begin{table}[h]
\centering
\scriptsize
\setlength{\tabcolsep}{4.6pt}
\caption{Main inference parameters of UMAS3D.}
\label{tab:inference_parameters}
\begin{tabular}{lcc}
\toprule
Component & Parameter & Value \\
\midrule
TRELLIS
& Stage~1 / Stage~2 steps & $100 / 25$ \\
\midrule
MARM
& Adaptive regions $K$ & $8$ \\
& Provisional macro regions & $16$ \\
& EMA coefficient $\lambda$ & $0.8$ \\
& Initial warm-up & first $35$ Stage~1 steps \\
& Region update interval & every $5$ steps \\
& Confirmation window / requirement & $3 / 2$ \\
& Edit$\rightarrow$Keep margin $\tau^K$ & $0.04$ \\
& Keep$\rightarrow$Edit margin $\tau^E$ & $0.15$ \\
\midrule
Stage~1 SGPS
& Sample optimization steps / branch & $1$ \\
& Base sample learning rate & $1\times10^{-5}$ \\
& Occupied / Empty update scale & $0.25 / 1.0$ \\
& Empty constraint activation & $t\leq0.5$ \\
\midrule
Stage~2 SGPS
& Sample optimization steps & $1$ \\
& Sample learning rate & $1\times10^{-5}$ \\
& SmoothL1 $\beta$ & $1.0$ \\
\bottomrule
\end{tabular}
\end{table}

During the first 35 Stage~1 steps, the initial partition $\mathbf{M}_0$ is kept
fixed to allow the target structure to establish a stable trajectory.
Thereafter, MARM evaluates region states every five inference steps. To suppress
transient switching, a transition is confirmed only when consistent evidence is
observed in at least two of the most recent three refinement checkpoints. We use
a larger margin for Keep$\rightarrow$Edit transitions ($\tau^E=0.15$) than for
Edit$\rightarrow$Keep transitions ($\tau^K=0.04$), making the release of
preserved source regions more conservative. The final partition obtained at the
end of Stage~1 is frozen as $\mathbf{M}_T$, and its Keep support
$\mathbf{M}_T^K$ is passed to Stage~2.

\noindent\textbf{Sample-State Update and Sampling Order.}
Section~\ref{4.4} defines the Stage~1 objectives $\mathcal{L}_1^o$ and
$\mathcal{L}_1^e$ and the Stage~2 objective $\mathcal{L}_2$. Here, we specify
how these objectives are applied within the TRELLIS sampler.

Let $\mathbf{x}$ denote the current Stage~1 structure sample state and
$\mathbf{Z}$ the current Stage~2 SLAT sample state. $\mathcal{T}_1$ and
$\mathcal{T}_2$ denote the ordered sets of consecutive sampling-time pairs
$(t,t')$ in Stage~1 and Stage~2, respectively. $\mathcal{U}_{\mathrm{MARM}}$
denotes the Stage~1 refinement checkpoints after the initial warm-up.
$\mathbf{M}$ denotes the current Edit/Keep partition and $\mathbf{M}^K$ its
current Keep support.

We use $\hat{\mathbf{x}}_0$ and $\hat{\mathbf{Z}}_0$ to denote the predicted
clean Structure and SLAT states. Superscript $a$ denotes a source-aligned sample
state, while $\tilde{\mathbf{x}}$ and $\tilde{\mathbf{Z}}$ denote the
SGPS-corrected states before frozen-flow re-evaluation.
$\mathbf{G}_o$ and $\mathbf{G}_e$ denote the localized supports of the occupied
and empty Stage~1 preservation branches. $\mathcal{A}_T$ denotes the final
active voxel coordinate set produced by Stage~1.

The final Keep support $\mathbf{M}_T^K$ is mapped to the fixed Stage~2
Keep-token support $\mathbf{G}_K$ by exact sparse integer-coordinate
correspondence. A narrow Boundary-token support $\mathbf{G}_B$ is additionally
derived around the final Edit/Keep interface. $\mathbf{G}_B$ is used only for
soft source--target blending and does not constitute an additional MARM state.
Finally, $\boldsymbol{\epsilon}_{\mathrm{init}}$ denotes the fixed initial
Stage~2 noise used for source-aligned Keep-token construction.

In Stage~1, source alignment first produces $\mathbf{x}^{a}$. The occupied and
empty preservation branches then perform two independent Adam updates, both
initialized from the same $\mathbf{x}^{a}$:
\begin{equation}
\Delta\mathbf{x}^{o}
=
\operatorname{AdamStep}
(\mathbf{x}^{a},\mathcal{L}_1^o,\eta\alpha_o)
-\mathbf{x}^{a},
\qquad
\Delta\mathbf{x}^{e}
=
\operatorname{AdamStep}
(\mathbf{x}^{a},\mathcal{L}_1^e,\eta\alpha_e)
-\mathbf{x}^{a}.
\end{equation}
The corrected state is
\begin{equation}
\tilde{\mathbf{x}}
=
\mathbf{x}^{a}
+
\mathbf{G}_o\odot\Delta\mathbf{x}^{o}
+
\mathbf{G}_e\odot\Delta\mathbf{x}^{e},
\end{equation}
where $\eta=10^{-5}$, $\alpha_o=0.25$, and $\alpha_e=1.0$. Thus, the two BCE
branches produce independent localized sample-state corrections from the same
source-aligned state. The frozen TRELLIS Structure Flow is then re-evaluated on
$\tilde{\mathbf{x}}$, followed by the original Euler sampling update.

Stage~2 follows the same sample-state correction principle. The source-aligned
state $\mathbf{Z}^{a}$ is constructed using the source SLAT features and fixed
initial-noise endpoint in $\mathbf{G}_K$, while tokens in $\mathbf{G}_B$ are
softly blended between source-aligned and target-driven states. All remaining
tokens stay target-driven. The SmoothL1 source supervision is applied only to
the Keep-token support:
\begin{equation}
\Delta\mathbf{Z}
=
\operatorname{AdamStep}
(\mathbf{Z}^{a},\mathcal{L}_2,\eta)
-\mathbf{Z}^{a},
\qquad
\tilde{\mathbf{Z}}
=
\mathbf{Z}^{a}
+
\mathbf{G}_K\odot\Delta\mathbf{Z}.
\end{equation}
The frozen SLAT Flow is subsequently re-evaluated on
$\tilde{\mathbf{Z}}$, followed by the standard Euler update over all tokens.
Boundary and Edit tokens receive no SmoothL1 source-feature gradient, although
they continue to evolve through the normal target-conditioned SLAT Flow.
A final exact Source-SLAT composition is applied only to the Keep tokens before
decoding.

\begin{algorithm}[t]
\caption{Inference Procedure of UMAS3D}
\label{alg:umas3d_inference}
\small
\begin{algorithmic}[1]

\Require Source 3D $\mathbf{X}_s$; four source views and calibrated cameras
$\{\mathbf{I}_s^v,\Pi_v\}_{v=1}^{V}$ with $V=4$; selected source image
$\mathbf{I}_s^{v^*}$; target image $\mathbf{I}_t$; frozen TRELLIS
\Ensure Edited 3D $\mathbf{X}_e$

\State $\mathbf{M}_0 \gets
\Call{InitializeMARM}
{\mathbf{X}_s,\{\mathbf{I}_s^v,\Pi_v\}_{v=1}^{V},
\mathbf{I}_s^{v^*},\mathbf{I}_t}$

\State $\mathbf{M}\gets\mathbf{M}_0$
\State $\mathbf{x}\gets\Call{InitStructureState}{}$

\For{$(t,t')\in\mathcal{T}_1$}

    \State $\hat{\mathbf{x}}_0
    \gets
    \Call{StructurePredict}
    {\mathbf{x},t,\mathbf{I}_t}$

    \If{$t\in\mathcal{U}_{\mathrm{MARM}}$}
        \State $\mathbf{M}
        \gets
        \Call{MARMUpdate}
        {\mathbf{M},\hat{\mathbf{x}}_0,\mathbf{X}_s}$
    \EndIf

    \State $\mathbf{x}^{a}
    \gets
    \Call{SourceAlign}
    {\mathbf{x},\hat{\mathbf{x}}_0,\mathbf{X}_s,\mathbf{M}^{K}}$

    \State $(\mathcal{L}_1^o,\mathcal{L}_1^e,\mathbf{G}_o,\mathbf{G}_e)
    \gets
    \Call{Stage1Loss}
    {\mathbf{x}^{a},\mathbf{X}_s,\mathbf{M}^{K}}$

    \State $\Delta\mathbf{x}^{o}
    \gets
    \Call{AdamStep}
    {\mathbf{x}^{a},\mathcal{L}_1^o,\eta\alpha_o}
    -\mathbf{x}^{a}$

    \State $\Delta\mathbf{x}^{e}
    \gets
    \Call{AdamStep}
    {\mathbf{x}^{a},\mathcal{L}_1^e,\eta\alpha_e}
    -\mathbf{x}^{a}$

    \State $\tilde{\mathbf{x}}
    \gets
    \mathbf{x}^{a}
    +\mathbf{G}_o\odot\Delta\mathbf{x}^{o}
    +\mathbf{G}_e\odot\Delta\mathbf{x}^{e}$

    \State $\mathbf{v}
    \gets
    \Call{StructureFlow}
    {\tilde{\mathbf{x}},t,\mathbf{I}_t}$

    \State $\mathbf{x}
    \gets
    \tilde{\mathbf{x}}-(t-t')\mathbf{v}$

\EndFor

\State $\mathbf{M}_T\gets\Call{Freeze}{\mathbf{M}}$

\State $\mathcal{A}_T
\gets
\Call{Stage1Finalize}
{\mathbf{x},\mathbf{X}_s,\mathbf{M}_T^K}$

\State $\mathbf{Z}_s
\gets
\Call{EncodeSourceSLAT}{\mathbf{X}_s}$

\State $(\mathbf{Z},\boldsymbol{\epsilon}_{\mathrm{init}})
\gets
\Call{InitSLATState}{\mathcal{A}_T}$

\State $\mathbf{G}_K
\gets
\Call{KeepTokenSupport}
{\mathbf{M}_T^K,\mathcal{A}_T}$

\State $\mathbf{G}_B
\gets
\Call{BoundaryTokenSupport}
{\mathbf{M}_T,\mathcal{A}_T}$

\For{$(t,t')\in\mathcal{T}_2$}

    \State $\hat{\mathbf{Z}}_0
    \gets
    \Call{SLATPredict}
    {\mathbf{Z},t,\mathbf{I}_t}$

    \State $\mathbf{Z}^{a}
    \gets
    \Call{SourceAlignSLAT}
    {\mathbf{Z},\hat{\mathbf{Z}}_0,\mathbf{Z}_s,
    \boldsymbol{\epsilon}_{\mathrm{init}},\mathbf{G}_K,\mathbf{G}_B}$

    \State $\mathcal{L}_2
    \gets
    \Call{KeepSmoothL1}
    {\mathbf{Z}^{a},\mathbf{Z}_s,\mathbf{G}_K}$

    \State $\Delta\mathbf{Z}
    \gets
    \Call{AdamStep}
    {\mathbf{Z}^{a},\mathcal{L}_2,\eta}
    -\mathbf{Z}^{a}$

    \State $\tilde{\mathbf{Z}}
    \gets
    \mathbf{Z}^{a}
    +\mathbf{G}_K\odot\Delta\mathbf{Z}$

    \State $\mathbf{v}
    \gets
    \Call{SLATFlow}
    {\tilde{\mathbf{Z}},t,\mathbf{I}_t}$

    \State $\mathbf{Z}
    \gets
    \tilde{\mathbf{Z}}-(t-t')\mathbf{v}$

\EndFor

\State $\mathbf{Z}_e
\gets
\Call{SourceCompose}
{\mathbf{Z},\mathbf{Z}_s,\mathbf{G}_K}$

\State $\mathbf{X}_e
\gets
\Call{Decode}{\mathbf{Z}_e}$

\State \Return $\mathbf{X}_e$

\end{algorithmic}
\end{algorithm}

The entire procedure uses a single Stage~1 trajectory followed by a single
Stage~2 trajectory, without an additional TRELLIS proposal pass.

\noindent\textbf{Computational Cost.}
On 50 samples, with 10 samples from each editing task, using an NVIDIA
A100-SXM4-40GB GPU and batch size 1, UMAS3D requires 245.80\,s per sample on
average and 11.55\,GiB peak allocated GPU memory (16.93\,GiB peak reserved
memory). Runtime is measured from target-image preprocessing to final edited
3D GLB export, including mask-free region discovery, Stage~1 and Stage~2
generation, final decoding, and mesh/texture/GLB export, while excluding model
and source-SLAT loading, four-view rendering, and GT evaluation.

\subsection{Evaluation Metrics}
\label{sec:appendix_evaluation_metrics}

We evaluate the final edited 3D results from two complementary aspects:
\textit{Unedited-region Preservation}, using Pres-IoU, Pres-PSNR, and
Pres-SSIM, and \textit{Overall Target Quality}, using DINO-I, CD-L1,
F@1\%, and NC.

\paragraph{Unedited-region Preservation.}

\textbf{Pres-IoU.}
Let $\mathcal{O}_s$ and $\hat{\mathcal{O}}$ denote the surface-occupancy
supports of the source asset $\mathbf{X}_s$ and prediction $\hat{\mathbf{X}}$
on a common $64^3$ Grid64. Let $\mathbf{M}_{\mathrm{gt}}^K$ denote the
evaluation Keep region derived from the aligned source and benchmark target.
The preserved surface supports are
\begin{equation}
\mathcal{O}_s^K
=
\mathcal{O}_s\cap\mathbf{M}_{\mathrm{gt}}^K,
\qquad
\hat{\mathcal{O}}^K
=
\hat{\mathcal{O}}\cap\mathbf{M}_{\mathrm{gt}}^K.
\end{equation}
Pres-IoU is defined as
\begin{equation}
\mathrm{Pres\text{-}IoU}
=
\frac{
|\mathcal{O}_s^K\cap\hat{\mathcal{O}}^K|
}{
|\mathcal{O}_s^K\cup\hat{\mathcal{O}}^K|
}.
\end{equation}
A higher value indicates better preservation of source geometry in the
unedited region.

\textbf{Pres-PSNR.}
For canonical view $v$, let $\mathbf{I}_s^v$ and
$\hat{\mathbf{I}}^v$ denote the source and predicted renderings, respectively,
and let $\mathbf{M}_{\mathrm{gt}}^{K,v}$ denote the corresponding evaluation
Keep mask. Images are normalized to $[0,1]$. The Keep-region MSE is
\begin{equation}
\mathrm{MSE}_K^v
=
\frac{1}{
3|\mathbf{M}_{\mathrm{gt}}^{K,v}|
}
\sum_{p\in\mathbf{M}_{\mathrm{gt}}^{K,v}}
\sum_{c=1}^{3}
\left(
\mathbf{I}_{s,c}^v(p)
-
\hat{\mathbf{I}}_c^v(p)
\right)^2,
\end{equation}
and
\begin{equation}
\mathrm{PSNR}_K^v
=
10\log_{10}
\left(
\frac{1}{
\max(\mathrm{MSE}_K^v,\epsilon)
}
\right),
\qquad
\epsilon=10^{-12}.
\end{equation}
Let
\begin{equation}
\mathcal{V}_K
=
\left\{
v\mid
\mathbf{M}_{\mathrm{gt}}^{K,v}\neq\varnothing
\right\}
\end{equation}
denote the valid Keep views. We compute
\begin{equation}
\mathrm{Pres\text{-}PSNR}
=
\frac{1}{|\mathcal{V}_K|}
\sum_{v\in\mathcal{V}_K}
\mathrm{PSNR}_K^v.
\end{equation}

\textbf{Pres-SSIM.}
We first compute the standard SSIM map between the complete source and predicted
views:
\begin{equation}
\mathbf{S}^v(p)
=
\operatorname{SSIMMap}
\left(
\mathbf{I}_s^v,
\hat{\mathbf{I}}^v
\right)_p,
\end{equation}
where the standard constants for $[0,1]$ images are
\begin{equation}
C_1=(0.01)^2,
\qquad
C_2=(0.03)^2.
\end{equation}
To reduce contamination from SSIM windows crossing the Edit/Keep boundary, the
evaluation Keep mask is slightly eroded before averaging. Let
$\widetilde{\mathbf{M}}_{\mathrm{gt}}^{K,v}$ denote the eroded Keep mask. Then
\begin{equation}
\mathrm{SSIM}_K^v
=
\frac{
\sum_p
\widetilde{\mathbf{M}}_{\mathrm{gt}}^{K,v}(p)
\mathbf{S}^v(p)
}{
\sum_p
\widetilde{\mathbf{M}}_{\mathrm{gt}}^{K,v}(p)
},
\end{equation}
and
\begin{equation}
\mathrm{Pres\text{-}SSIM}
=
\frac{1}{|\mathcal{V}_K|}
\sum_{v\in\mathcal{V}_K}
\mathrm{SSIM}_K^v.
\end{equation}
A higher Pres-SSIM indicates better structural and visual preservation in the
unedited region.

\paragraph{Overall Target Quality.}

\textbf{DINO-I.}
Let $\mathbf{I}_{\mathrm{gt}}^v$ denote the benchmark-target rendering at
canonical view $v$. Using the same frozen DINOv2 backbone $\phi$, let
$g_\phi(\cdot)$ denote its image-level representation. We compute
\begin{equation}
\mathrm{DINO\text{-}I}
=
\frac{1}{V}
\sum_{v=1}^{V}
\cos
\left(
g_\phi(\hat{\mathbf{I}}^v),
g_\phi(\mathbf{I}_{\mathrm{gt}}^v)
\right),
\qquad
V=4.
\end{equation}
This metric is evaluated over the complete rendered object. A higher value
indicates stronger visual-semantic consistency with the benchmark target.

\textbf{CD-L1.}
Let $\mathcal{Q}_{\hat{X}}$ and $\mathcal{Q}_{\mathrm{gt}}$ denote surface
point sets sampled from $\hat{\mathbf{X}}$ and
$\mathbf{X}_{\mathrm{gt}}$, respectively. We use the symmetric Chamfer
distance
\begin{equation}
\mathrm{CD\text{-}L1}
=
\frac{1}{2}
\left[
\frac{1}{|\mathcal{Q}_{\hat{X}}|}
\sum_{\hat q\in\mathcal{Q}_{\hat{X}}}
\min_{q\in\mathcal{Q}_{\mathrm{gt}}}
\|\hat q-q\|_2
+
\frac{1}{|\mathcal{Q}_{\mathrm{gt}}|}
\sum_{q\in\mathcal{Q}_{\mathrm{gt}}}
\min_{\hat q\in\mathcal{Q}_{\hat{X}}}
\|q-\hat q\|_2
\right].
\end{equation}
A lower value indicates better overall geometric consistency with the
benchmark target.

\textbf{F@1\%.}
Let $D_{\mathrm{gt}}$ denote the 3D bounding-box diagonal of the benchmark
target and define
\begin{equation}
\tau=0.01D_{\mathrm{gt}}.
\end{equation}
The surface precision and recall are
\begin{equation}
\mathrm{Prec}_{\tau}
=
\frac{
\left|
\left\{
\hat q\in\mathcal{Q}_{\hat{X}}:
\min_{q\in\mathcal{Q}_{\mathrm{gt}}}
\|\hat q-q\|_2\le\tau
\right\}
\right|
}{
|\mathcal{Q}_{\hat{X}}|
},
\end{equation}
and
\begin{equation}
\mathrm{Rec}_{\tau}
=
\frac{
\left|
\left\{
q\in\mathcal{Q}_{\mathrm{gt}}:
\min_{\hat q\in\mathcal{Q}_{\hat{X}}}
\|q-\hat q\|_2\le\tau
\right\}
\right|
}{
|\mathcal{Q}_{\mathrm{gt}}|
}.
\end{equation}
We then compute
\begin{equation}
\mathrm{F@1\%}
=
\frac{
2\,\mathrm{Prec}_{\tau}\mathrm{Rec}_{\tau}
}{
\mathrm{Prec}_{\tau}+\mathrm{Rec}_{\tau}
}.
\end{equation}

\textbf{NC.}
Normal Consistency measures agreement between the predicted and benchmark-target
surface normals under bidirectional nearest-neighbor correspondence. Define
\begin{equation}
\pi_{\mathrm{gt}}(\hat q)
=
\arg\min_{q\in\mathcal{Q}_{\mathrm{gt}}}
\|\hat q-q\|_2,
\qquad
\pi_{\mathrm{pred}}(q)
=
\arg\min_{\hat q\in\mathcal{Q}_{\hat{X}}}
\|q-\hat q\|_2.
\end{equation}
Let $\hat{\mathbf n}(\hat q)$ and
$\mathbf n_{\mathrm{gt}}(q)$ denote the corresponding unit normals. We compute
\begin{equation}
\mathrm{NC}
=
\frac{1}{2}
\left[
\frac{1}{|\mathcal{Q}_{\hat{X}}|}
\sum_{\hat q\in\mathcal{Q}_{\hat{X}}}
\left|
\hat{\mathbf n}(\hat q)^\top
\mathbf n_{\mathrm{gt}}
\bigl(\pi_{\mathrm{gt}}(\hat q)\bigr)
\right|
+
\frac{1}{|\mathcal{Q}_{\mathrm{gt}}|}
\sum_{q\in\mathcal{Q}_{\mathrm{gt}}}
\left|
\mathbf n_{\mathrm{gt}}(q)^\top
\hat{\mathbf n}
\bigl(\pi_{\mathrm{pred}}(q)\bigr)
\right|
\right].
\end{equation}
A higher NC indicates better local surface-orientation consistency with the
benchmark target.

The evaluation Keep regions
$\mathbf{M}_{\mathrm{gt}}^K$ and
$\mathbf{M}_{\mathrm{gt}}^{K,v}$ are used only for metric computation and are
independent of the Edit/Keep partition estimated by MARM during inference. If a
sample contains no valid Keep region for a preservation metric, that metric is
excluded for the sample rather than assigned an artificial value. Dataset-level
scores are therefore averaged independently over the valid samples of each
metric.

\subsection{Qualitative 3D Editing Results}

\begin{figure*}[t]
\centering

\setlength{\tabcolsep}{0pt}
\renewcommand{\arraystretch}{1.0}

\newcommand{\aEightFinalImg}[1]{%
    \includegraphics[
        width=0.112\textwidth,
        height=1.55cm,
        keepaspectratio
    ]{#1}%
}

%
\newcommand{\aEightFinalRow}[8]{%
\begin{tikzpicture}[baseline=(current bounding box.center)]

\node[
    draw=black!70,
    dashed,
    line width=0.55pt,
    rounded corners=5pt,
    inner xsep=4pt,
    inner ysep=4pt
] (rowbox) {%
    \begin{minipage}{0.975\textwidth}
        \centering

        \begin{tabular*}{\linewidth}{
            @{\extracolsep{\fill}}
            c c c c c c c c
            @{}
        }

        \aEightFinalImg{#1}
        &
        \aEightFinalImg{#2}
        &
        \aEightFinalImg{#3}
        &
        \aEightFinalImg{#4}
        &
        \aEightFinalImg{#5}
        &
        \aEightFinalImg{#6}
        &
        \aEightFinalImg{#7}
        &
        \aEightFinalImg{#8}

        \end{tabular*}

    \end{minipage}%
};

\draw[
    black!60,
    dashed,
    line width=0.55pt
]
(rowbox.north) -- (rowbox.south);

\end{tikzpicture}%
}

\newcommand{\aEightFinalHeader}{%
\begin{minipage}{0.975\textwidth}
    \centering
    \scriptsize

    \makebox[0.48\linewidth][c]{\textbf{Source 3D}}%
    \hfill
    \makebox[0.48\linewidth][c]{\textbf{Edited 3D}}%

\end{minipage}%
}

\begin{tabular}{@{}c@{}}

\aEightFinalHeader
\\[1.5mm]

\aEightFinalRow
    {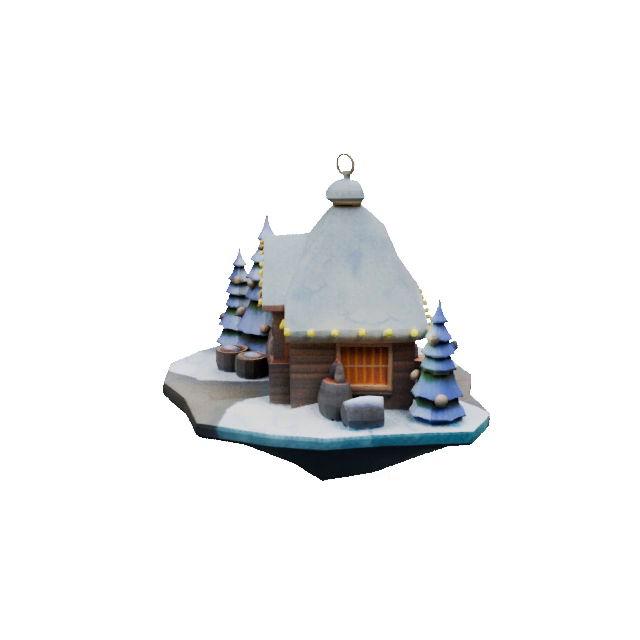}
    {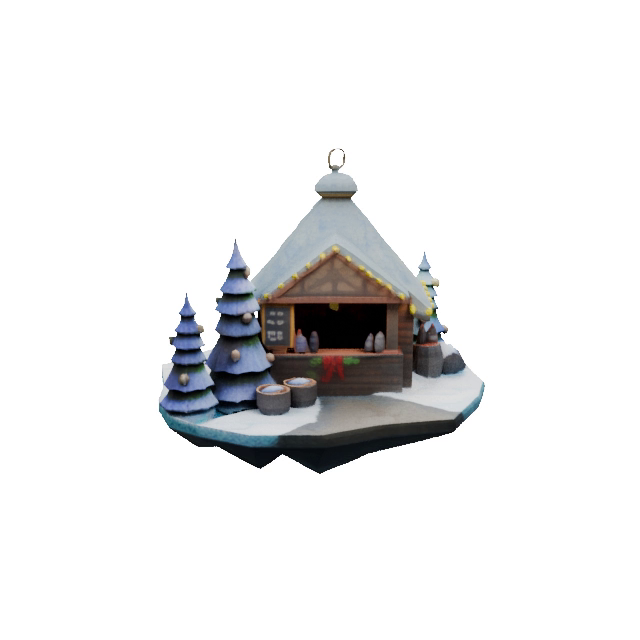}
    {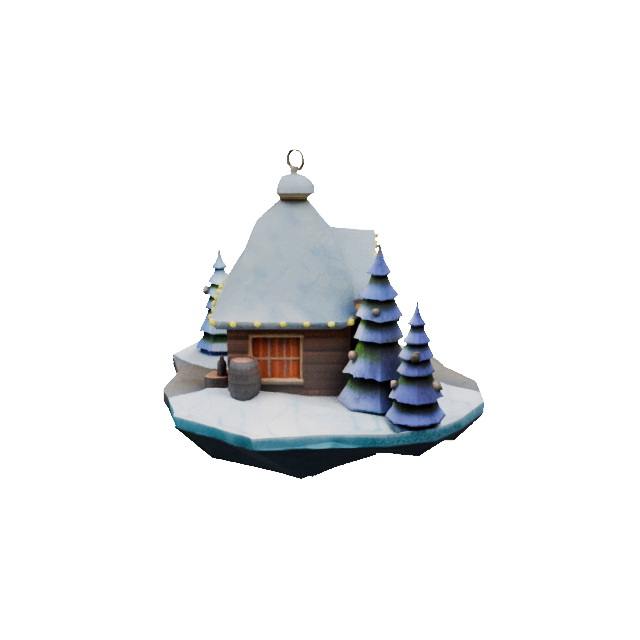}
    {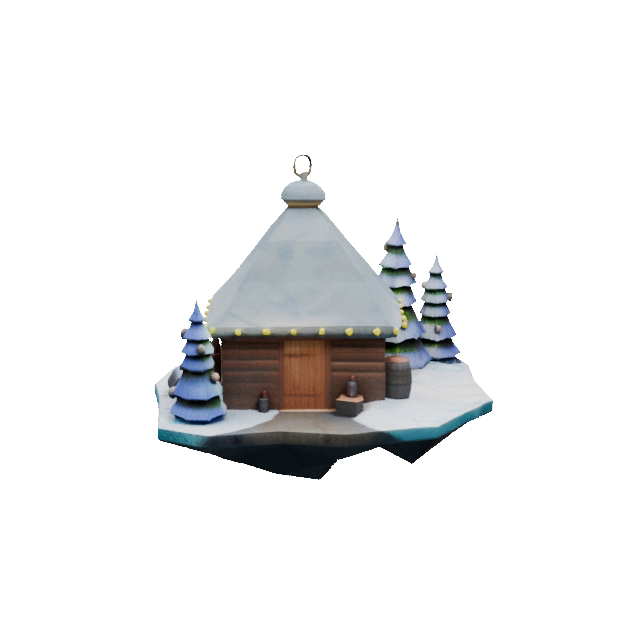}
    {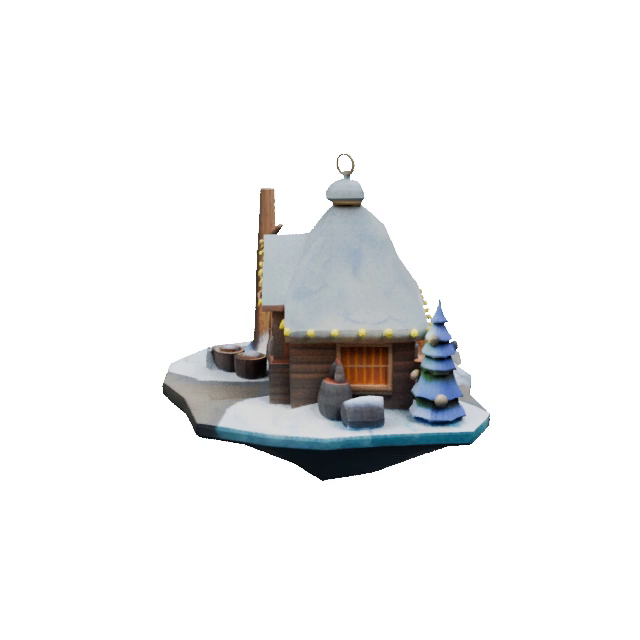}
    {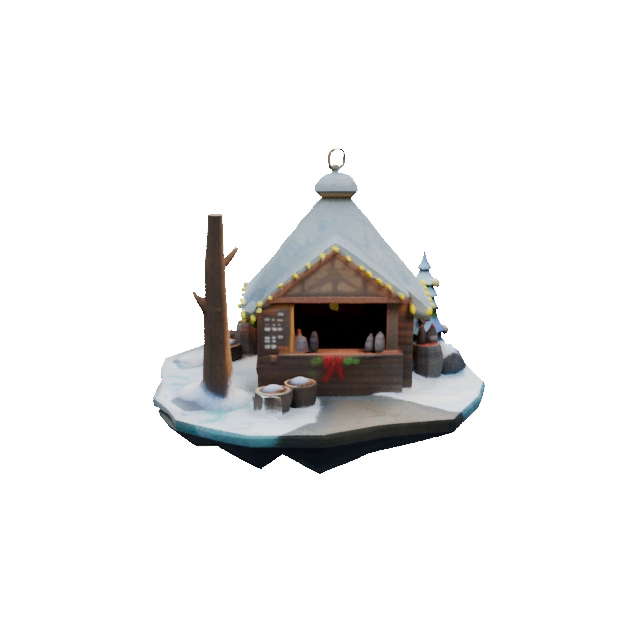}
    {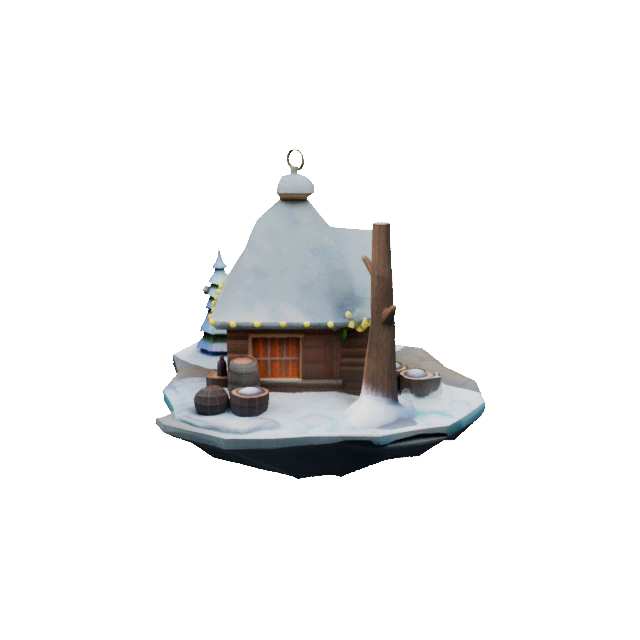}
    {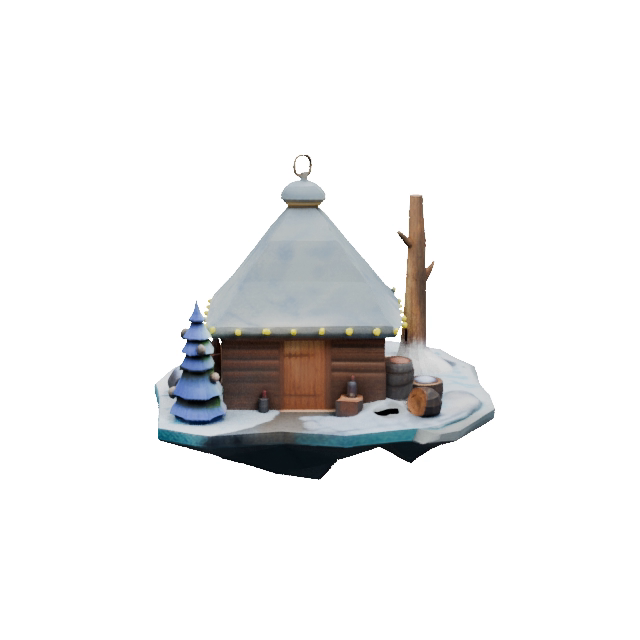}
\\[9mm]

\aEightFinalRow
    {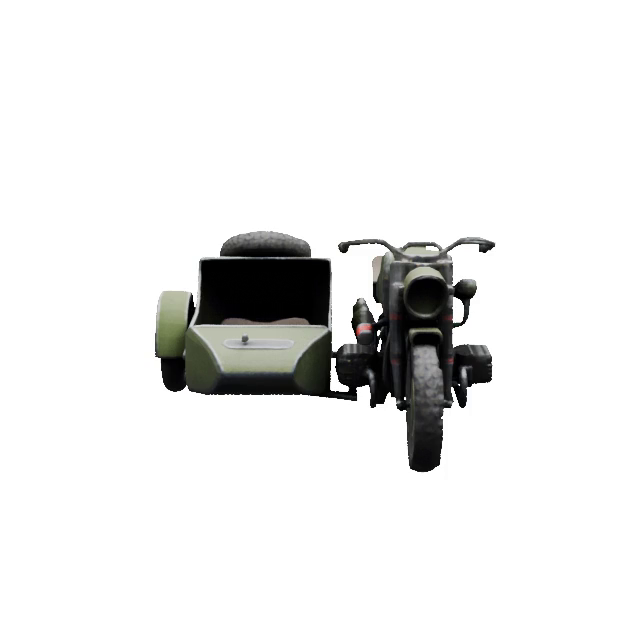}
    {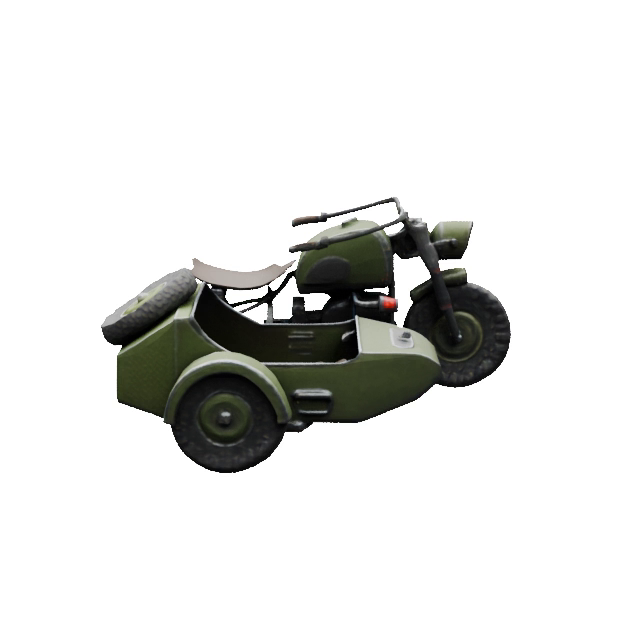}
    {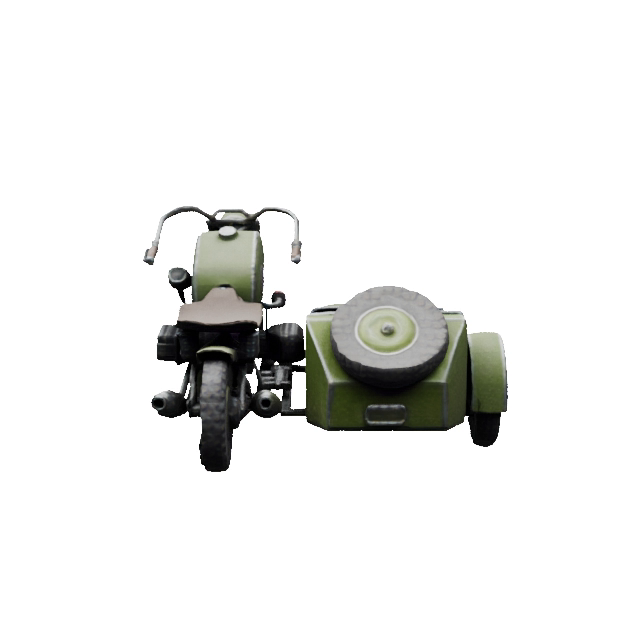}
    {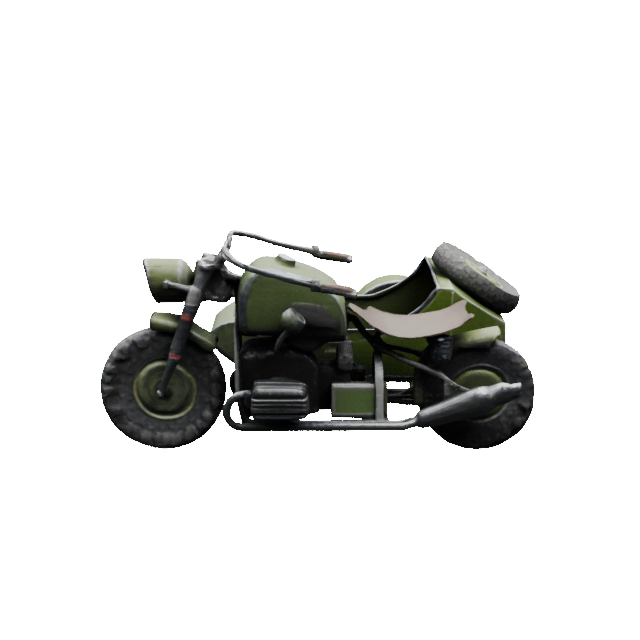}
    {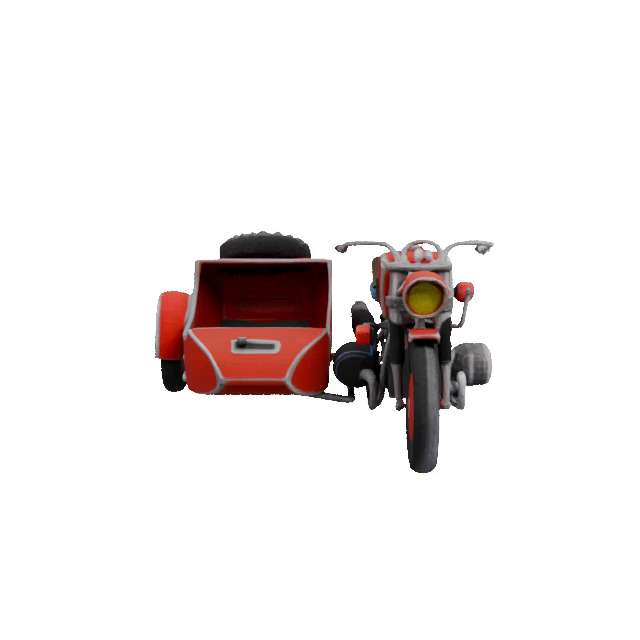}
    {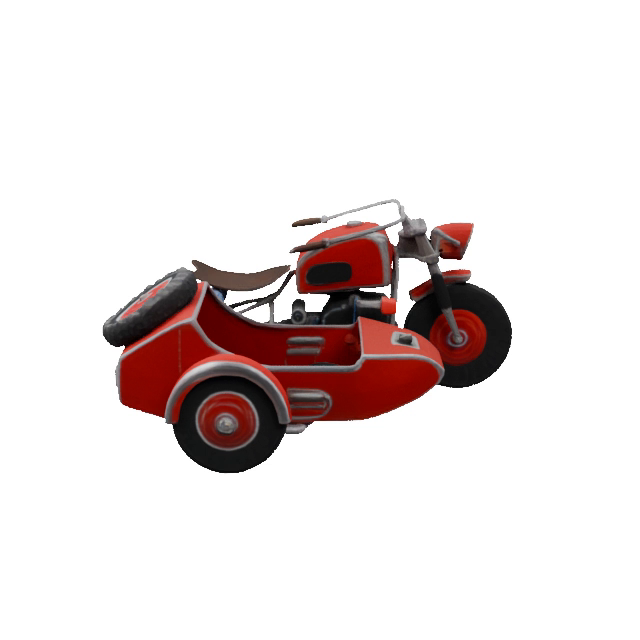}
    {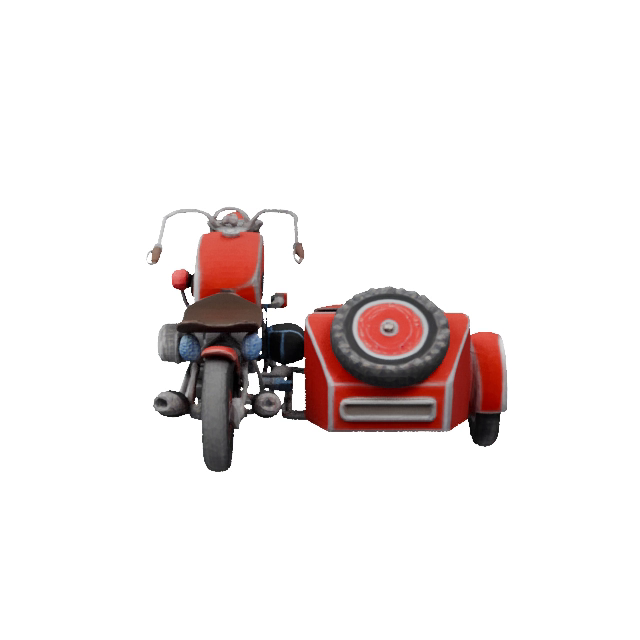}
    {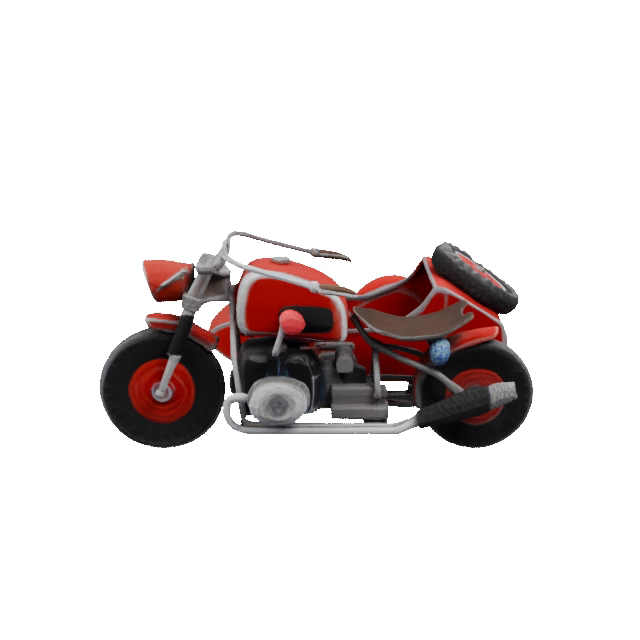}
\\[9mm]

\aEightFinalRow
    {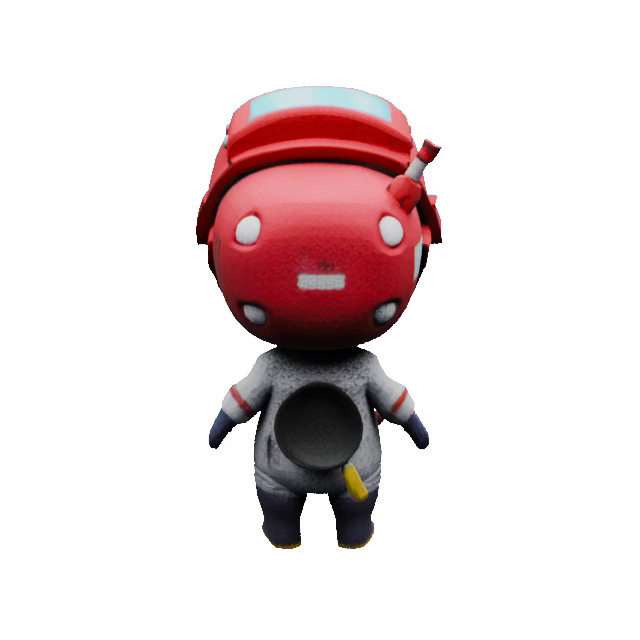}
    {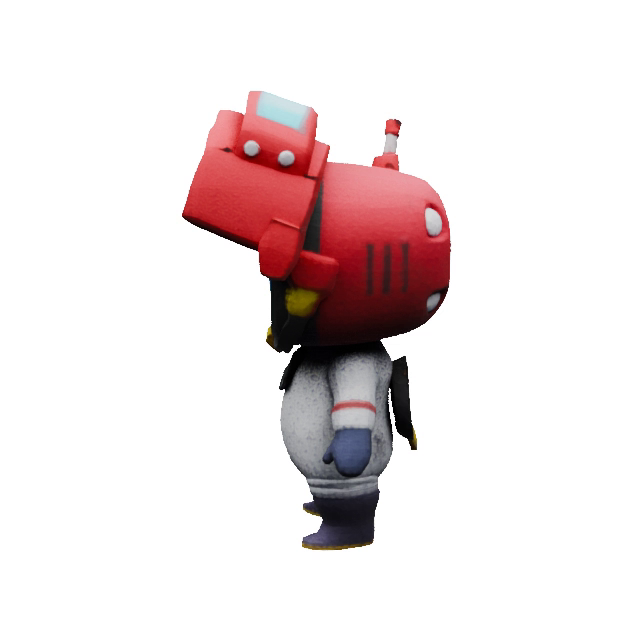}
    {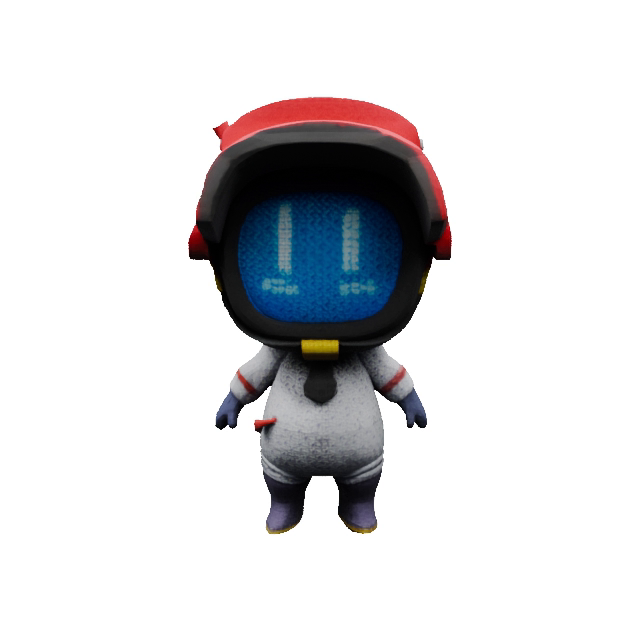}
    {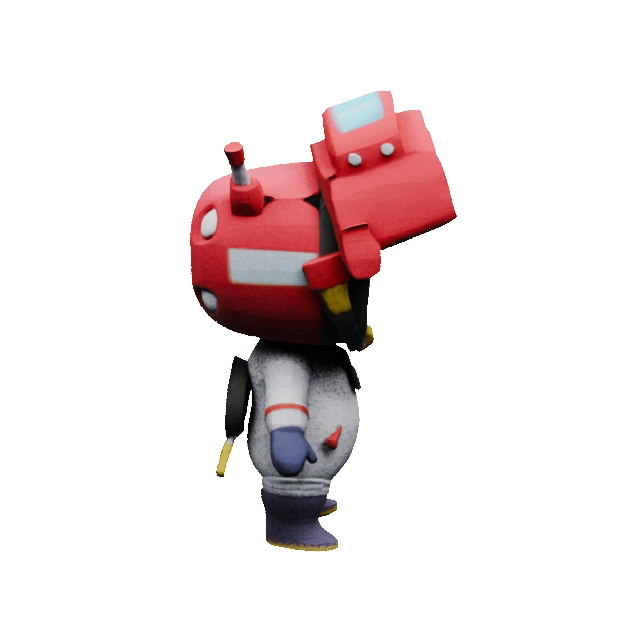}
    {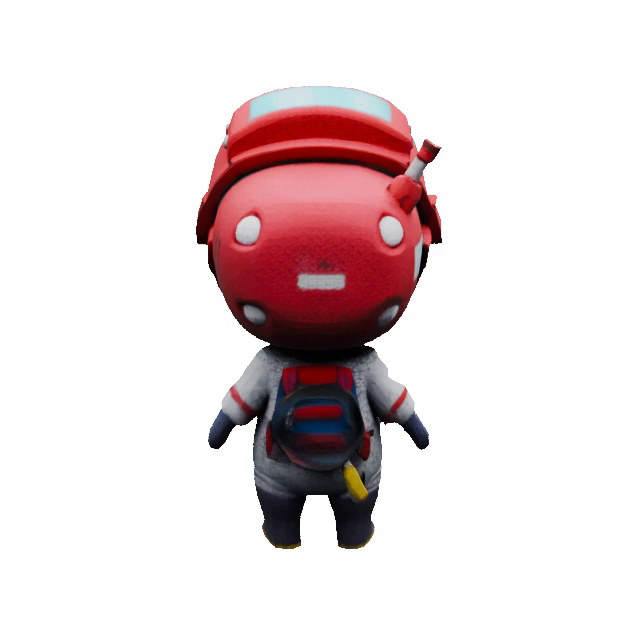}
    {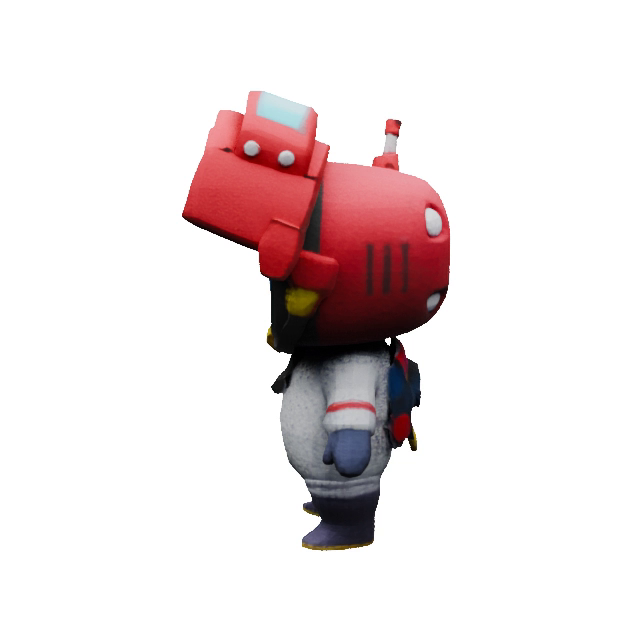}
    {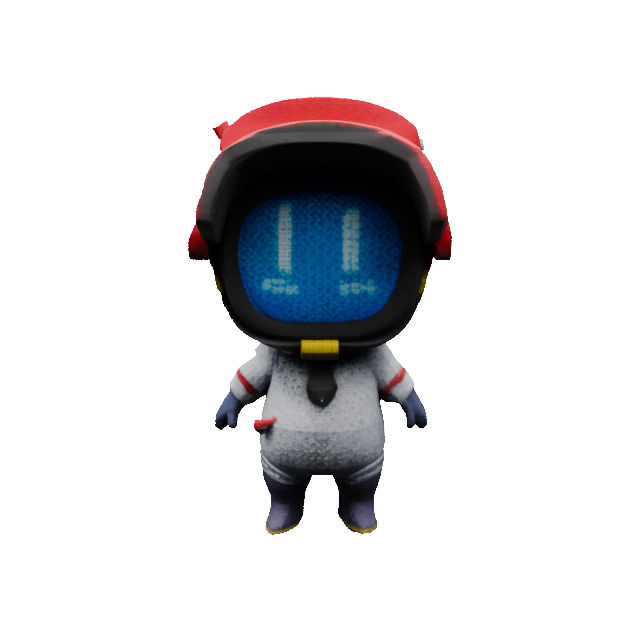}
    {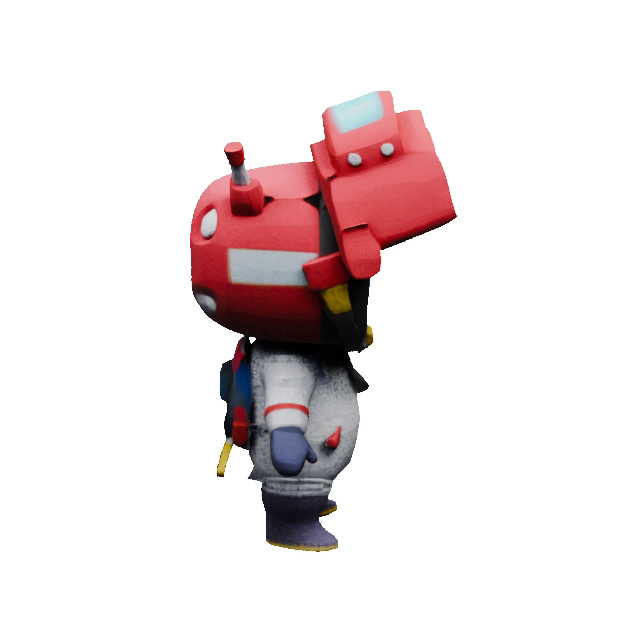}
\\[9mm]

\aEightFinalRow
    {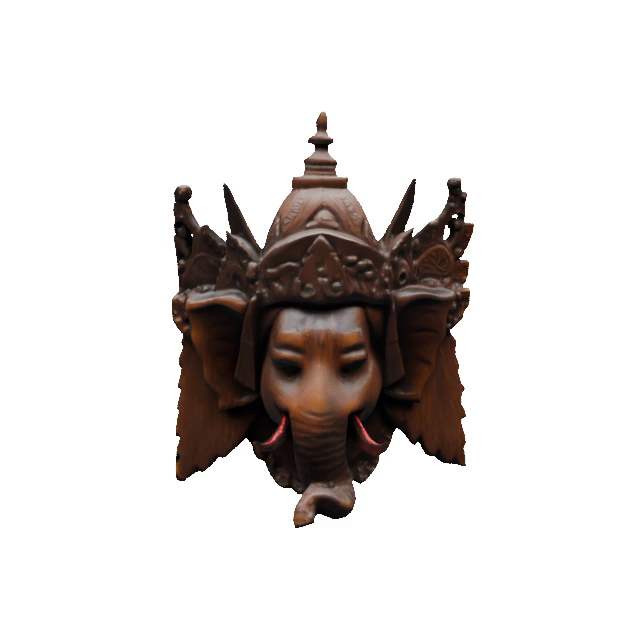}
    {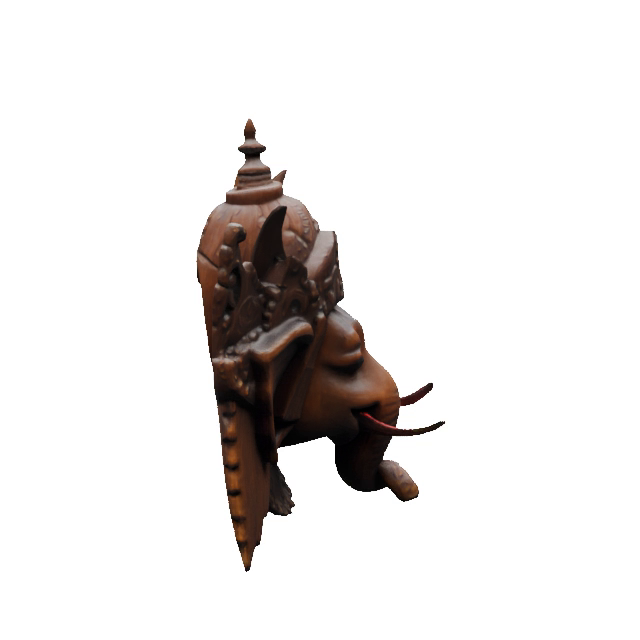}
    {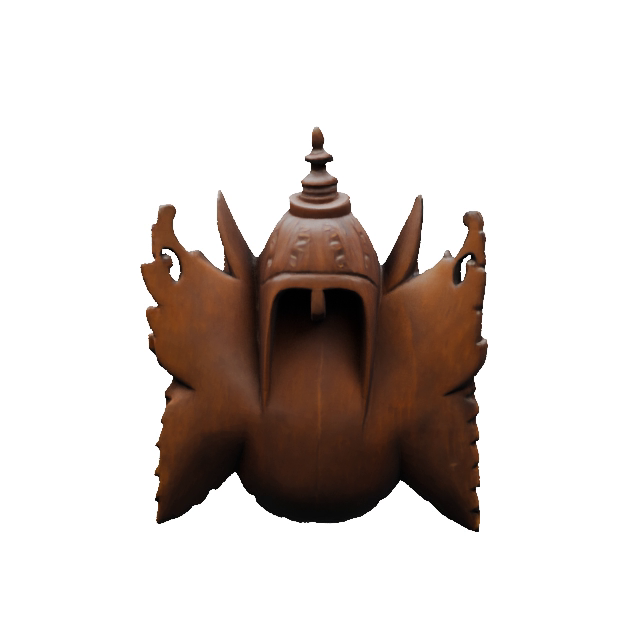}
    {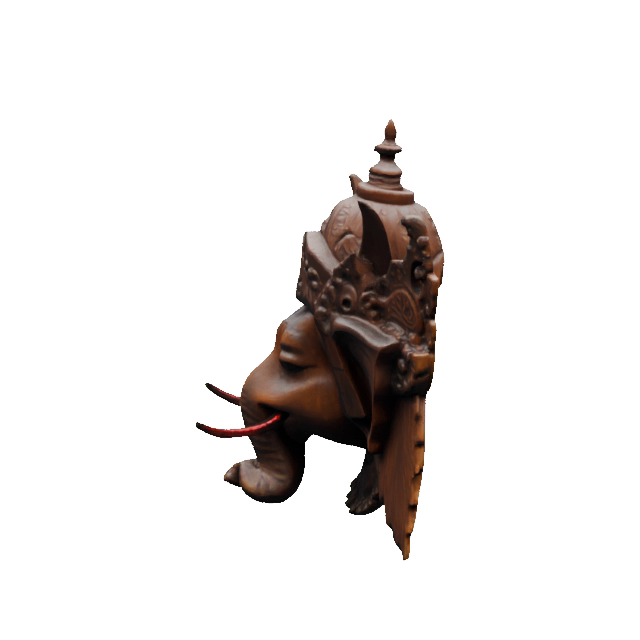}
    {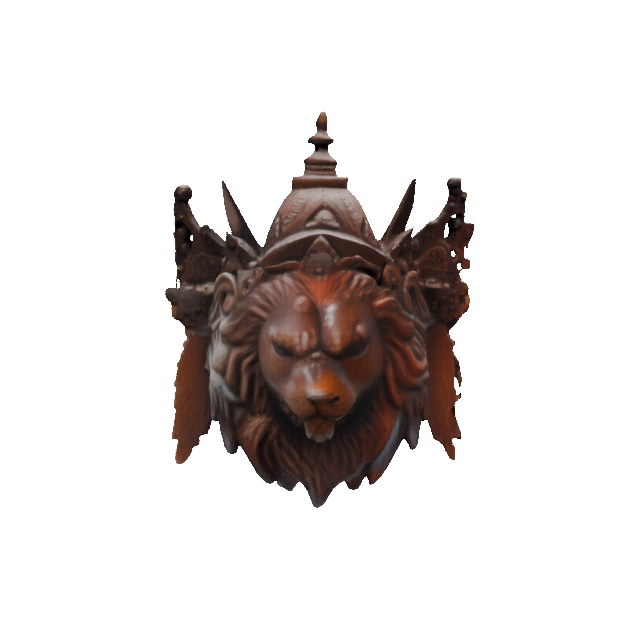}
    {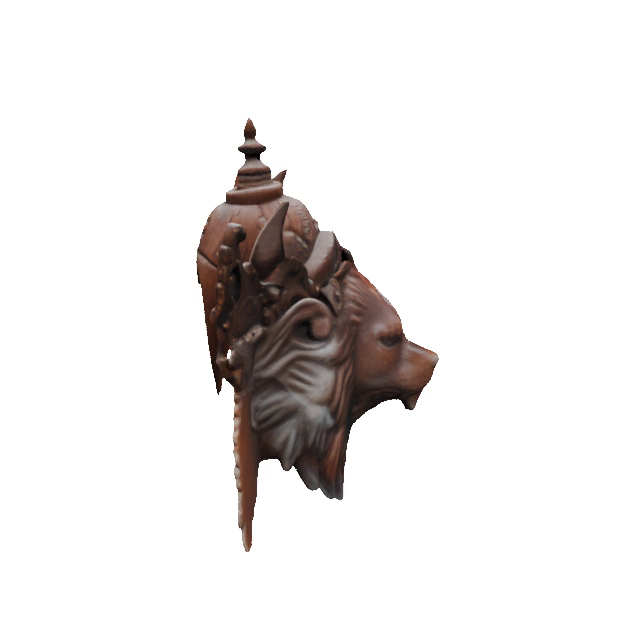}
    {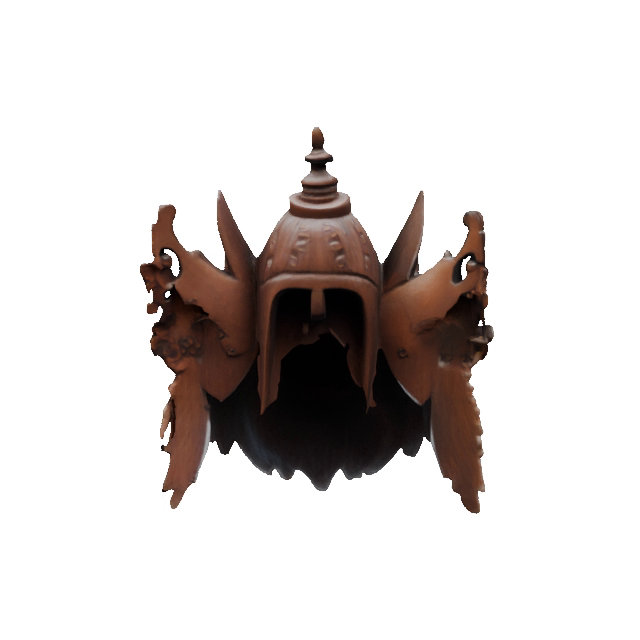}
    {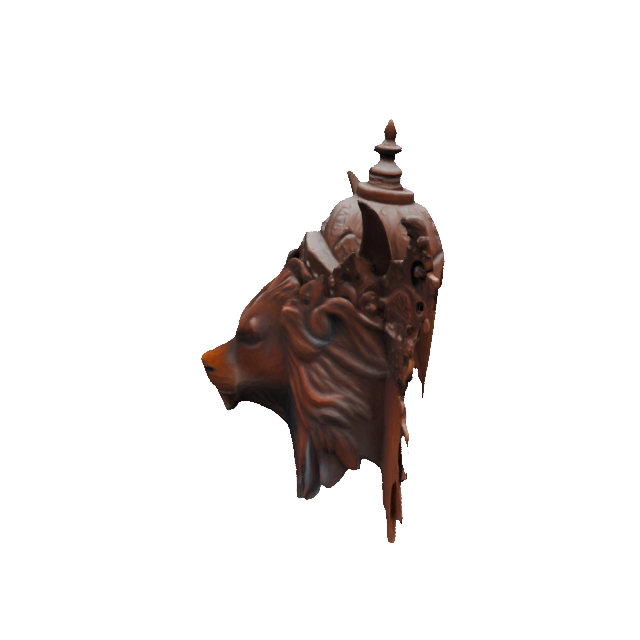}
\\[9mm]

\aEightFinalRow
    {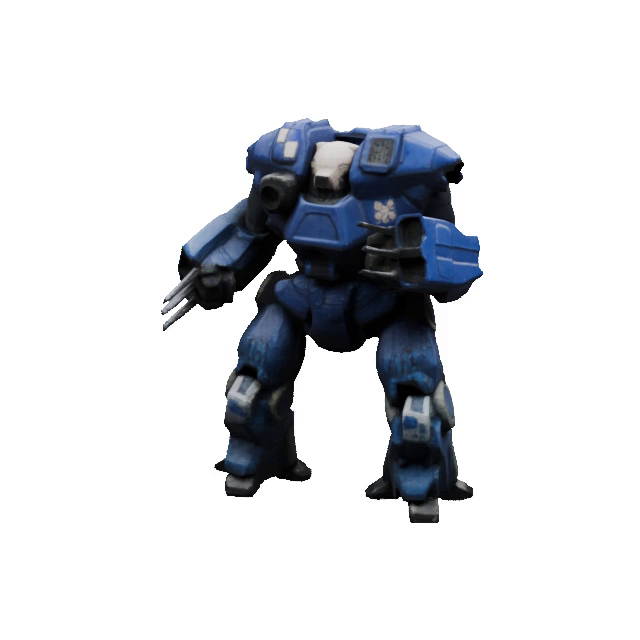}
    {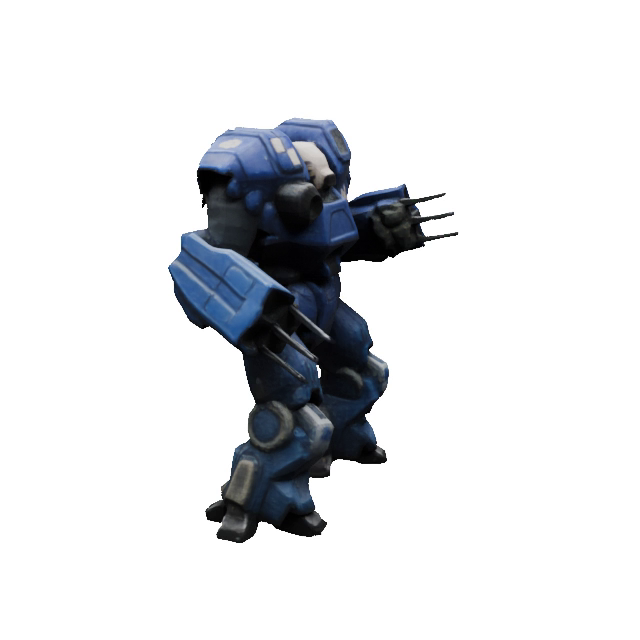}
    {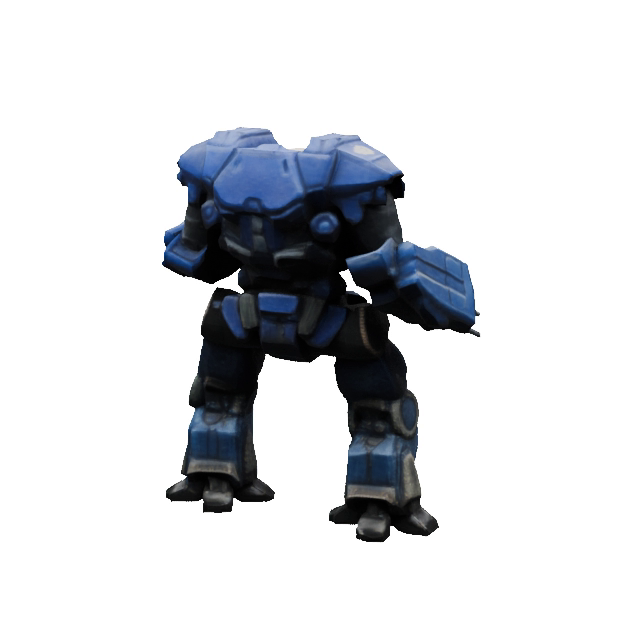}
    {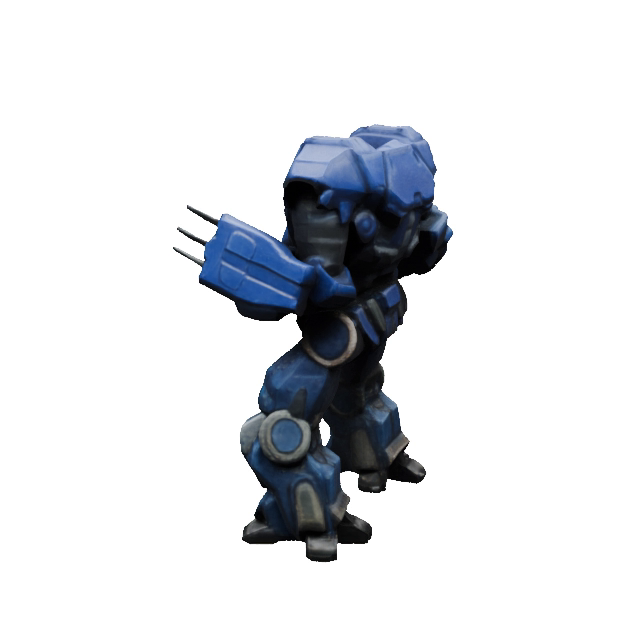}
    {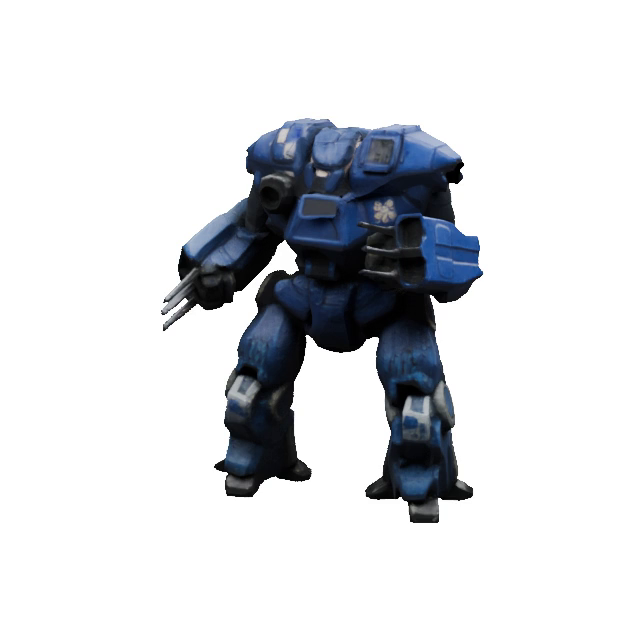}
    {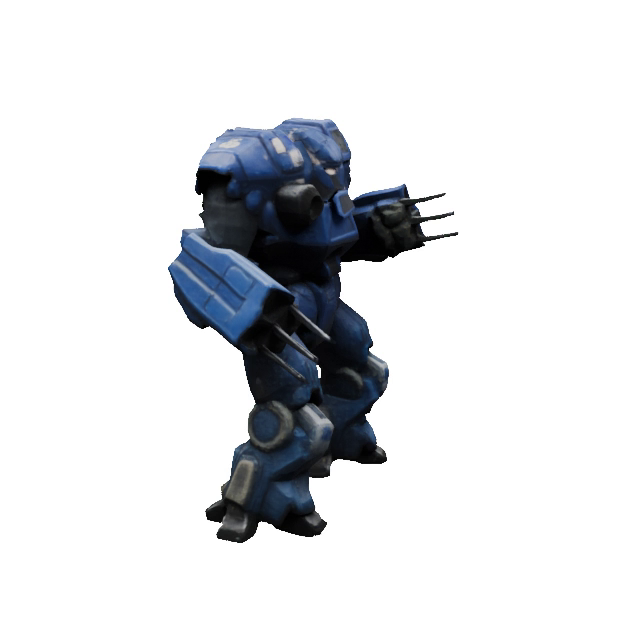}
    {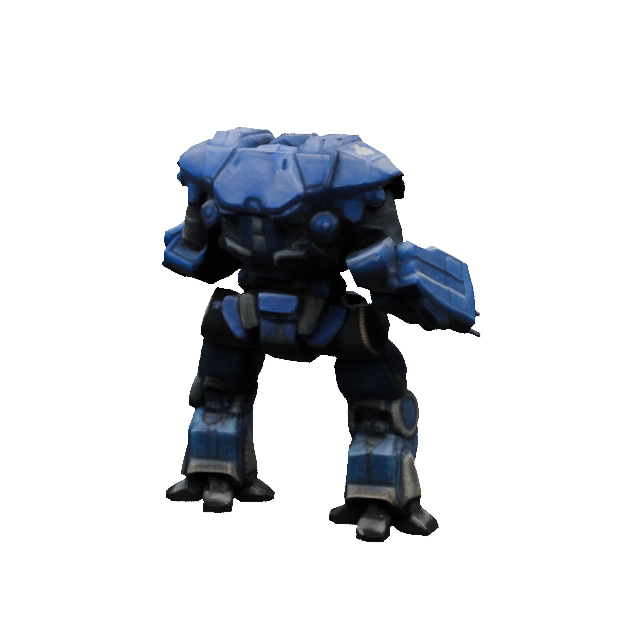}
    {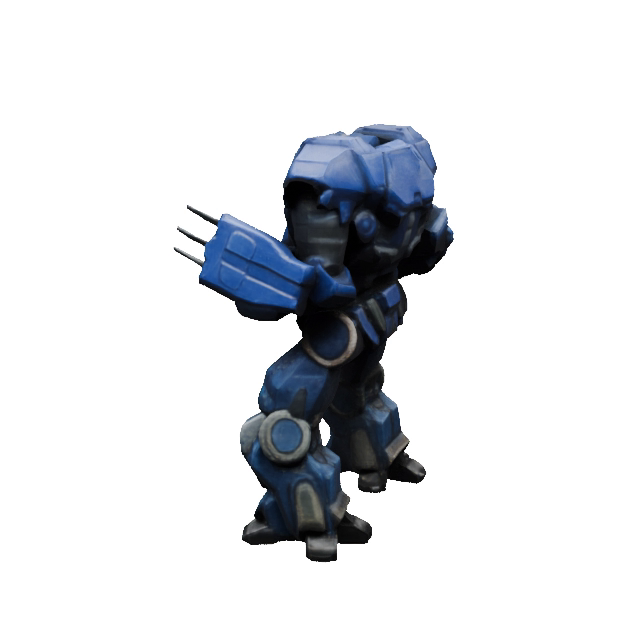}
\\[9mm]


\aEightFinalRow
    {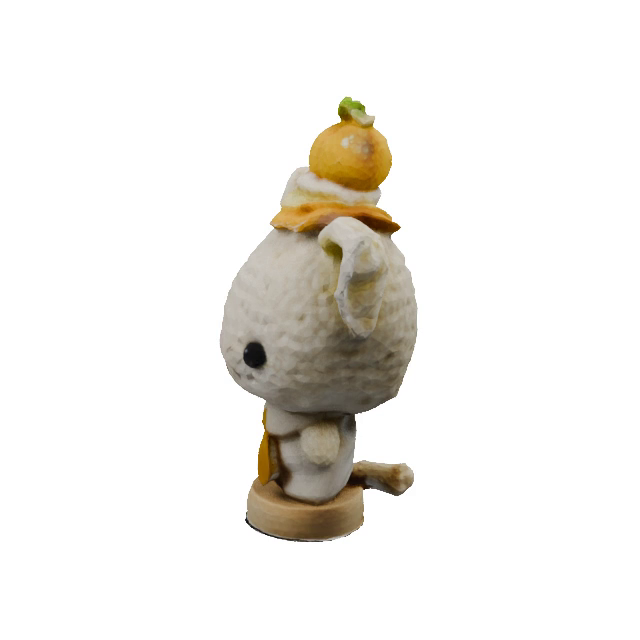}
    {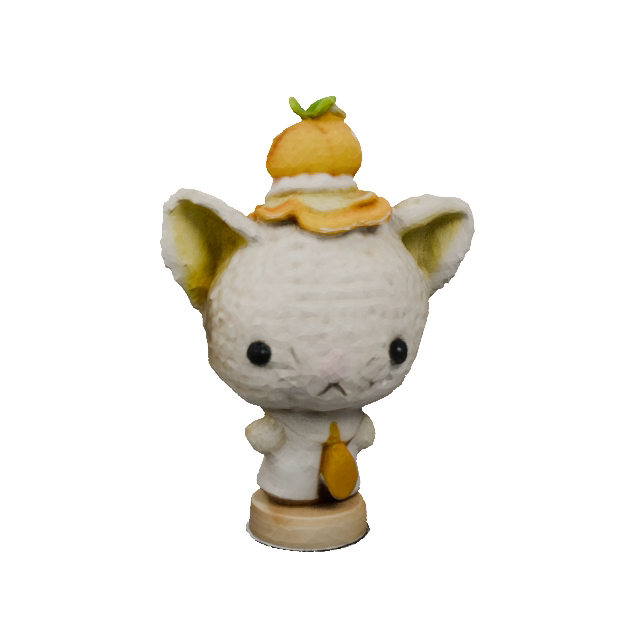}
    {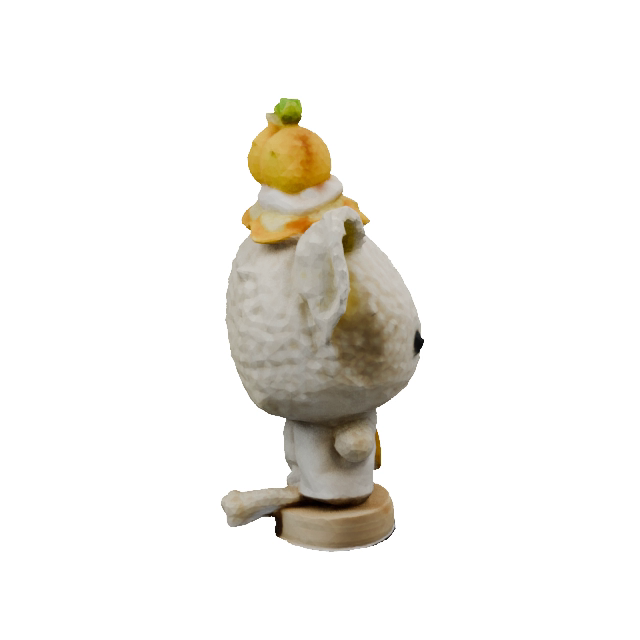}
    {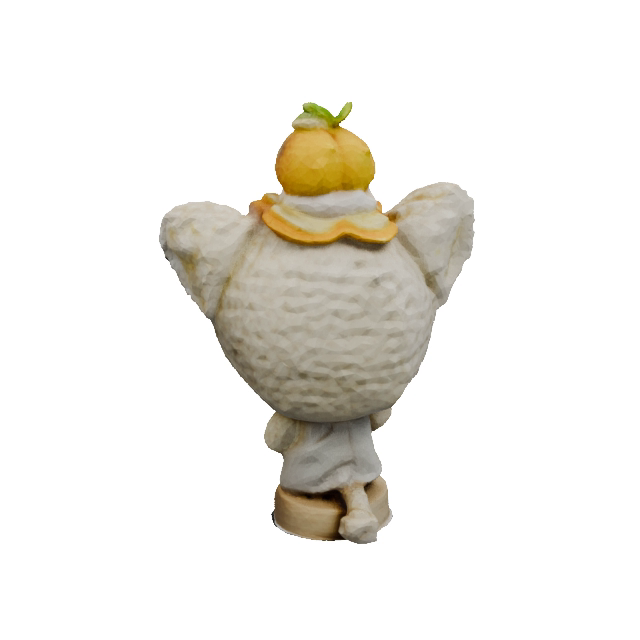}
    {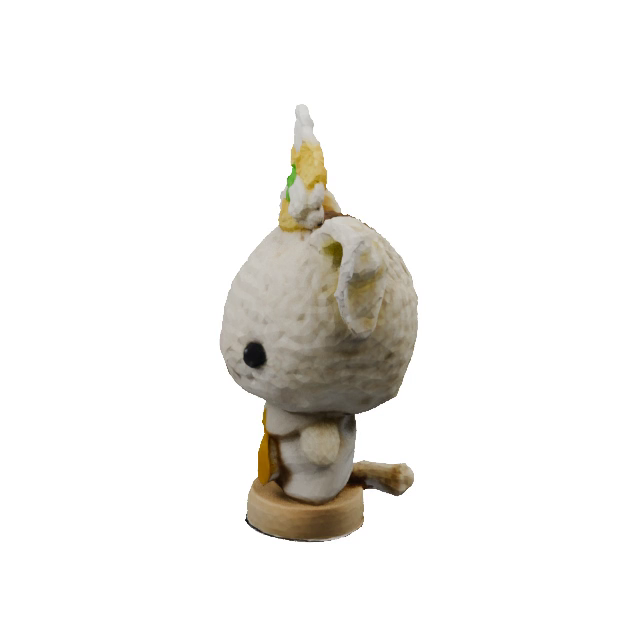}
    {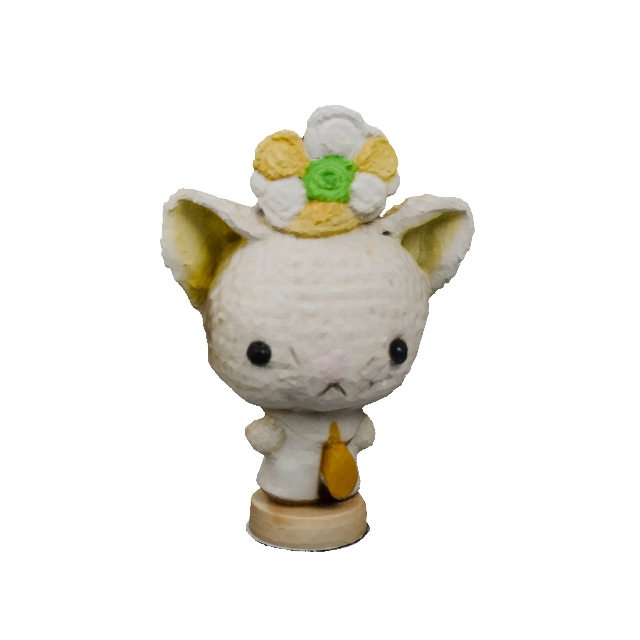}
    {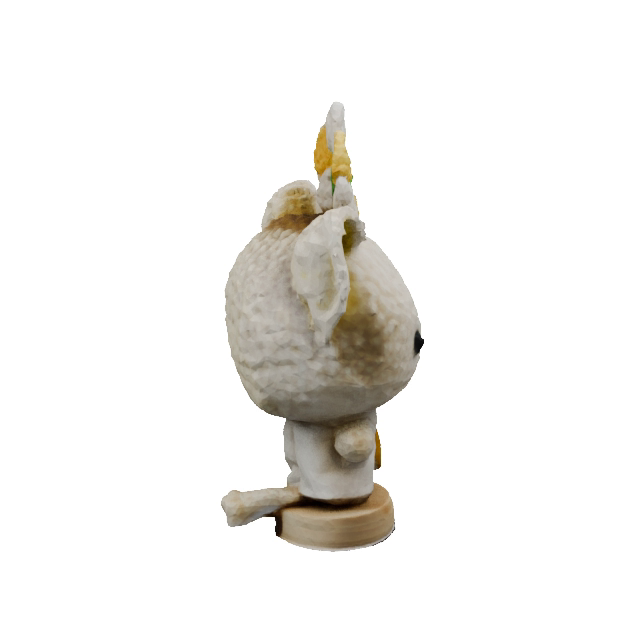}
    {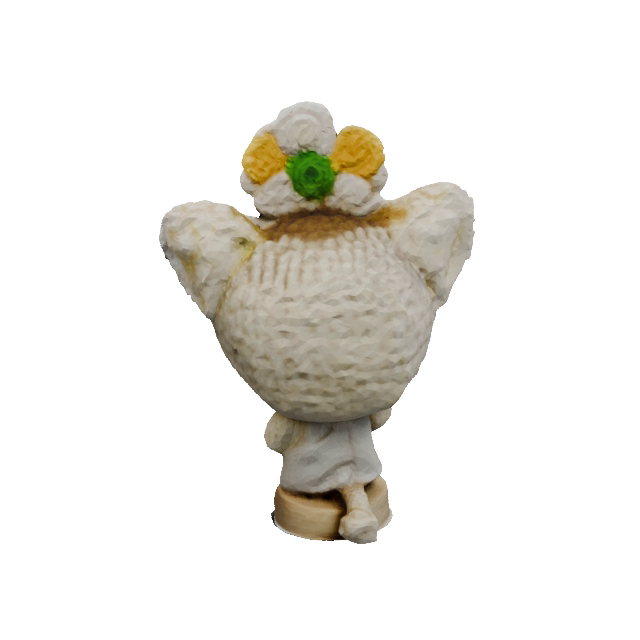}
\\[9mm]

\aEightFinalRow
    {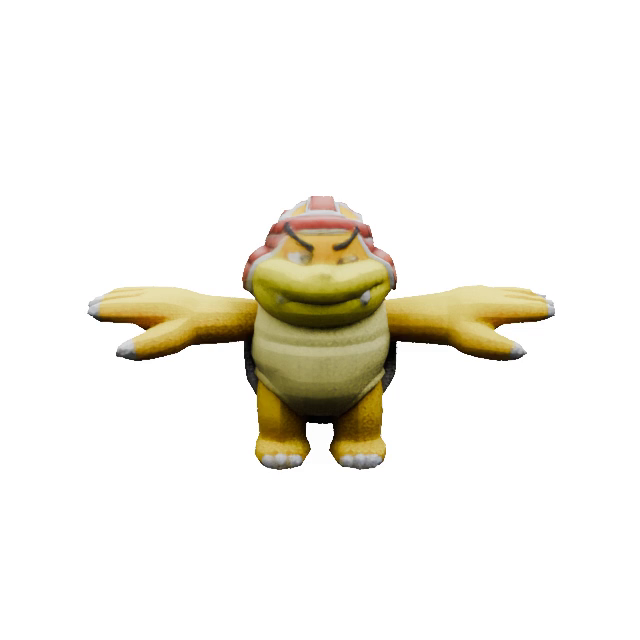}
    {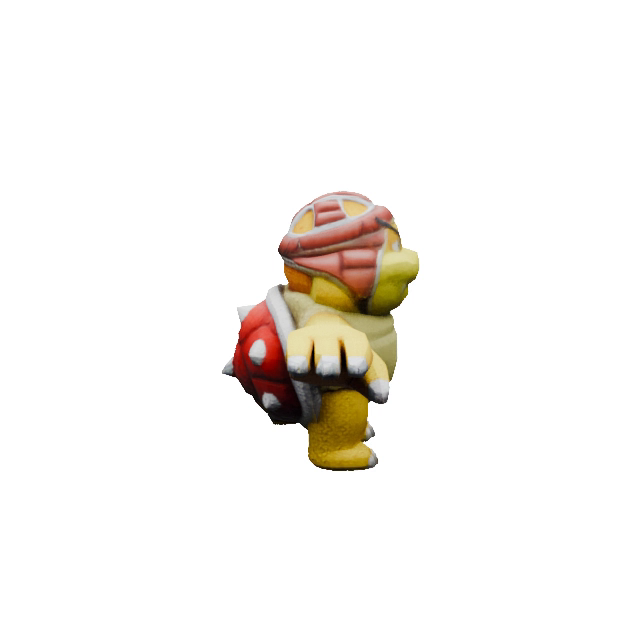}
    {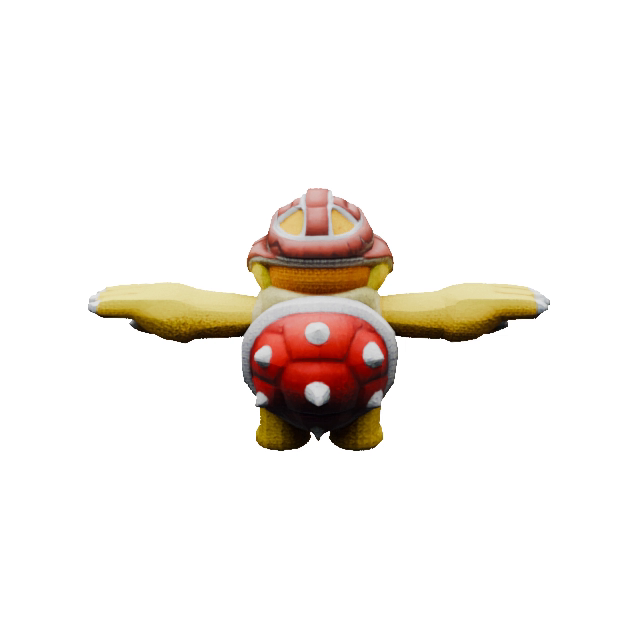}
    {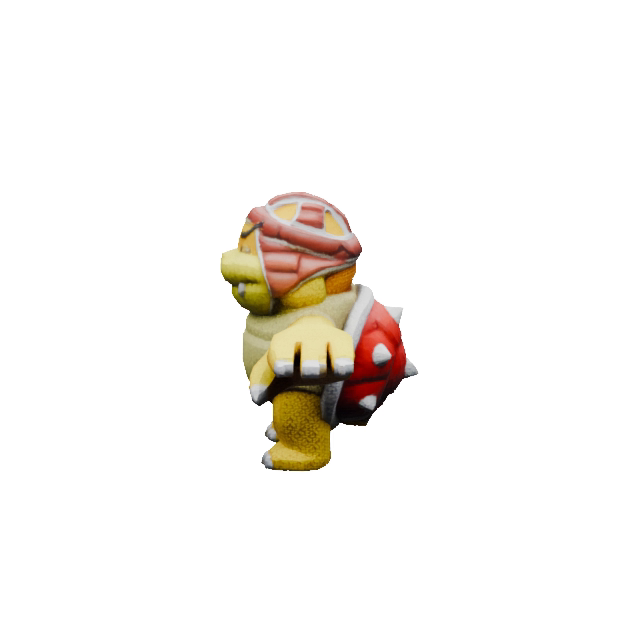}
    {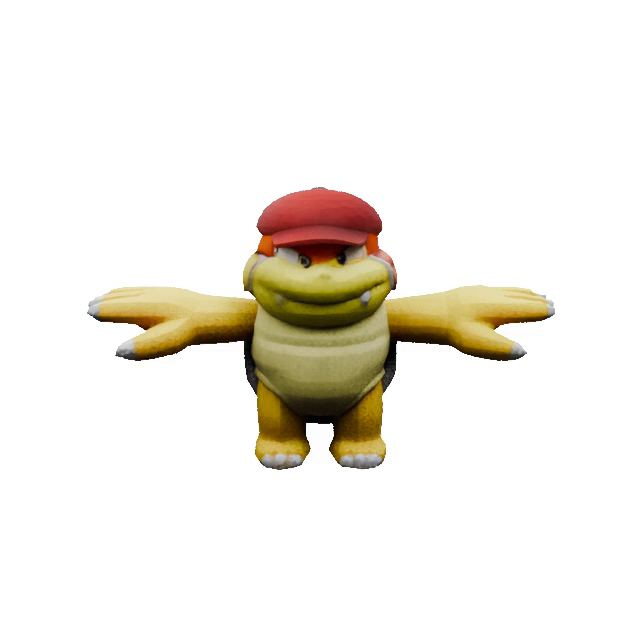}
    {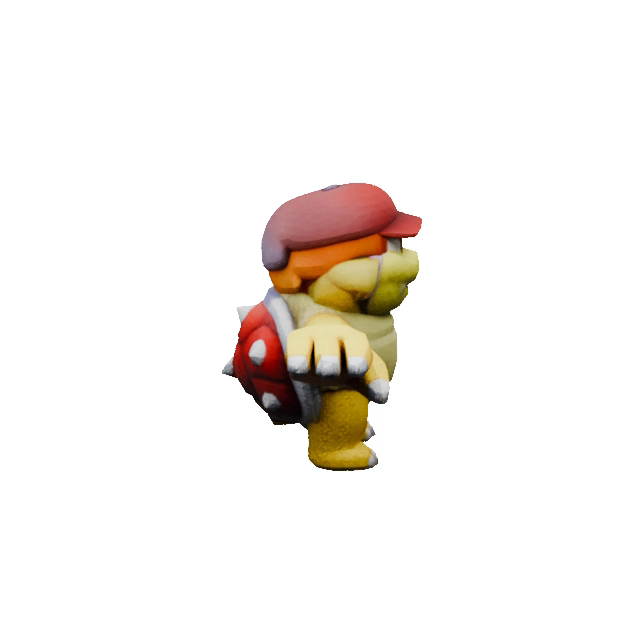}
    {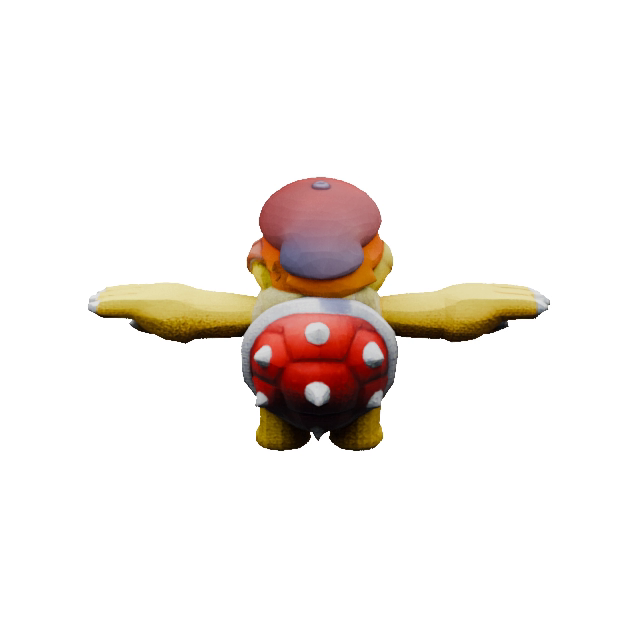}
    {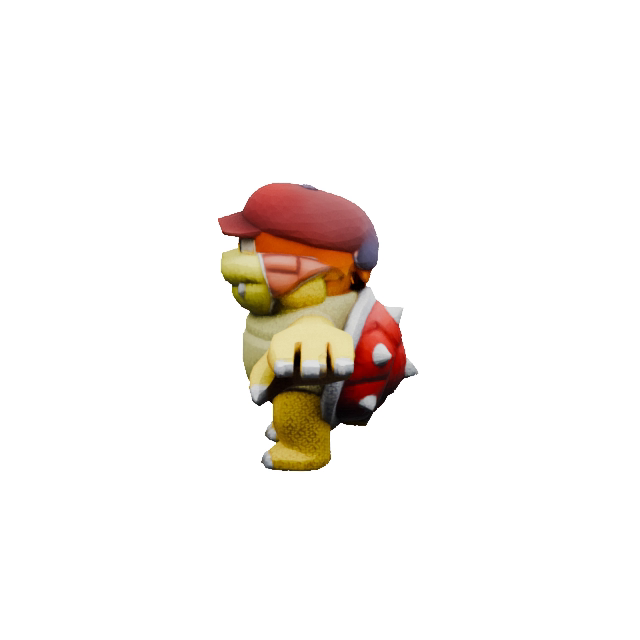}
\\[9mm]

\aEightFinalRow
    {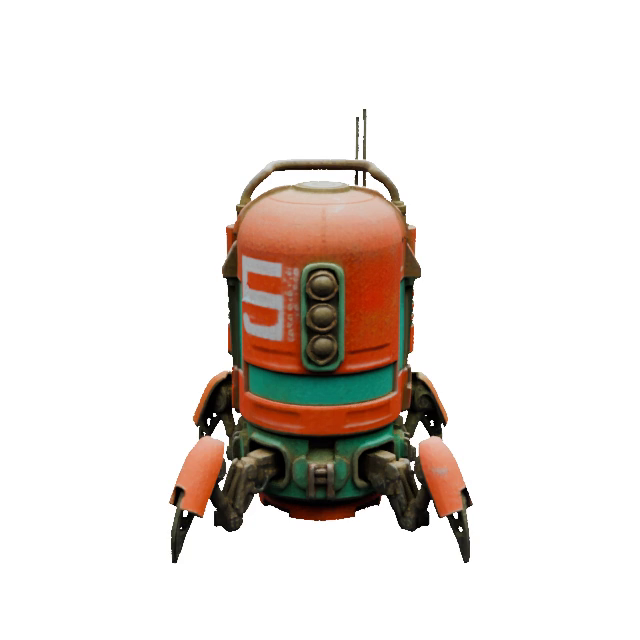}
    {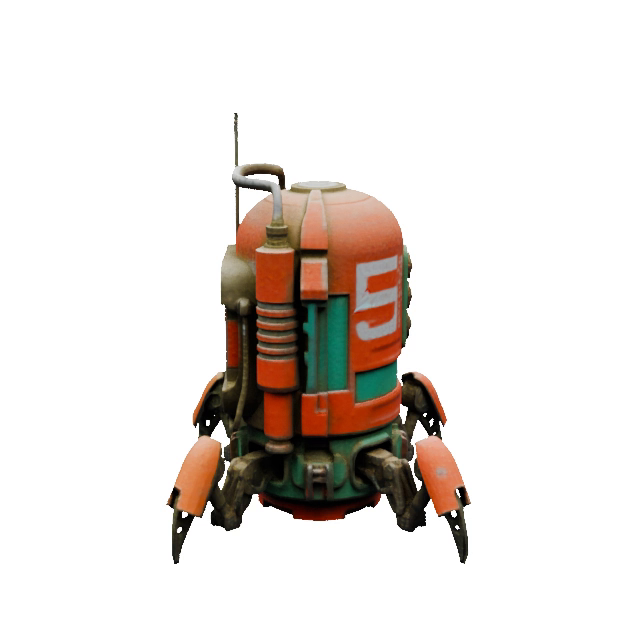}
    {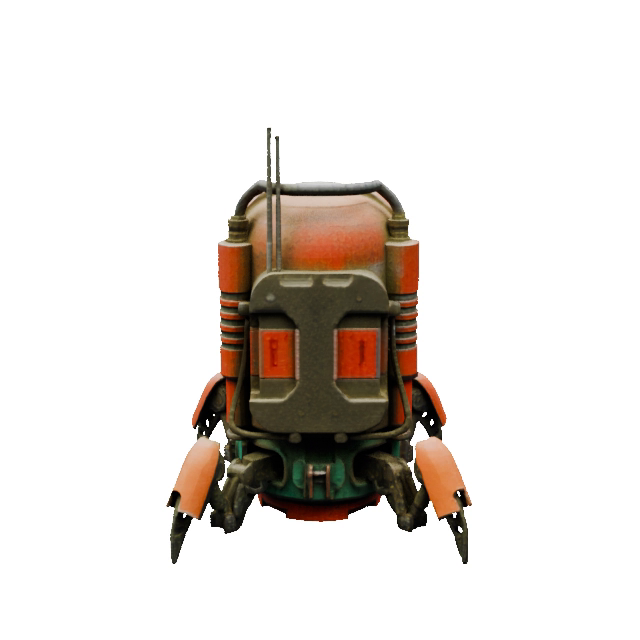}
    {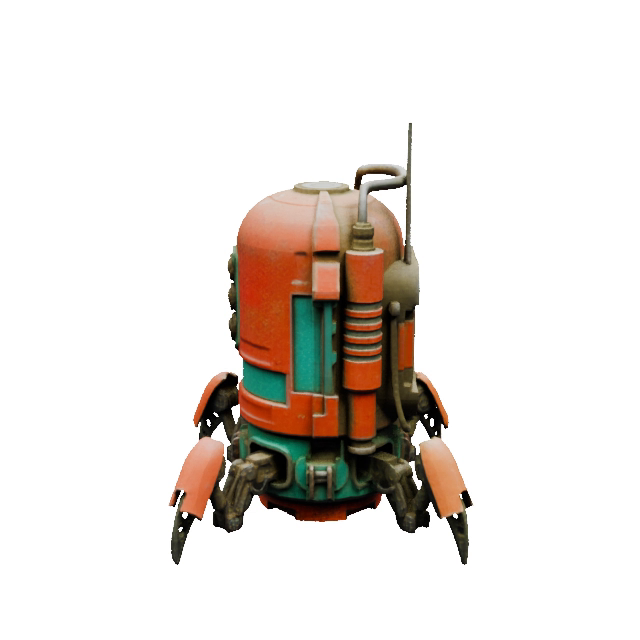}
    {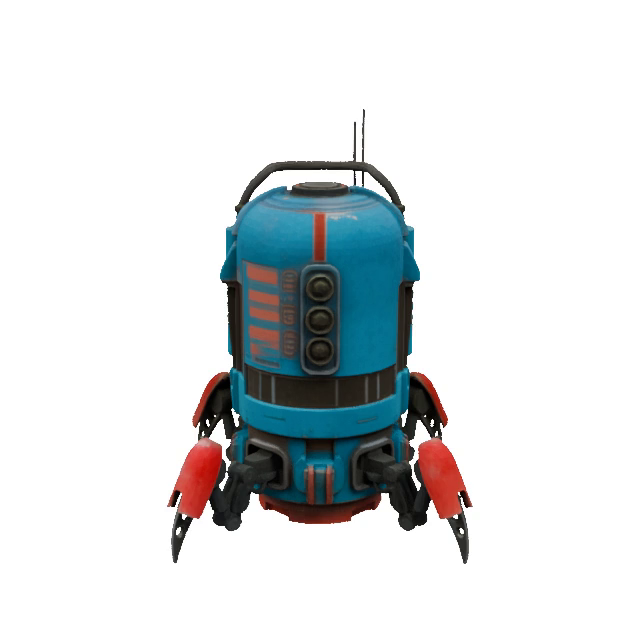}
    {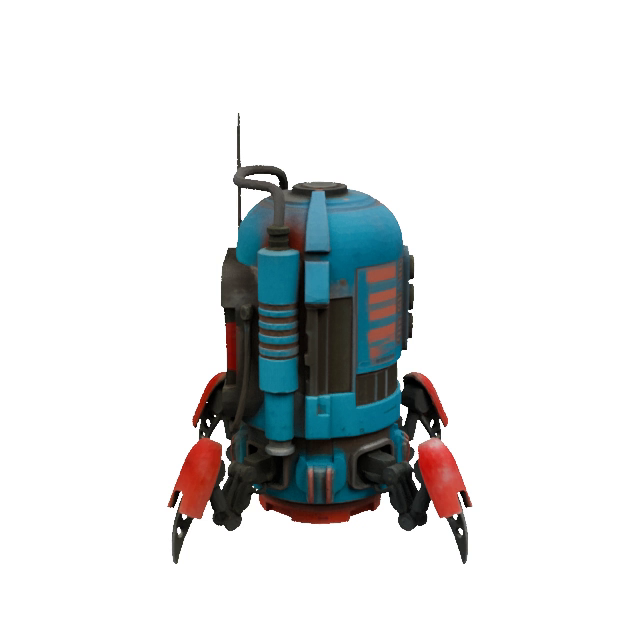}
    {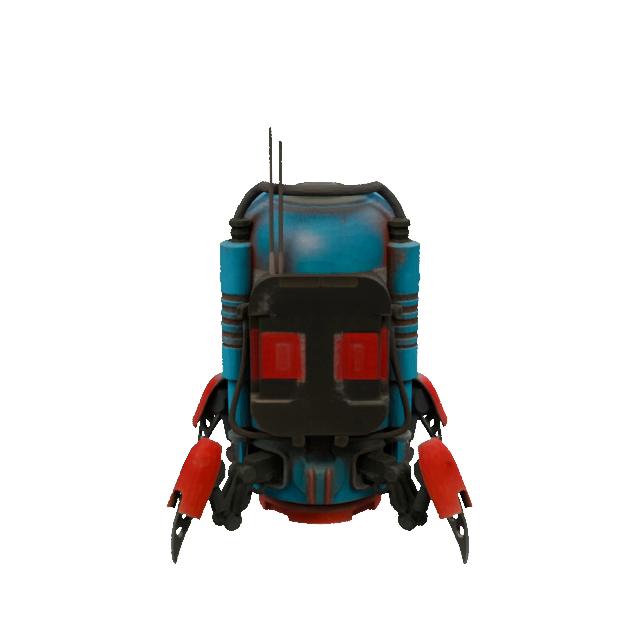}
    {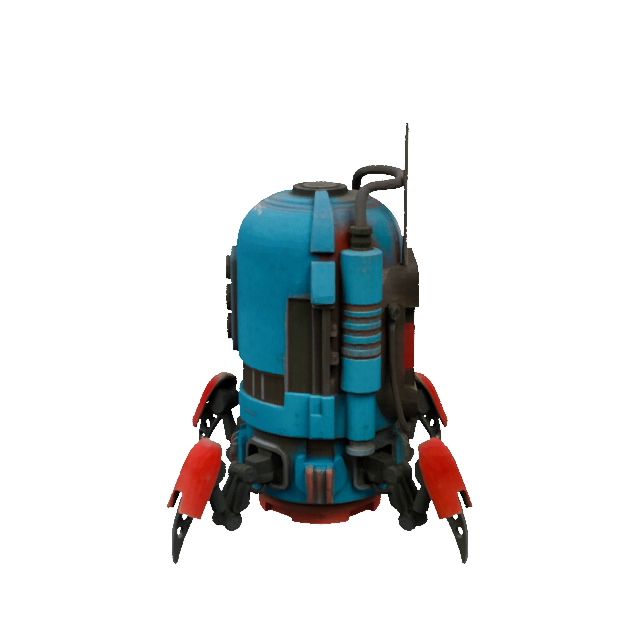}
\\[9mm]

\aEightFinalRow
    {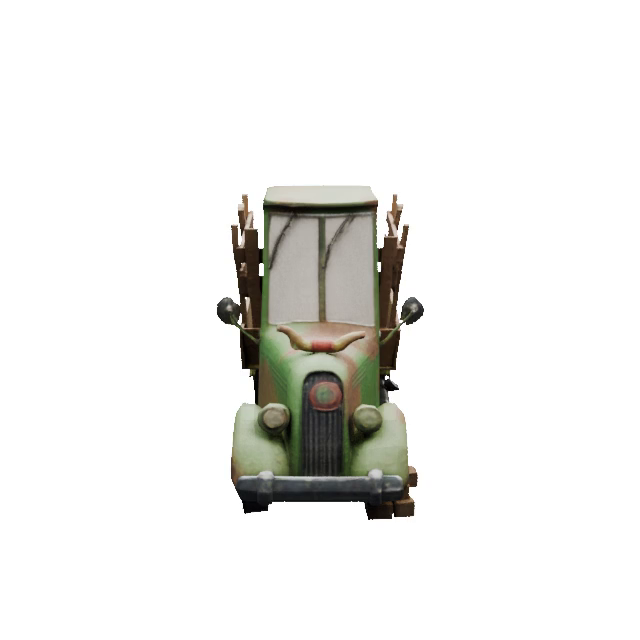}
    {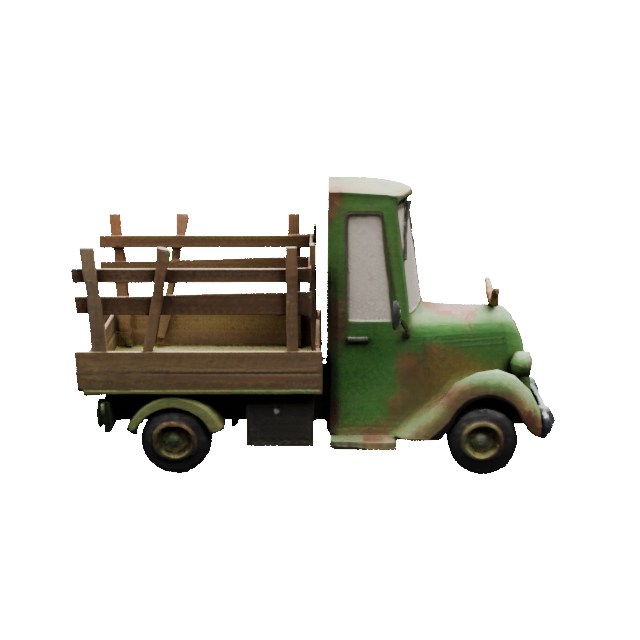}
    {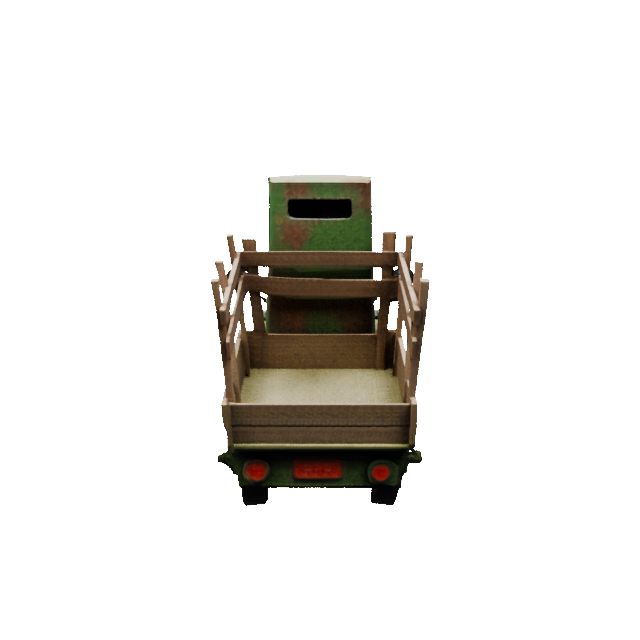}
    {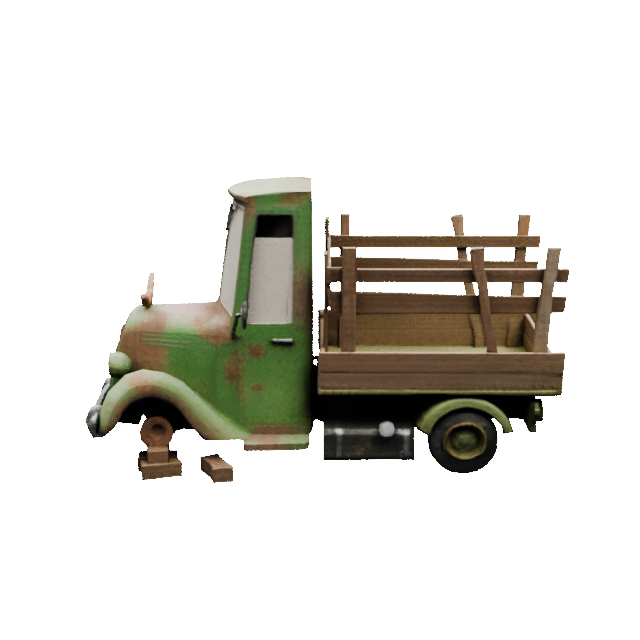}
    {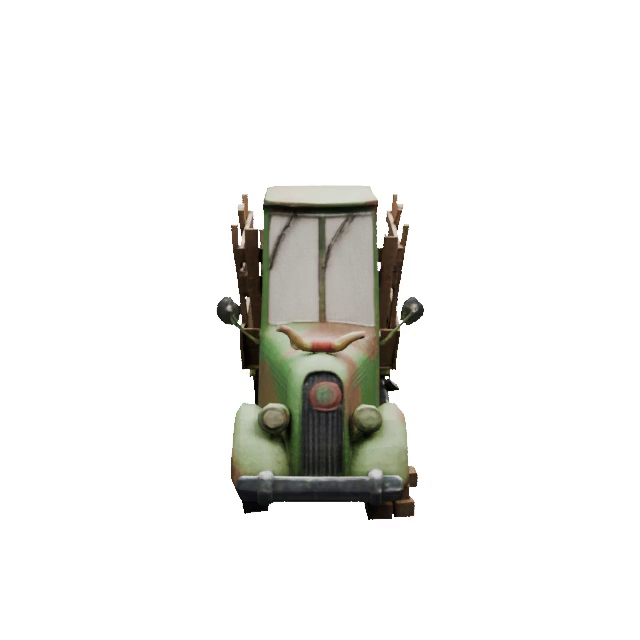}
    {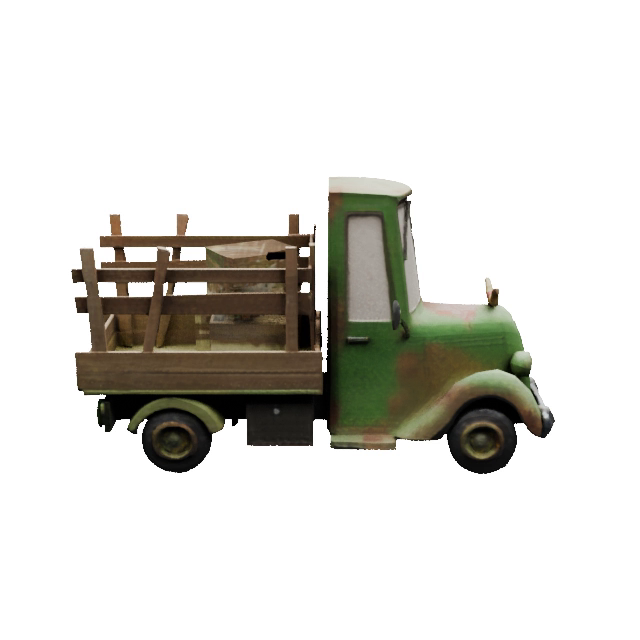}
    {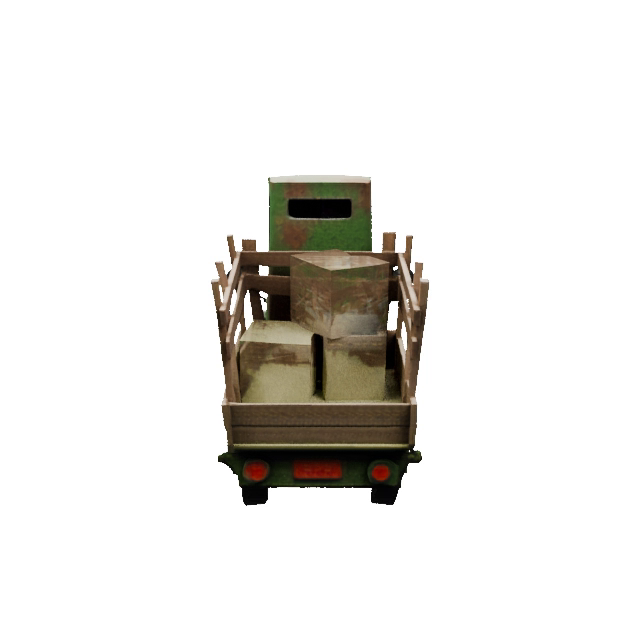}
    {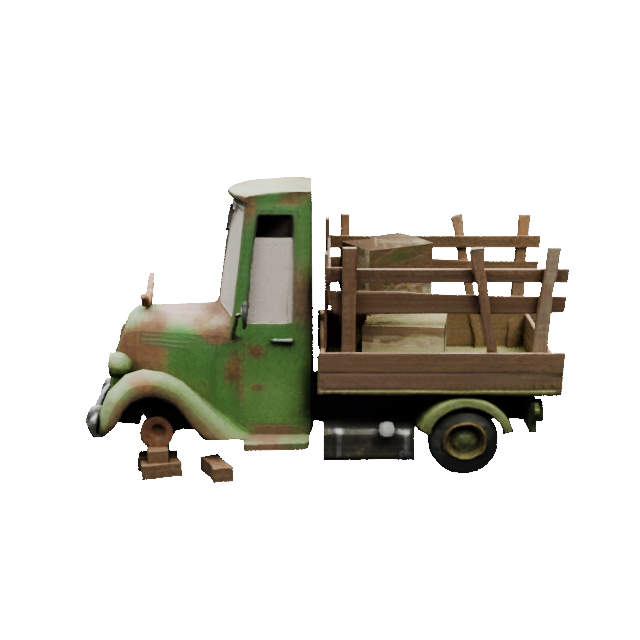}
\\[9mm]
\aEightFinalRow
    {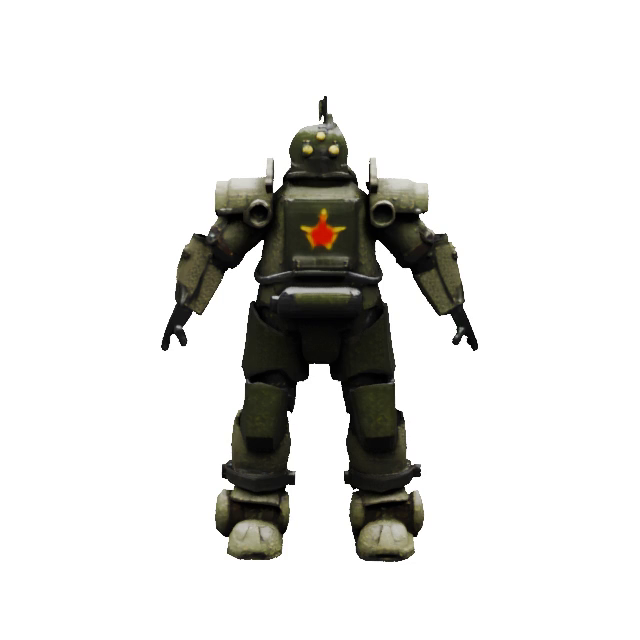}
    {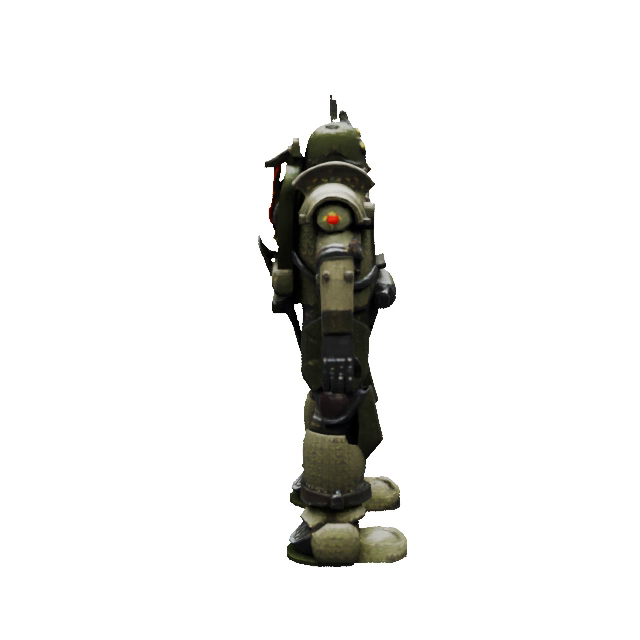}
    {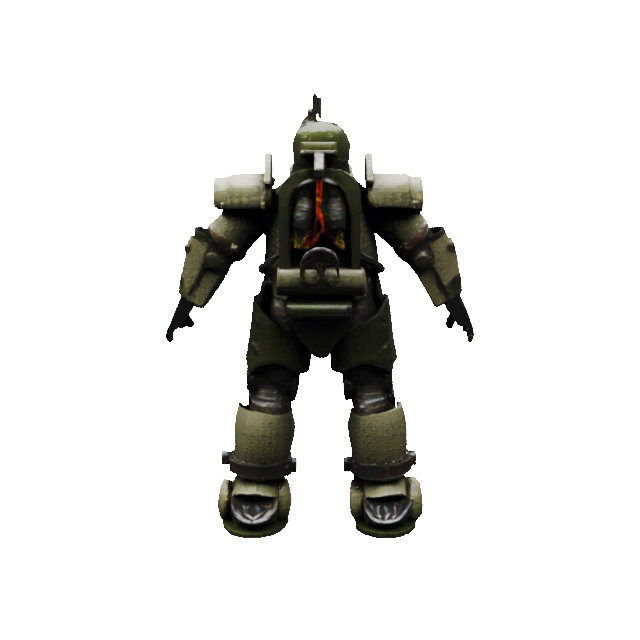}
    {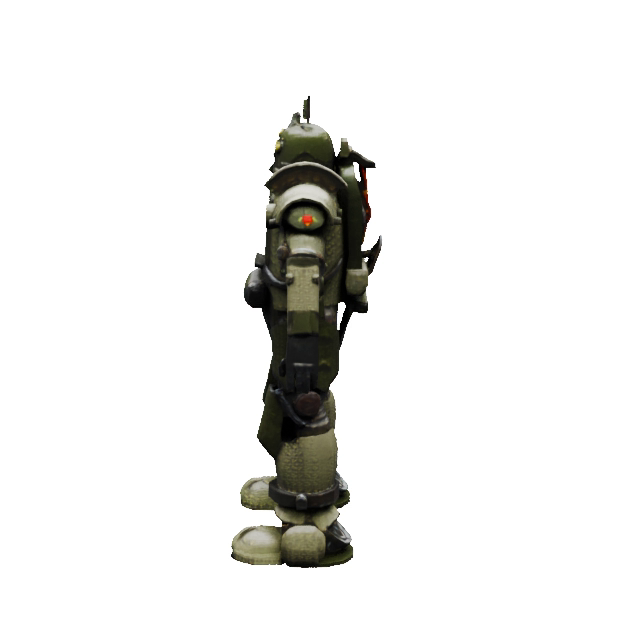}
    {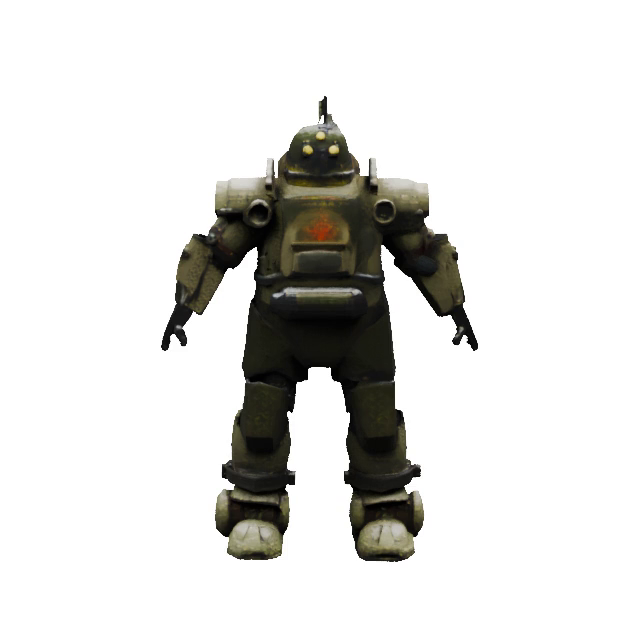}
    {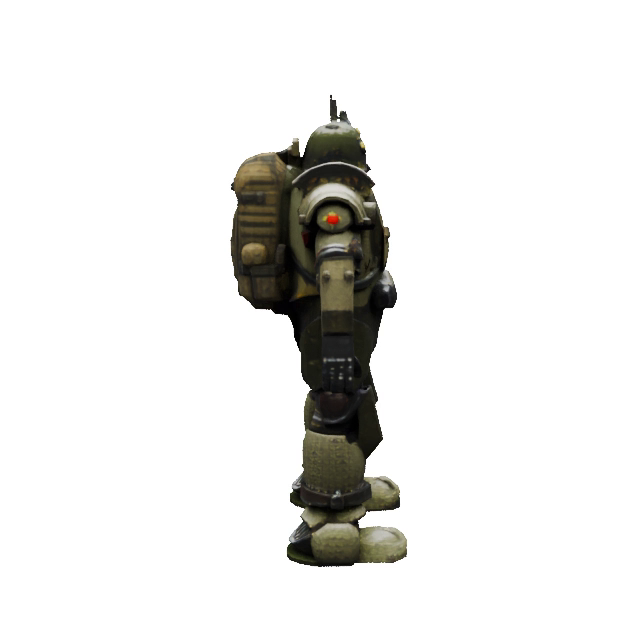}
    {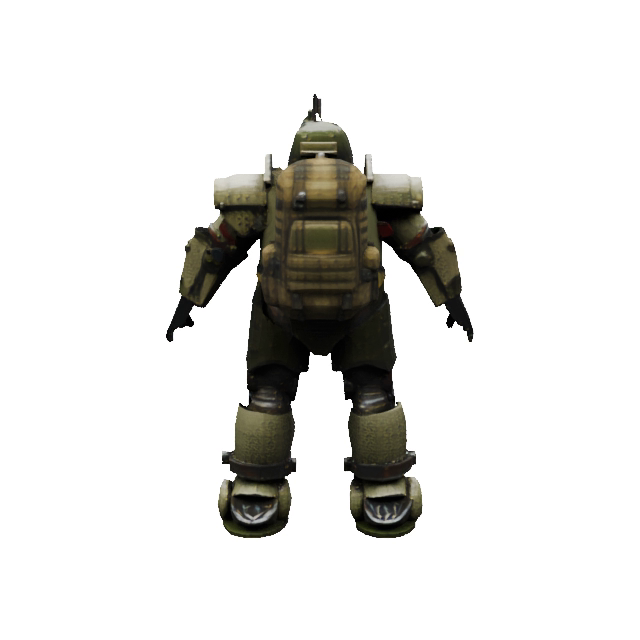}
    {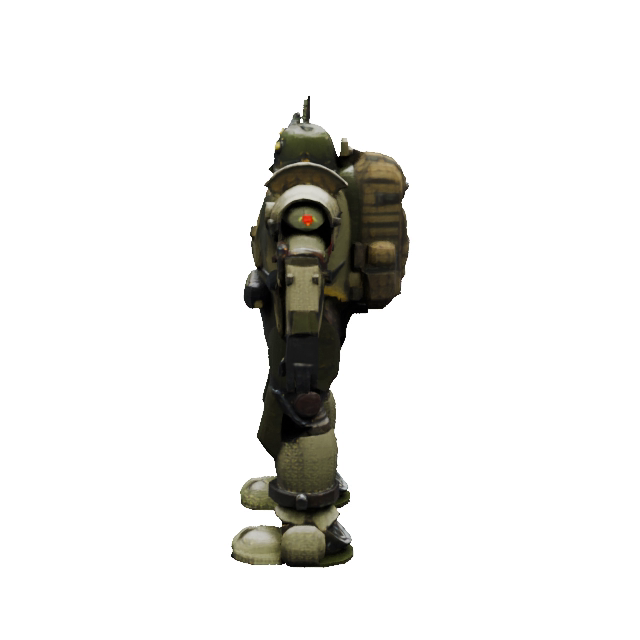}

\end{tabular}

\caption{
Additional qualitative 3D editing results of UMAS3D.
Each dashed box presents four views of the source 3D and four corresponding
views of the edited 3D, separated by a vertical dashed line.
}
\label{fig:A6}

\end{figure*}

To further demonstrate the editing capability of UMAS3D across different
tasks and object categories, Figure~\ref{fig:A6} presents additional
qualitative results covering Addition, Deletion, Modification, Appearance,
and Pose. For each example, we show four corresponding views of the source
3D asset and the final edited 3D result, enabling direct inspection of both
the requested modification and the preservation of unrelated regions.

The results show that UMAS3D can perform diverse structural and appearance
edits while largely preserving the geometry and visual attributes of
unrelated source regions. In addition to samples from Pxform and Omni3DEdit,
we include several Objaverse assets to illustrate the applicability of the
method to more diverse object categories. These additional samples are used
only for qualitative visualization and are excluded from all quantitative
evaluations.

\end{document}

%% file: math_commands.tex
\usepackage{amsmath,amsfonts,bm}

\def\eqref#1{equation~\ref{#1}}

\def\1{\bm{1}}

\DeclareMathAlphabet{\mathsfit}{\encodingdefault}{\sfdefault}{m}{sl}
\SetMathAlphabet{\mathsfit}{bold}{\encodingdefault}{\sfdefault}{bx}{n}



%% file: secs/intro.tex
Recent advances~\citep{xiang2025structured,zhao2025hunyuan3d} in 3D generative models and representation techniques have substantially improved 3D asset quality, enabling applications in digital entertainment, AR/VR, and interactive content creation. Building on this progress, controllable 3D editing~\citep{barda2025instant3dit,bar2025editp23,xia2025towards} further enhances content-creation flexibility by enabling users to modify and customize existing 3D assets under specified editing conditions. Given a source 3D asset and an editing condition, the goal is to realize the requested modification while preserving the geometry, appearance, and identity of unaffected regions.

Existing 3D editing methods can be broadly categorized into training-based and training-free approaches. Training-based methods learn editing transformations from paired 3D data through additional model optimization~\citep{gat2026shapeup,ma2025feedforward}, whereas training-free methods reuse pretrained 3D generative priors through mechanisms such as source-3D inversion, source--target trajectory manipulation, or latent-space operations~\citep{li2026voxhammer,chi2026vinedresser3d,zhou2026anchorflow}. Existing methods also differ in the user-facing editing conditions and editing tasks they support~\citep{barda2025instant3dit,bar2025editp23,xia2025towards}. For localized editing, some approaches additionally rely on a 3D editing mask either provided as input or estimated before generation to specify where the edit is applied~\citep{barda2025instant3dit,li2026voxhammer,nie2026editflow3d}.

These differences motivate three requirements for a more general 3D editor. \textbf{First}, editing intent may be expressed through text, bounding boxes, reference images, or their combinations; mapping these inputs to a common interface allows them to share the same downstream editing process. \textbf{Second}, a training-free formulation avoids dependence on paired 3D editing data and editing-specific model optimization, enabling the same pretrained generative prior to be reused across tasks. \textbf{Third}, fixed edit regions computed before generation cannot reflect subsequent changes in the target geometry. As the target evolves, such a region may exclude areas that later require editing or include source content that should remain unchanged, resulting in incomplete edits or unintended modifications. This motivates adjusting edit localization during generation to keep it aligned with the evolving target. The resulting challenge is to unify diverse conditions and tasks in a training-free framework while supporting adaptive edit localization.

To address this challenge, we propose~\textbf{UMAS3D}, a unified 3D editing framework that operates without additional model training, source-3D inversion, or externally supplied 3D masks. As illustrated in Figure~\ref{fig:1}, \textbf{UMAS3D} achieves two levels of unification: text, bounding boxes, reference images, and their combinations are converted into common target-view guidance, while addition, deletion, modification, appearance, and pose editing share the same downstream 3D editing pipeline. Its core editing mechanism consists of two complementary modules. \textbf{Mask-Free Adaptive Region Modeling (MARM)} constructs a coarse Edit/Keep prior from multi-view source–target visual differences and adaptively refines the regional assignments as the target structure evolves. \textbf{Source-Guided Preservation Steering (SGPS)} then applies representation-specific source constraints only within the current Keep regions while leaving Edit regions target-driven. Thus, MARM determines where source content should be retained, whereas SGPS controls how it is preserved, allowing source preservation to follow the evolving target while keeping source steering spatially localized.

We evaluate UMAS3D on 150 source objects from Pxform~\citep{weng2026feedforward} and Omni3DEdit~\citep{omni3dedit2026}, covering five editing tasks. Under a controlled fixed-target-view protocol, UMAS3D achieves the best reported results across all evaluated metrics among the compared baselines.  Qualitative results under five constructed multimodal input settings and a user study provide additional evidence of the framework’s flexibility and perceptual quality.

Overall, our contributions can be summarized as follows:
\begin{itemize}[leftmargin=*, topsep=2pt, itemsep=1pt, parsep=0pt, partopsep=0pt]
   \item We present UMAS3D, a unified framework that supports heterogeneous conditions and five editing tasks through a shared 3D editing pipeline, without additional training, source-3D inversion, or externally supplied 3D masks.
    
   \item We introduce Mask-Free Adaptive Region Modeling (MARM) and Source-Guided Preservation Steering (SGPS) for adaptive source supervision. MARM updates an explicit Edit/Keep partition from source–target evidence, while SGPS preserves source content only within the current Keep regions.
    
   \item Experiments on 150 objects show that UMAS3D achieves the best reported results across all evaluated metrics among the compared baselines. Ablations, multimodal results, and a user study further support its effectiveness.
 \end{itemize}